\documentclass{bmvc2k}

\usepackage{times}
\usepackage{epsfig}
\usepackage{booktabs}
\usepackage{algorithm, algpseudocode}
\usepackage{overpic}
\usepackage{amsmath,amssymb,amsfonts}
\usepackage{graphicx}
\usepackage{caption}
\usepackage{subcaption}
\usepackage{cancel}
\usepackage{diagbox}
\usepackage{textcomp}
\usepackage{xcolor}
\usepackage{color}
\usepackage{hyperref}
\usepackage{cite}

\def\A{\mathbf{A}}

\def\I{\mathbf{I}}

\def\e{\mathbf{e}}

\def\g{\mathbf{g}}

\def\x{\mathbf{x}}
\def\y{\mathbf{y}}
\def\z{\mathbf{z}}

\def\bSigma{\boldsymbol{\Sigma}}

\def\bepsilon{\boldsymbol{\epsilon}}

\def\bmu{\boldsymbol{\mu}}

\title{ADIR: Adaptive Diffusion for Image Reconstruction}

\addauthor{Shady Abu-Hussein}{shady.abh@gmail.com}{1}
\addauthor{Tom Tirer}{tirer.tom@biu.ac.il}{1}
\addauthor{Raja Giryes}{raja@tauex.tau.ac.il}{2}

\addinstitution{
 Tel Aviv University\\
 Tel Aviv, Israel
}
\addinstitution{
 Bar Ilan University\\
 Ramat Gan, Israel
}
\addinstitution{
 Tel Aviv University\\
 Tel Aviv, Israel
}

\runninghead{Abu-Hussein et. al}{ADIR: Adaptive Diffusion for Image Reconstruction}

\begin{document}

\maketitle

\begin{abstract}
Denoising diffusion models have recently achieved remarkable success in image generation, capturing rich information about natural image statistics. This makes them highly promising for image reconstruction, where the goal is to recover a clean image from a degraded observation. In this work, we introduce a conditional sampling framework that leverages the powerful priors learned by diffusion models while enforcing consistency with the available measurements. 
To further adapt pre-trained diffusion models to the specific degradation at hand, we propose a novel fine-tuning strategy. In particular, we employ LoRA-based adaptation using images that are semantically and visually similar to the degraded input, efficiently retrieved from a large and diverse dataset via an off-the-shelf vision–language model. 
We evaluate our approach on two leading publicly available diffusion models—Stable Diffusion and Guided Diffusion—and demonstrate that our method, termed \textbf{A}daptive \textbf{D}iffusion for \textbf{I}mage \textbf{R}econstruction (\textbf{ADIR}), yields substantial improvements across a range of image reconstruction tasks. Code is available at \url{https://github.com/shadyabh/ADIR}.
\end{abstract}

\maketitle

\section{Introduction}
Image reconstruction problems appear in a wide range of applications, where one would like to reconstruct an unknown clean image $\x \in\mathbb{R}^n$ from its degraded version $\y \in\mathbb{R}^m$, which can be noisy, blurry, low-resolution, etc. The acquisition (forward) model of $\y$ in many important degradation settings can be formulated using the following linear model
\begin{equation}
    \y = \mathbf{A}\x + \e,
    \label{eq:observ_model}
\end{equation}
where $\mathbf{A}\in\mathbb{R}^{m\times n}$ is the measurement operator (blurring, masking, sub-sampling, etc.) and $\e \in \mathbb{R}^m \sim \mathcal{N}(0, \sigma^2 \I_m)$ is the measurement noise. 
Typically, just fitting the observation model is not sufficient for recovering $\x$ successfully; thus, prior knowledge of the characteristics of $\x$ is needed.

Over the past decade, many works suggested solving the inverse problem in \eqref{eq:observ_model} using 
a single execution of a deep neural network, trained using pairs of clean $\{\x_{i}\}$ images and their degraded versions $\{\y_i\}$ obtained by applying the forward model in \eqref{eq:observ_model} on $\{\x_{i}\}$ \citep{dong2015image,Sun2015LearningAC,lim2017enhanced,zhang2017beyond,lugmayr2020srflow,Liang2021SwinIR}. 
Yet, these approaches tend to overfit the observation model and perform poorly on setups that have not been considered in training and several methods have been proposed to overcome that \citep{shocher2018zero,tirer2019super,corrFlt,Ji2020Real,wei2020cdc,wang2021real,Zhang_2021_ICCV,zhang2022practical}.

\begin{figure}[ht]
    \centering

    \begin{subfigure}[b]{0.24\linewidth}
        \centering
        \includegraphics[width=\linewidth, trim={0 12 0 11}, clip]{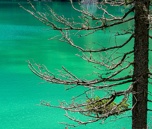}
        \caption{Ground Truth}
    \end{subfigure}
    \begin{subfigure}[b]{0.24\linewidth}
        \centering
        \includegraphics[width=\linewidth, trim={0 12 0 11}, clip]{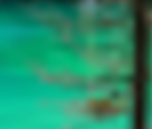}
        \caption{Bicubic $\times$8}
    \end{subfigure}
    \begin{subfigure}[b]{0.24\linewidth}
        \centering
        \includegraphics[width=\linewidth, trim={0 12 0 11}, clip]{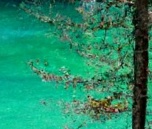}
        \caption{Guided Diffusion}
    \end{subfigure}
    \begin{subfigure}[b]{0.24\linewidth}
        \centering
        \includegraphics[width=\linewidth, trim={0 12 0 11}, clip]{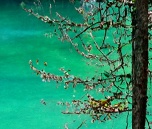}
        \caption{ADIR (ours)}
    \end{subfigure}\\

    \begin{subfigure}[b]{0.24\linewidth}
        \centering
        \includegraphics[width=\linewidth, trim={40 120 120 0}, clip]{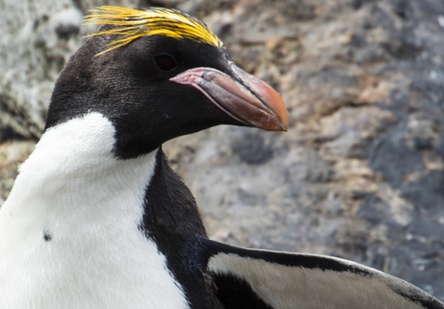}
        \caption{Ground Truth}
    \end{subfigure}
    \begin{subfigure}[b]{0.24\linewidth}
        \centering
        \includegraphics[width=\linewidth, trim={40 120 120 0}, clip]{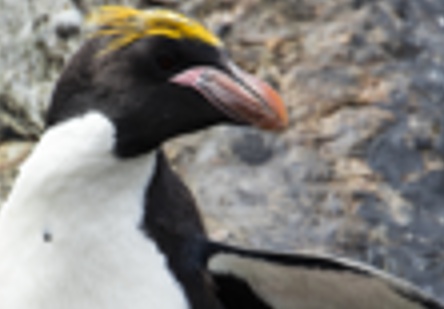}
        \caption{Bicubic $\times$4}
    \end{subfigure}
    \begin{subfigure}[b]{0.24\linewidth}
        \centering
        \includegraphics[width=\linewidth, trim={40 120 120 0}, clip]{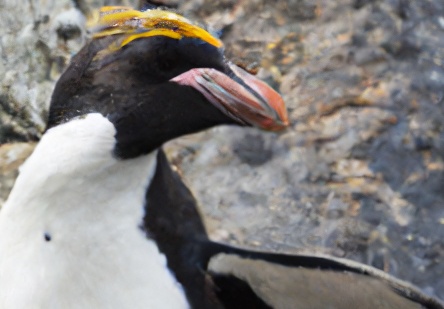}
        \caption{Stable Diffusion}
    \end{subfigure}
    \begin{subfigure}[b]{0.24\linewidth}
        \centering
        \includegraphics[width=\linewidth, trim={40 120 120 0}, clip]{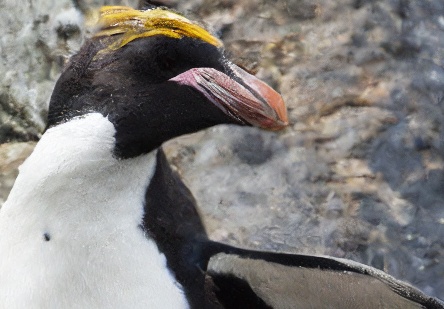}
        \caption{ADIR (ours)}
    \end{subfigure}\\
    \caption{Super-resolution with scale factors 4 and 8, using Stable Diffusion \citep{rombach2022high}, Guided Diffusion \citep{guidedDiff}, and our method ADIR.}
    \label{fig:LDM_SR_teaser}
    \vspace{-15pt}
\end{figure}

Several approaches such as Deep Image Prior \citep{ulyanov2018deep}, zero-shot-super-resolution \citep{shocher2018zero} or GSURE-based test-time optimization \citep{PGSURE} rely solely on the observation image $\y$. They utilize the implicit bias of deep neural networks and gradient-based optimizers, as well as
the self-recurrence of patterns in natural images when training a neural model directly on the observation and in this way reconstruct the original image.
Although these methods are not limited to a family of observation models, they usually perform worse than data-driven methods. 
The alternative popular approach that exploits external data while remaining flexible to the observation model, uses deep models for imposing only the prior. It typically uses pretrained deep denoisers \citep{zhang2017learning,arjomand2017deep,tirer2018image,zhang2021plug} or generative models \citep{bora2017compressed,dhar2018modeling,IAGAN} within the optimization scheme, where consistency of the reconstruction with the observation $\y$ is maintained by minimizing a data-fidelity term.  

Recently, diffusion models \citep{guidedDiff, nichol2021improved, sohl2015deep, DDPM} have shown remarkable capabilities in generating high-fidelity images and videos \citep{ho2022video}. 
These models are based on a Markov chain diffusion process performed on each training sample. They learn the reverse process, namely, the denoising operation between each two points in the chain. 
Sampling images via pretrained diffusion models is performed by starting from a pure white Gaussian noise image, then progressively denoise and sample a less noisy image, until reaching a clean image.
Since diffusion models capture prior knowledge of the data, one may utilize them as deep priors/regularization for inverse problems\citep{song2021solving,lugmayr2022repaint,avrahami2022blended,kawar2022denoising,choi2021ilvr,rombach2022high}.

In this work, we propose an Adaptive Diffusion framework for Image Reconstruction (ADIR). 
First, we devise a diffusion guidance sampling scheme that solves \eqref{eq:observ_model} while restricting the reconstruction of $\x$ to the range of a pretrained diffusion model.
Then, we propose a technique that uses the observations $\y$ to adapt the diffusion network to patterns beneficial for recovering the unknown $\x$. 
Adapting the model's parameters is based on $K$ external images similar to $\y$ in some neural embedding space that is not sensitive to the degradation of $\y$. These images are retrieved from a diverse dataset and the embedding can be calculated using an off-the-shelf encoder model for images such as CLIP \citep{CLIP}.

This work mainly focuses on image reconstruction tasks, yet, we also showcase that ADIR can be employed for text-guided image editing. Note that for the latter, we just show the potential of our strategy and that it can be combined with existing editing techniques. We leave further exploration of applying ADIR to editing to a future work. 

\begin{figure}[t]
    \centering
    \includegraphics[width=0.9\linewidth]{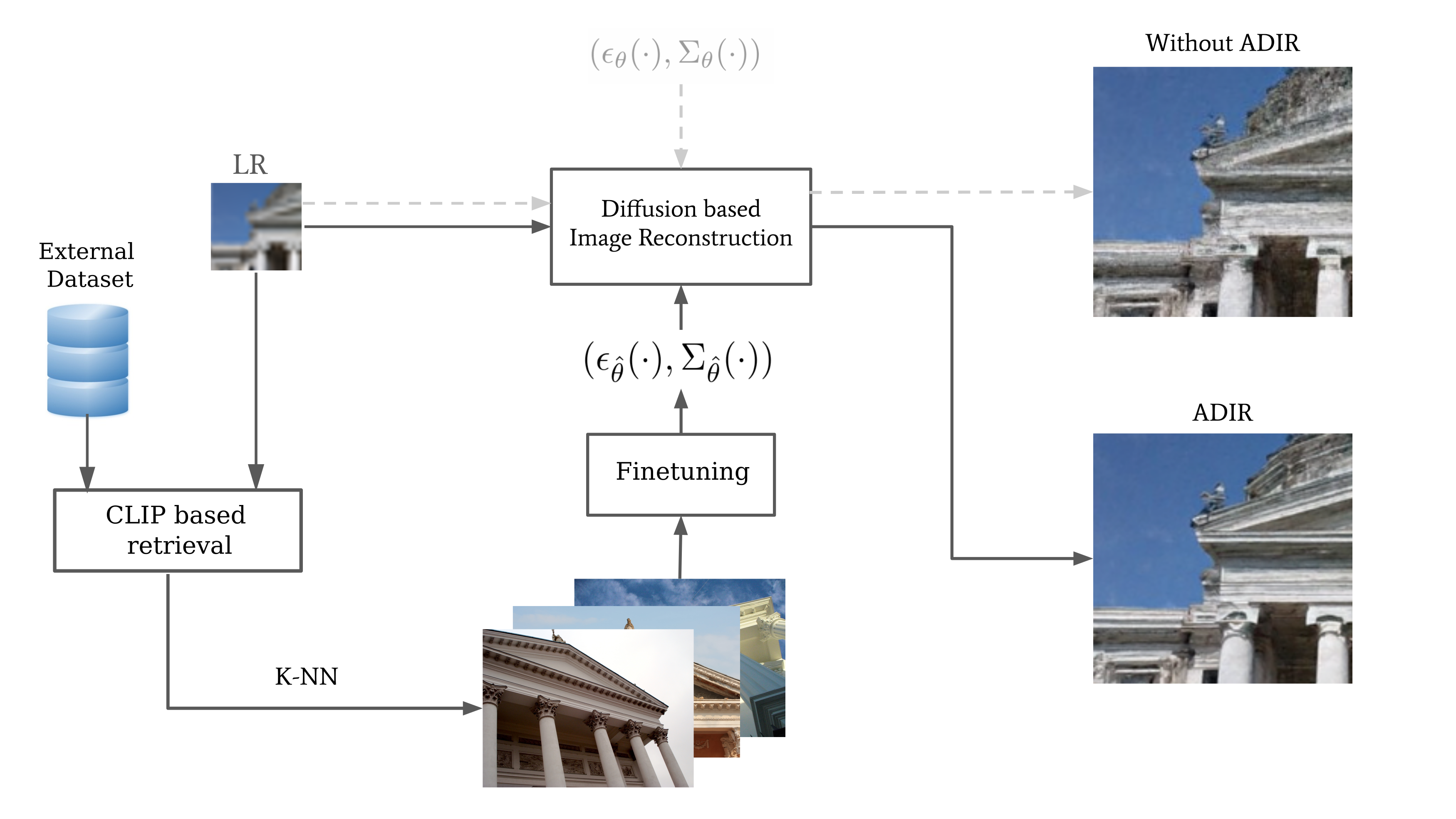}
    \caption{\small Diagram of our proposed method ADIR (Adaptive Diffusion for Image Reconstruction) applied to the super resolution task. Given a pretrained diffusion model $(\bepsilon_\theta(\cdot), \bSigma_\theta(\cdot))$ and a Low Resolution (LR) image, we look for the $K$ nearest neighbor images to the LR image, then using ADIR we adapt the diffusion model and use it for reconstruction.}
    \vspace{-10pt}
    \label{fig:ADIR}
\end{figure}

\section{Related Work}
\label{sec:related}

\paragraph{Diffusion models} Denoising diffusion models \citep{sohl2015deep, DDPM} are latent variable generative models, with latent variables $\x_1, \x_2,...,\x_T \in \mathbb{R}^n$.
Given a training sample $\x_0 \sim q_\x$, these models construct a Markov chain from $\x_0$ to $\x_T$ of the form
\begin{equation}
    q(\x_{1:T}|\x_0) := \prod_{t=1}^{T} q(\x_t|\x_{t-1}),\text{ where } q(\x_t|\x_{t-1}) := \mathcal{N}(\sqrt{ 1 - \beta_t}\x_{t-1}, \beta_t\I_n),
    \label{eq:forward_dist}
\end{equation}
By utilizing the Markov property of the process, $\x_t|\x_0$ can be obtained via the following parametrization  \citep{DDPM}:
\begin{align}
\label{eq:xt_given_x0}
    \x_t = \sqrt{\bar{\alpha}_t} \x_0 + \sqrt{1 - \bar{\alpha}_t}\bepsilon,\ \bepsilon \sim \mathcal{N}(\mathbf{0}, \I_n),
\end{align}
where $\bar{\alpha}_t := \prod_{s=1}^t\alpha_s$ and $\alpha_s := 1 - \beta_s$.
The goal in these models is to learn the distribution of the reverse chain from $\x_T$ to $\x_0$, which is parameterized as the Markov chain 
\begin{equation}
    p_{\theta}(\x_{0:T}) := p(\x_T)\prod_{t=1}^{T} p_{\theta}(\x_{t-1}|\x_{t}), \text{ where } 
    p_{\theta}(\x_{t-1}|\x_t) := \mathcal{N}(\bmu_{\theta}(\x_t,t), \bSigma_{\theta}(\x_t, t)). 
    \label{eq:backward_dist}
\end{equation}

The parameters $\theta$ of the diffusion model 
are optimized by minimizing evidence lower bound \citep{sohl2015deep}, a simplified score-matching loss \citep{DDPM,song2019generative}, or a combination of both \citep{guidedDiff,nichol2021improved}.
For example, the simplified loss involves the minimization of
\begin{align}
    \ell_{\text{simple}}(\x_0,\bepsilon_{\theta},t) = \| \bepsilon - \bepsilon_{\theta}( \sqrt{\bar{\alpha}_t}\x_0 + \sqrt{1-\bar{\alpha}_t}\bepsilon ,t) \|_2^2,
    \label{eq:diff_simple_loss}
\end{align}
w.r.t.~$\theta$ in each training iteration, where $\x_0$ is drawn from the training data, $t$ uniformly drawn from $\{1,...,T\}$ and the noise $\bepsilon \sim \mathcal{N}(\mathbf{0},\I_n)$.

Given a trained diffusion model $(\bepsilon_\theta(\x_t, t), \bSigma_\theta(\x_t, t))$, one may generate a sample $\x_0$ from the learned data distribution $p_\theta$ by initializing $\x_T \sim \mathcal{N}(\mathbf{0},\I_n)$ and 
running the reverse diffusion process
by sampling
\begin{align}
\label{eq:synthesis}
\x_{t-1} \sim \mathcal{N}(\bmu_{\theta}(\x_t,t), \bSigma_{\theta}(\x_t, t)), \text{ where  } \bmu_{\theta}(\x_t,t) := \frac{1}{\sqrt{\alpha_t}}(\x_t - \frac{1-\alpha_t}{\sqrt{1-\bar{\alpha}_t}}\bepsilon_{\theta}(\x_t,t)).
\end{align}

The class-guided sampling method that has been proposed in \citep{guidedDiff}  modifies the sampling procedure in \eqref{eq:synthesis} by adding to the mean of the Gaussian a term that depends on the gradient of an offline-trained classifier, which has been trained using noisy images $\{\x_t\}$ for each $t$, and approximates the likelihood $p_{c|\x_t}$, where $c$ is the desired class. 
This procedure has been shown to improve the quality of the samples generated for the learned classes.

  In recent years, many works utilized diffusion models for image manipulation and reconstruction tasks \citep{choi2021ilvr,rombach2022high, kawar2022imagic, kawar2022denoising, whang2022deblurring, SR3, zhu2023denoising, ozdenizci2023restoring, delbracio2023inversion,garber2023image}, where a denoising network is trained to learn the prior distribution of the data, then at test time, some conditioning mechanism is combined with the learned prior to solve very challenging imaging tasks \citep{avrahami2022blended, avrahami2022blendedLatent, chung2022mr}.
Note that our novel adaptive diffusion ingredient 
can be incorporated with any conditional sampling scheme that is based on diffusion models.

In \citep{whang2022deblurring,SR3} the problems of deblurring and super-resolution were considered. Specifically, a diffusion model is trained to perform the task. In this way, the model learns to carry out the deblurring or super-resolution task directly. Notice that these models are trained for one specific task and cannot be used for the other as is. 

The closest works to us are \citep{Giannone2022Few,Sheynin2022KNN,kawar2022imagic}. These recent works consider the task of image editing and perform an adaptation of the used diffusion model with the provided input and external data. Yet, notice that neither of these works consider the task of image reconstruction as we do here or apply our proposed sampling scheme for this task.

\paragraph{Image-Adaptive Reconstruction} Adaptation of pretrained deep models, which serve as priors in inverse problems, to the unknown true $\x$ through its observations at hand was proposed in \citep{IAGAN, tirer2019super}.
These works improve the reconstruction performance by fine-tuning the parameters of pretrained deep denoisers \citep{tirer2019super} and GANs \citep{IAGAN} via the observed image $\y$ instead of keeping them fixed during inference time.
The image-adaptive GAN (IAGAN) approach \citep{IAGAN} has led to many follow up works with different applications, e.g., \citep{bhadra2020medical,pan2021exploiting,roich2022pivotal,nitzan2022mystyle}. Recently, it has been shown that one may even fine-tune a masked-autoencoder to the input data at test-time for improving the adaptivity of classification neural networks to new domains \citep{gandelsman2022test}.

This paper considers test-time adaptation of diffusion models for inverse problems, where little attention was given to such type of adaptation. 
Furthermore, while existing works fine-tune the deep priors directly using $\y$, we propose an improved strategy where the tuning is based on $K$ external images similar to $\y$ that are automatically retrieved from an external dataset.

\section{Method}

We turn to present our proposed approach. We start with a brief description of  our proposed strategy for modifying the sampling scheme of diffusion models for the image reconstruction tasks. Then, we present our suggested adaptation scheme.

\subsection{Diffusion based Image Reconstruction} \label{sec:method_guided_diff}

We turn to extend the guidance method of \citep{guidedDiff} to image reconstruction.
First, we generalize their framework to inverse problems in the form of \eqref{eq:observ_model}. Namely, given the observed image $\y$, we modify the guided reverse diffusion process to generate possible reconstructions of $\x$ that are associated with $\y$ rather than arbitrary samples of a certain class. 
Similar to \citep{guidedDiff}, the guiding direction at iteration $t$ should follow (the gradient of) the likelihood function $p_{\y|\x_t}$.

The key difference between our framework and \citep{guidedDiff} is that we need to base our method on the specific degraded image $\y$ rather than on 
a classifier that has been trained for each level of noise of $\{ \x_t \}$.
However, only the likelihood function $p_{\y|\x_0}$ is known, 
i.e., of the clean image $\x_0$ that is 
available only at the end of the procedure, and not for every $1 \leq t \leq T$. To overcome this issue, we propose a surrogate for the intermediate likelihood functions $p_{\y|\x_t}$.
Our relaxation resembles the one in a recent concurrent work \citep{chung2022improving}. Yet, their sampling scheme is significantly different and has no adaptation ingredient.

Formally, we are interested in sampling from the posterior
\begin{equation}
    p_\theta(\x_{t} | \x_{t+1}, \y) \propto p_\theta(\x_{t}|\x_{t+1}) p_{\y|\x_t}(\y|\x_{t}),
    \label{eq:posterior}
\end{equation}
where $p_{\y|\x_t}(\cdot|\x_{t})$ is the distribution of $\y$ conditioned on $\x_t$, and $p_\theta(\x_{t}|\x_{t+1}) = \mathcal{N}(\bmu_{\theta}(\x_{t+1};t+1)), \bSigma_{\theta}(\x_{t+1}; t+1))$ is the learned diffusion prior.
For brevity, we omit the arguments of $\bmu_{\theta}$ and $\bSigma_{\theta}$ in the rest of this subsection. Under the assumption that the likelihood $\log p_{\y|\x_t}(\y|\cdot)$ has low curvature compared to $\bSigma_\theta^{-1}$ \citep{guidedDiff}, the following Taylor expansion is valid
\begin{align}
    \log p_{\y|\x_t}(\y|\x_t)  &\approx \log p_{\y|\x_t}(\y|\x_t) |_{\x_t=\bmu_\theta} + (\x_t - \bmu_\theta)^\top \ \nabla_{\x_t} \log p_{\y|\x_t}(\y|\x_t) |_{\x_t=\bmu_\theta} \nonumber \\
    &= (\x_t - \bmu_\theta)^\top \g + C_1,
    \label{eq:likelihood_apprx}
\end{align}
where $\g=\nabla_{\x_t} \log p_{\y|\x_t}(\y|\x_t) |_{\x_t=\bmu_\theta}$, and $C_1$ is a constant that does not depend on $\x_t$.
Then, similar to the computation in \citep{guidedDiff}, we can use \eqref{eq:likelihood_apprx} to express the posterior in \eqref{eq:posterior}, i.e.,
\begin{align}
    \log(p_{\theta}(\x_t|\x_{t+1})p_{\y|\x_t}(\y|\x_t)) &\approx C_2 + \log p(\z),
    \label{eq:posterior_apprx}
\end{align}  
where $\z\sim \mathcal{N}(\bmu_\theta + \bSigma_\theta \g, \bSigma_\theta)$, and $C_2$ is some constant that does not depend on $\x_t$.
Therefore, for conditioning the diffusion reverse process on $\y$, one needs to evaluate the derivative $\g$ from a (different) log-likelihood function $\log p_{\y|\x_t}(\y|\cdot)$ at each iteration $t$. 

Observe that we know the exact log-likelihood function for $t=0$. 
Since the noise $\e$ in \eqref{eq:observ_model} is white Gaussian with variance $\sigma^2$, we therefore 
have following distribution 
\begin{align}
\label{eq:pyx_prop_gaussian}
    p_{\y|\x}(\y|\x) = \mathcal{N}(\A\x, \sigma^2 \I_m) 
             \propto e^{-\frac{1}{2\sigma^2}\|\y - \A\x\|_2^2}.
\end{align}
In the denoising diffusion setup, $\y$ is related to $\x_0$ using the observation model in \eqref{eq:observ_model}. Therefore,
\begin{equation}
\label{eq:logLike_x0}
    \log p_{\y|\x_0}(\y|\x_0) \propto -\|\A\x_0 - \y\|_2^2.
\end{equation}

Motivated by the expression above, we use the following approximation
\begin{align}
\label{eq:likelihood_approx}
    \log p_{\y|\x_t}(\y|\x_t) &\approx \log p_{\y|\x_0}(\y|\hat{\x}_0(\x_t)), 
\end{align}
where 
\begin{align}
    \hat{\x}_0(\x_t) := \left ( \x_t - \sqrt{1-\bar{\alpha}_t}\bepsilon_{\theta}(\x_t, t) \right )/ \sqrt{\bar{\alpha}_t}
\end{align}
is an estimation of $\x_0$ from $\x_t$, which is based on the (stochastic) relation of $\x_t$ and $\x_0$ in \eqref{eq:xt_given_x0} and the random noise $\bepsilon$ is replaced by its estimation $\bepsilon_{\theta}(\x_t, t)$.

From \eqref{eq:pyx_prop_gaussian} and \eqref{eq:likelihood_approx} it follows that $\g$ in \eqref{eq:likelihood_apprx} can be approximated at each iteration $t$ by evaluating (e.g., via automatic-differentiation)
\begin{align}
    \g \approx -\nabla_{\x_t} \|\A\hat{\x}_0(\x_t) - \y\|_2^2 |_{\x_t=\bmu_\theta}.
    \label{eq:log_likelihood_approx}
\end{align}

Note that existing methods \citep{chung2022improving, kawar2022denoising, song2021solving} either use a term that resembles \eqref{eq:log_likelihood_approx} with the naive approximation $\hat{\x}_0(\x_t) = \x_t$ \citep{kawar2022denoising, song2021solving}, or significantly modify \eqref{eq:log_likelihood_approx} before computing it via the automatic derivation framework \citep{chung2022improving} (we observed that trying to compute exactly \eqref{eq:log_likelihood_approx} is unstable due to numerical issues). For example, in the official implementation of \citep{chung2022improving}, which uses automatic derivation, the squaring of the norm in \eqref{eq:log_likelihood_approx} is dropped even though this is not stated in their paper (otherwise, the reconstruction suffers from significant artifacts). In our case, we use the following relaxation to overcome the stability issue of using \eqref{eq:log_likelihood_approx} directly. For a pretrained denoiser predicting $\bepsilon_\theta$ from $\x_t$ and $0 < t \leq T$ we have
\begin{align}
    \|\A\hat{\x}_0(\x_t) - \y\|_2^2 
    &= \| \A(\x_t - \sqrt{1-\bar{\alpha}_t}\bepsilon_\theta)/\sqrt{\bar{\alpha}_t} - \y \|_2^2 
    \propto \| \A\x_t - \sqrt{1-\bar{\alpha}_t}\A\bepsilon_\theta - \sqrt{\bar{\alpha}_t} \y \|_2^2 \nonumber \\
    &= \| \A\x_t - \sqrt{\bar{\alpha}_t} \y - \sqrt{1-\bar{\alpha}_t}\A\bepsilon_\theta \|_2^2 
    = \| \A\x_t - \y_t \|_2^2,
    \label{eq:likelihood_apprx_diff}
\end{align}
where $\y_t := \sqrt{\bar{\alpha}_t} \y + \sqrt{1-\bar{\alpha}_t}\A\bepsilon_\theta$.
From our experiments we found that omitting the recursive dependency of $\bepsilon_\theta$ on $\x_t$ is sufficient.
Consequently, we propose to replace the expression for $\g$ in \eqref{eq:log_likelihood_approx} with a surrogate obtained by evaluating the derivative of \eqref{eq:likelihood_apprx_diff} w.r.t. $\x_t$, formally
\begin{align}
    \g &= -(2\A^T(\A\x_t - \y_t) - 2\cancel{(\nabla_{\x_t}\y_t)^T}(\A\x_t - \y_t))|_{\x_t = \bmu_\theta}
    \approx -2\A^T(\A\x_t - \y_t)|_{\x_t = \bmu_\theta},\label{eq:likelihood_deriv}
\end{align}
which can be used for sampling the posterior distribution as detailed in Algorithm \ref{alg:guided_diff}. 

\subsection{Adaptive Diffusion}\label{sec:method_IA}

Having defined the guided inverse diffusion flow for image reconstruction, we turn to discuss how one may adapt a given diffusion model to a given degraded image $\y$ as defined in \eqref{eq:observ_model}. Assume we have a pretrained diffusion model $(\bepsilon_\theta(\cdot), \bSigma_\theta(\cdot))$, then the adaptation scheme is defined by the following minimization problem
\begin{align}
    \hat{\theta} = \arg \min_\theta \sum_{t = 1}^T \ell_\text{simple}(\y, \bepsilon_\theta, t) 
    \label{eq:IA0_loss}
\end{align}
with $\ell_\text{simple}$ defined in \eqref{eq:diff_simple_loss}, 
which can be solved using stochastic gradient descent, where at each iteration the gradient step is performed on a single term of the sum above, for $0<t\leq T$ chosen randomly. Although the original work \citep{guidedDiff} trains the network to predict the posterior variance $\Sigma_{\theta}$, in our case, we did not see any benefit of including it in the adaptation loss.

Adapting the denoising network to the measurement image $\y$, allows it to learn cross-scale features recurring in the image, which is a well studied property of natural images \citep{ulyanov2018deep, mataev2019deepred, shaham2019singan, michaeli2014blind}. Such an approach has been proven to be very helpful in reconstruction-based algorithms \citep{IAGAN, tirer2019super}. 
However, in some cases where the image does not satisfy the assumption of recurring patterns across scales, this approach can lose some of the sharpness captured in training. Therefore, in this work we extend the approach of measurement based fine-tuning, by developing an
algorithm for retrieving $K$ images similar to $\x$ from a large dataset of diverse images, using off-the-shelf embedding distance,
e.g., CLIP \citep{CLIP}, BLIP \citep{li2022blip}, or CyCLIP \citep{goel2022cyclip}). 

We then use the $K$-Nearest Neighbor ($K$-NN) images $\{\z_{k}\}_{k=1}^K$ to fine-tune the diffusion model, 
which adapts the denoising network to the context of $\y$. 
Specifically, we modify the denoiser parameters $\theta$ in a similar fashion to \eqref{eq:IA0_loss}, but with $\{\z_{k}\}_{k=1}^K$ rather than $\y$. We stochastically solve the following minimization problem
\begin{align}
    \hat{\theta} = \arg \min_\theta \sum_{k=1}^K\sum_{t = 1}^T \ell_\text{simple}(\z_k, \bepsilon_\theta, t) 
    \label{eq:IA_loss}
\end{align}
We refer to this K-NN based adaptation technique as ADIR (Adaptive Diffusion for Image Reconstruction), which is described schematically in Figure \ref{fig:ADIR}.

\section{Experiments}

We evaluate our method on two state-of-the-art diffusion models, Guided Diffusion (GD) \citep{guidedDiff} and Stable Diffusion (SD) \citep{rombach2022high}, showing results for super-resolution, colorization and deblurring. In addition, we show how adaptive diffusion can be used for the task of text-based editing using stable diffusion.

GD \citep{guidedDiff} provides several models with a conditioning mechanism built-in to the denoiser. However, in our case, we perform the conditioning using the $\log$-likelihood term. Therefore, we used the unconditional model that was trained on ImageNet \citep{russakovsky2015imagenet} and produces images of size $256\times 256$. 
In addition to GD, we demonstrate the improvement that can be achieved using stable diffusion \citep{rombach2022high}, where we use its publicly available super-resolution and text-based editing models.

In all cases, we adapt the diffusion models in the image adaptive scheme presented in section \ref{sec:method_IA}, using the Google Open Dataset \citep{OpenImages} as the external dataset 
, from which we retrieve $K$ images, where $K=20$ for GD and $K=50$ for SD (several examples of retrieved images are shown Figure \ref{fig:supp_SRx8_NN}). 

Since the Nearest Neighbor (NN) search is performed in the embedding space, we can efficiently retrieve the $K$ images from the $1.7M$ images using a K-D Tree structure. This significantly accelerates the retrieval procedure, as can be seen in Table \ref{table:KNN_ablation}. We compare the reconstruction performance and the runtimes when using random NN images, MSE-based NN, and using our approach. 

For optimizing the network parameters we use LoRA \citep{hu2021lora} with rank $r=16$ and scaling $\alpha=8$ for all the convolution layers, which is then optimized using
Adam \citep{kingma2014adam}.
The specific implementation configurations are detailed in Table \ref{table:appndx_configs}. We run all of our experiments on a NVIDIA RTX A6000 48GB card, which allows us to fine-tune the models by randomly sampling a batch of 6 images from $\{\z_{k}\}_{k=1}^K$, where in each iteration we use the same $0 < t \leq T$ for images in the batch.

\subsection{Super Resolution}
In the Super-Resolution (SR) task one would like to reconstruct a high resolution image $\x$ from its low resolution image $\y$, where in this case $\A$ represents an anti-aliasing filter followed by sub-sampling with stride $\gamma$, which we refer to as the scaling factor. In our setup we use a bicubic anti-aliasing filter and assume $\e=0$, similarly to many SR works.

Here we apply our approach on two different diffusion based SR methods, Stable Diffusion \citep{rombach2022high}, and section \ref{sec:method_guided_diff} approach combined with the unconditional diffusion model from \citep{guidedDiff}. In Stable Diffusion, the low-resolution image $\y$ is upscaled from $256\times 256$ to $1024\times 1024$, while in Guided Diffusion we use the unconditional model trained on $256\times 256$ images. When adapting Stable diffusion, we downsample random crops of the $K$-NN images using $\A$, which we encode using the VAE and plug into the network conditioning mechanism. We fine-tune both models using random crops of the $K$-NN images, to which we then add noise using the scheduler provided by each model. 

The perception preference of generative models-based image reconstruction has been seen in many works \citep{IAGAN, bora2017compressed, blau2018perception}. Therefore, we chose a perception-based measure to evaluate the performance of our method. Specifically, we use the state-of-the-art AVA-MUSIQ and KonIQ-MUSIQ perceptual quality assessment measures \citep{ke2021musiq}. We report our results using the two measures averaged on the first 50  validation images of the DIV2K \citep{div2k} dataset. As can be seen in Table  \ref{table:SR}, our method significantly outperforms both Stable Diffusion, GD-based reconstruction approaches, and several other state-of-the-art technique. We also compare ADIR to adaptation using random images from the dataset, as well as MSE-based retrieval, and report the results in Table \ref{table:KNN_ablation} and in Figure \ref{fig:Faces}, where the obvious advantage of ADIR can be seen clearly, emphasizing the effectiveness of our adaptation approach.

\subsection{Deblurring}
In deblurring, $\y$ is obtained by applying a blur filter (uniform blur of size $5\times 5$ in our case) on $\x$, followed by adding measurement noise $\e\sim \mathcal{N}(0, \sigma^2I_n)$, where in our setting $\sigma = 10$. 
We apply our proposed approach in Section \ref{sec:method_guided_diff} for the Guided Diffusion unconditional model \citep{guidedDiff} to solve the task. 
As a baseline, we use the unconditional diffusion model provided by GD \citep{guidedDiff}, which was trained on $256\times 256$ size images. Yet, in our tests, we solve the deblurring task on images of sizes $256\times 256$ and $512\times 512$, which emphasizes the remarkable benefit of the adaptation, as it allows the model to generalize to resolutions not seen during training.
Similar to SR, in Table \ref{table:GD_deblur} we report the KonIQ-MUSIQ and AVA-MUSIQ \citep{ke2021musiq} measures, averaged on the first 50 DIV2K validation images \citep{div2k}.
Visual comparisons are also available in Figure \ref{fig:GD_Deblur}, where a significant improvement can be seen in both robustness to noise and reconstructing details. We also compare ADIR to the scenario where we adapt the denoiser on random images from the dataset, as well as MSE-based retrieval; as can be seen in Table \ref{table:KNN_ablation} and Figure \ref{fig:Faces}.

\subsection{Colorization}
In colorization, $\y$ is obtained by averaging the colors of $\x$ using RGB2Gray transform. 
Similar to deblurring, we apply our proposed approach in Section \ref{sec:method_guided_diff} to solve the task. In this case, $\A$ can be implemented by averaging the color dimension of $\x$, while $\A^T$ can simply be viewed as a replication of the color dimension. We use the unconditional diffusion model provided by GD \citep{guidedDiff} as a baseline for coloring $256\times256$ images. Visual comparison of the results can be seen in Figure \ref{fig:GD_color}. We report the average MUSIQ \citep{ke2021musiq} perceptual measure for this case, as shown in Table \ref{table:GD_color}. Note that we do not report LPIPS as there are many colorization solutions and therefore the reconstructed image may differ a lot from the ground truth. Thus, we focus on non-reference based perceptual measures for the colorization task.

\subsection{Text-Guided Editing}
Text-guided image editing is the task of completing a masked region of  $\x$ according to a prompt provided by the user. In this case, the diffusion model needs to predict objects and textures correspondent to the provided prompt, therefore we chose to adapt the network on $\{\z_{k}\}_{k=1}^K$ retrieved using the text encoder.
For evaluating our method for this application, we use the inpainting variant of Stable Diffusion \citep{rombach2022high}. 
When adapting the network, we follow the training scheme of Stable Diffusion, where we use random masks and the classifier-free conditioning \citep{ho2022classifier}. Notice that we cannot compare to  \citep{Giannone2022Few,Sheynin2022KNN,kawar2022imagic} as there is no code available for them. For some of them, we do not even have access to the diffusion model that they adapt \citep{Saharia2022Photorealistic}. Note though that our goal is not to show state-of-the-art editing results but rather to show here the potential contribution of ADIR to text-guided editing. As it is a general framework, it may be used also with other existing editing techniques in order to improve them. 
Figure~\ref{fig:Editing_GLIDE} presents the editing results and compares them to both stable diffusion and GLIDE. 
The images of GLIDE are taken from the paper. 
Since Stable Diffusion was trained using a lossy latent representation with smaller dimensionality than the data, it is clear that GLIDE can achieve better results. 
However, because our method adapts the network to a specific scenario, it enables the model to produce cleaner and more accurate generations, as can be seen in Figure \ref{fig:Editing_GLIDE}. 

\paragraph{Limitation.} One limitation of our approach is that as is the case with all diffusion models, there is randomness in the generation process of the results. Therefore, the quality of the output may depend on the random seed being used.  
For a fair comparison, we used the same seed both for ADIR and the baseline. 
In the appendix, we provide more examples with different random seeds.
We still find that when 
we compare our approach and the baseline with the same seed, 
we consistently get an improvement. 
Another limitation of ADIR is that it works sequentially, i.e. we first look for $K$-NN images and then fine-tune the denoiser network on these images, therefore, an additional run-time is added to the standard diffusion flow. Also, in this work, we assume that the observation operator $\A$ is known (non-blind setting), while in many real-world applications, it is usually inaccessible. As a result, one needs to run the guidance scheme (section \ref{sec:method_guided_diff}) with an estimated version of $\A$, which is suboptimal.
Additionally, for optimal performance, one should use a relatively diverse dataset to retrieve images that match the degraded image context. Otherwise, the adaptation can lead to negligible advantage. We leave exploring these questions to a future research.

\section{Conclusion}
We have presented the Adaptive Diffusion Image Reconstruction (ADIR) method, in which we improve the reconstruction results in several imaging tasks using off-the-shelf diffusion models. We have demonstrated how our adaptation can significantly improve existing state-of-the-art methods, e.g. Stable Diffusion for super resolution, where the exploitation of external data with the same context as $\y$, combined with our adaptation scheme leads to a significant improvement. Specifically, the produced images are sharper and have more details than the original ground truth image.
Importantly, note that our novel adaptive diffusion ingredient 
can be incorporated into any conditional sampling scheme that is based on diffusion models, beyond those that are examined in this paper. One such possible direction is integrating our method with advanced diffusion models-based editing techniques \citep{meng2022sdedit,Kim_2022_CVPR,mokady2022null,bar2023multidiffusion,molad2023dreamix,wei2023elite,lhhuang2023composer,qi2023fatezero,liu2023videop2p}.


\bibliography{egbib}

\clearpage

\onecolumn
\appendix 
\section*{Additional Results}

\begin{figure}[!hb]
\captionsetup[subfigure]{labelformat=empty}
    \centering
    \begin{subfigure}[b]{0.19\linewidth}
        \centering
        \begin{overpic}[width=\linewidth]{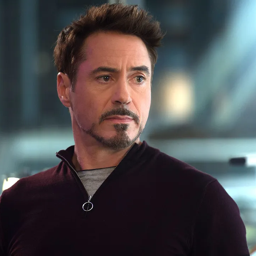}
            \put(0.5\linewidth,0){\includegraphics[width=0.5\linewidth, trim={80 100 100 100}, clip]{Faces/deblur_GT.png}}  
        \end{overpic}
        \caption{\footnotesize Ground truth}
    \end{subfigure}
    \begin{subfigure}[b]{0.19\linewidth}
        \centering
        \begin{overpic}[width=\linewidth]{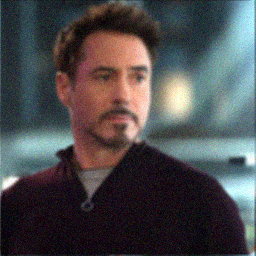}
            \put(0.5\linewidth,0){\includegraphics[width=0.5\linewidth, trim={80 100 100 100}, clip]{Faces/deblur_box_y.png}}  
        \end{overpic}
        \caption{\footnotesize Blurry}
    \end{subfigure}
     \begin{subfigure}[b]{0.19\linewidth}
        \centering
        \begin{overpic}[width=\linewidth]{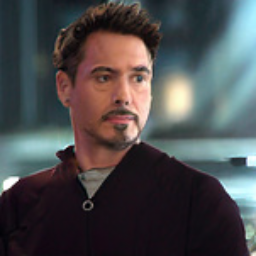}
            \put(0.5\linewidth,0){\includegraphics[width=0.5\linewidth, trim={80 100 100 100}, clip]{Faces/deblur_box_GD256_s5.png}}  
        \end{overpic}
        \caption{\footnotesize Guided Diffusion}
    \end{subfigure}
    \begin{subfigure}[b]{0.19\linewidth}
        \centering
        \begin{overpic}[width=\linewidth]{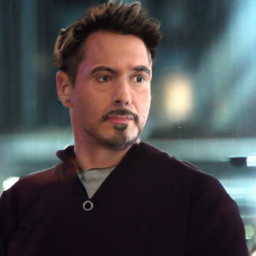}
            \put(0.5\linewidth,0){\includegraphics[width=0.5\linewidth, trim={80 100 100 100}, clip]{Faces/deblur_box_s5.0_LS_IA1000_random_KNN.png}}  
        \end{overpic}
        \caption{\footnotesize ADIR Random $K$-NN}
    \end{subfigure}
    \begin{subfigure}[b]{0.19\linewidth}
        \centering
        \begin{overpic}[width=\linewidth]{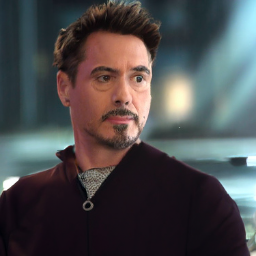}
            \put(0.5\linewidth,0){\includegraphics[width=0.5\linewidth, trim={80 100 100 100}, clip]{Faces/deblur_box_s5.0_LS_IA1000test.png}}  
        \end{overpic}
        \caption{\footnotesize ADIR (ours)}
    \end{subfigure}\\
    \begin{subfigure}[b]{0.19\linewidth}
        \centering
        \begin{overpic}[width=\linewidth]{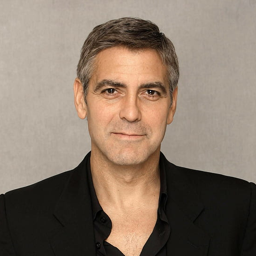}
            \put(0.5\linewidth,0){\includegraphics[width=0.5\linewidth, trim={140 260 190 30}, clip]{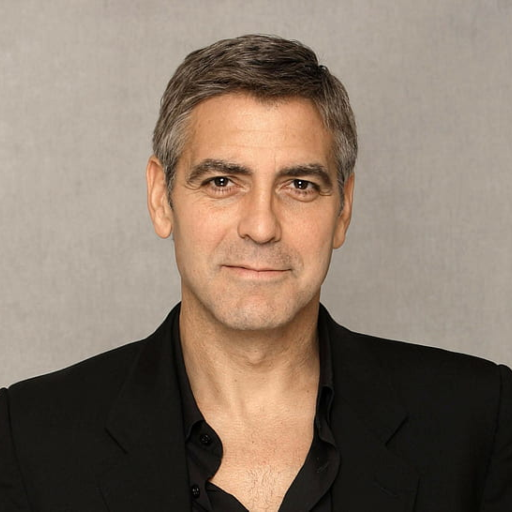}}  
        \end{overpic}
        \caption{\footnotesize Ground truth}
    \end{subfigure}
    \begin{subfigure}[b]{0.19\linewidth}
        \centering
        \begin{overpic}[width=\linewidth]{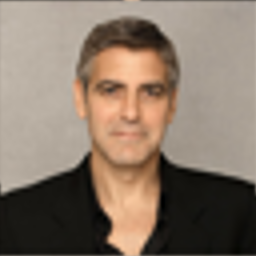}
            \put(0.5\linewidth,0){\includegraphics[width=0.5\linewidth, trim={140 260 190 30}, clip]{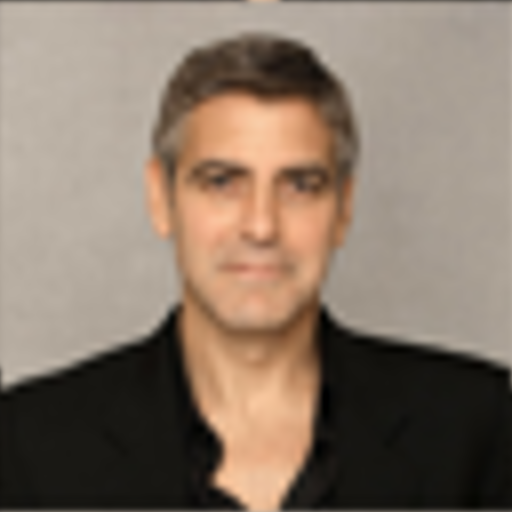}}  
        \end{overpic}
        \caption{\footnotesize Bicubic x4}
    \end{subfigure}
     \begin{subfigure}[b]{0.19\linewidth}
        \centering
        \begin{overpic}[width=\linewidth]{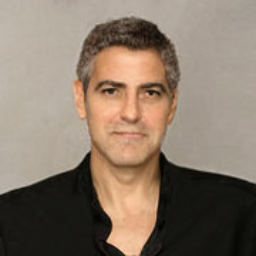}
            \put(0.5\linewidth,0){\includegraphics[width=0.5\linewidth, trim={140 260 190 30}, clip]{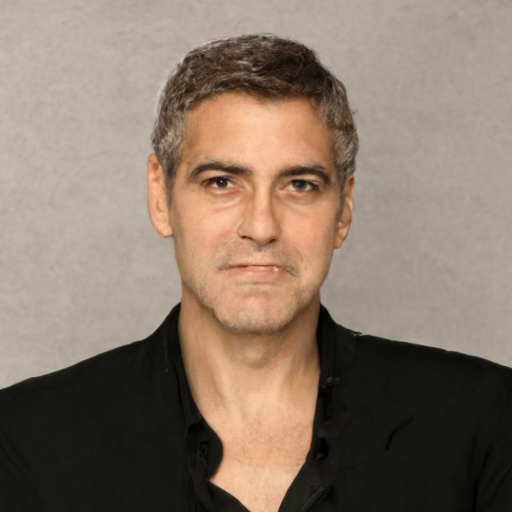}}  
        \end{overpic}
        \caption{\footnotesize Guided Diffusion}   
    \end{subfigure}
    \begin{subfigure}[b]{0.19\linewidth}
        \centering
        \begin{overpic}[width=\linewidth]{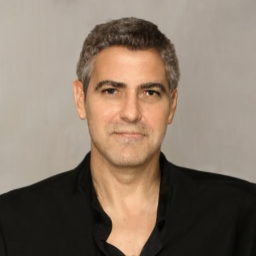}
            \put(0.5\linewidth,0){\includegraphics[width=0.5\linewidth, trim={140 260 190 30}, clip]{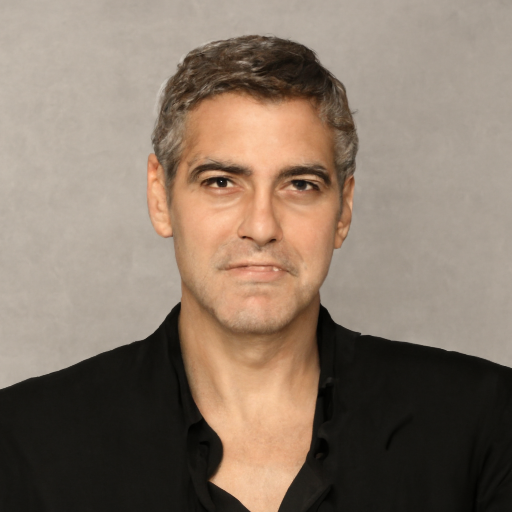}}  
        \end{overpic}
        \caption{\footnotesize ADIR Random $K$-NN}
    \end{subfigure}
    \begin{subfigure}[b]{0.19\linewidth}
        \centering
        \begin{overpic}[width=\linewidth]{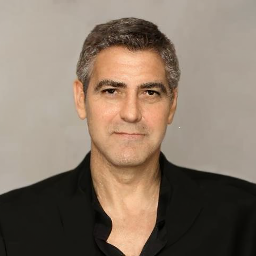}
            \put(0.5\linewidth,0){\includegraphics[width=0.5\linewidth, trim={140 260 190 30}, clip]{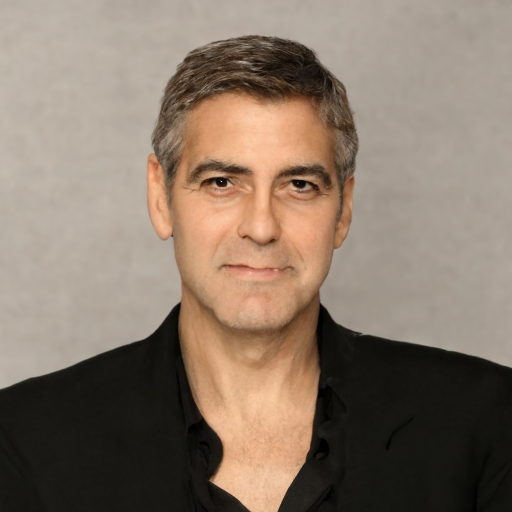}}  
        \end{overpic}
        \caption{\footnotesize ADIR (ours)}
    \end{subfigure}

    \caption{Ablation study on the benefit of ADIR compared to adapting the denoiser on random images for deblurring and super-resolution of celebrity images. }
    \label{fig:Faces}
\end{figure}

\begin{figure*}
\captionsetup[subfigure]{labelformat=empty}
    \centering
    \begin{subfigure}[b]{0.24\linewidth}
        \centering
        \includegraphics[width=\linewidth, height=101pt]{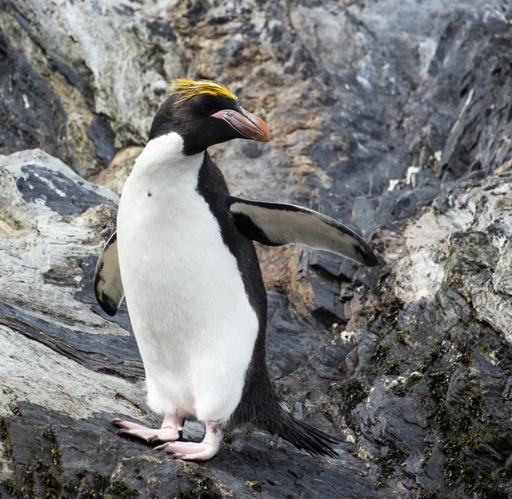}
   
    \end{subfigure}
     \begin{subfigure}[b]{0.24\linewidth}
        \centering
        \includegraphics[width=\linewidth, height=101pt]{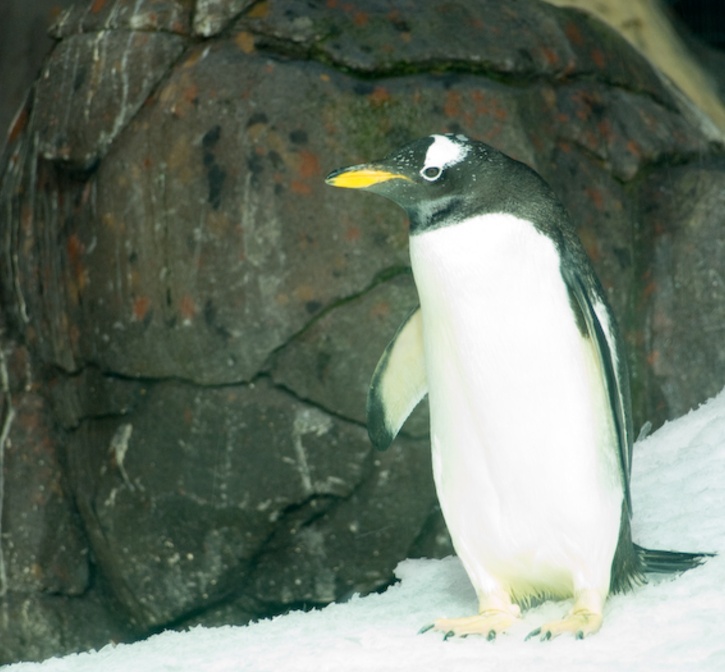}
   
    \end{subfigure}
    \begin{subfigure}[b]{0.24\linewidth}
        \centering
        \includegraphics[width=\linewidth, height=101pt]{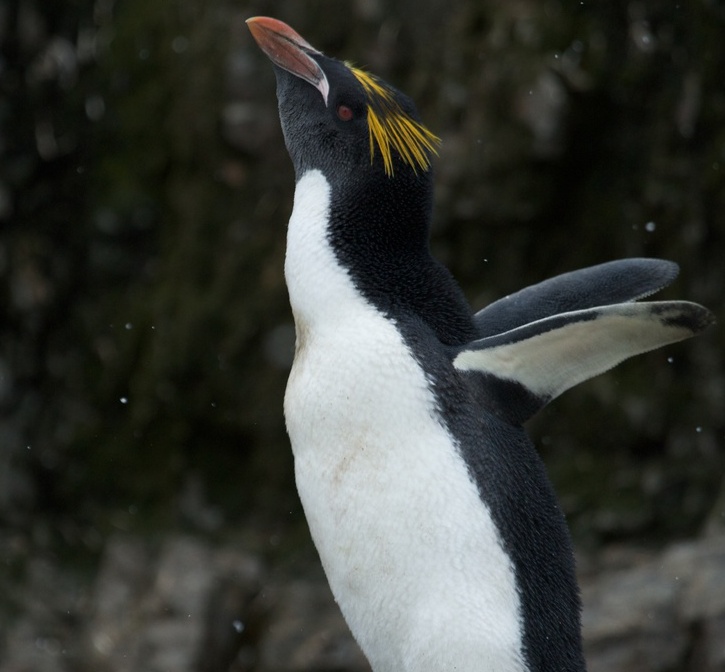}

    \end{subfigure}
    \begin{subfigure}[b]{0.24\linewidth}
        \centering
        \includegraphics[width=\linewidth, height=101pt]{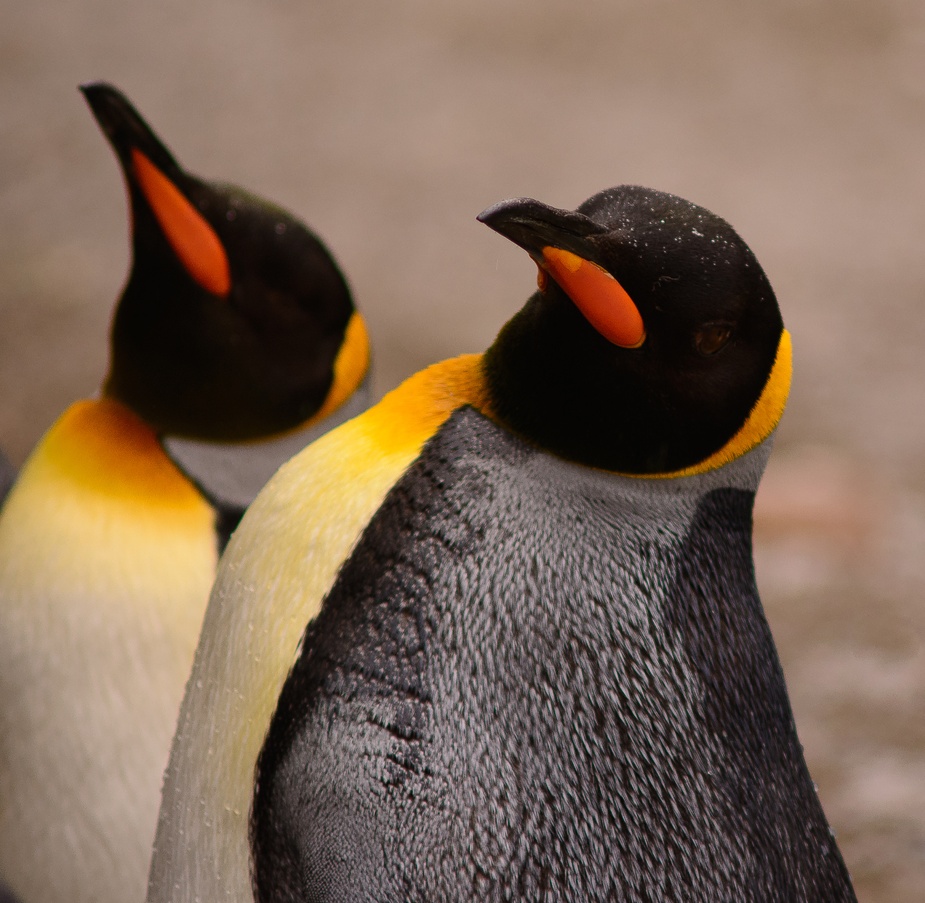}
    \end{subfigure}\\
    \begin{subfigure}[b]{0.24\linewidth}
        \centering
        \includegraphics[width=\linewidth, height=101pt]{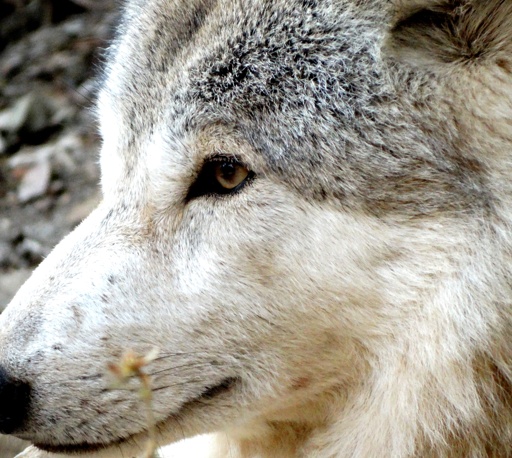}
   
    \end{subfigure}
     \begin{subfigure}[b]{0.24\linewidth}
        \centering
        \includegraphics[width=\linewidth, height=101pt]{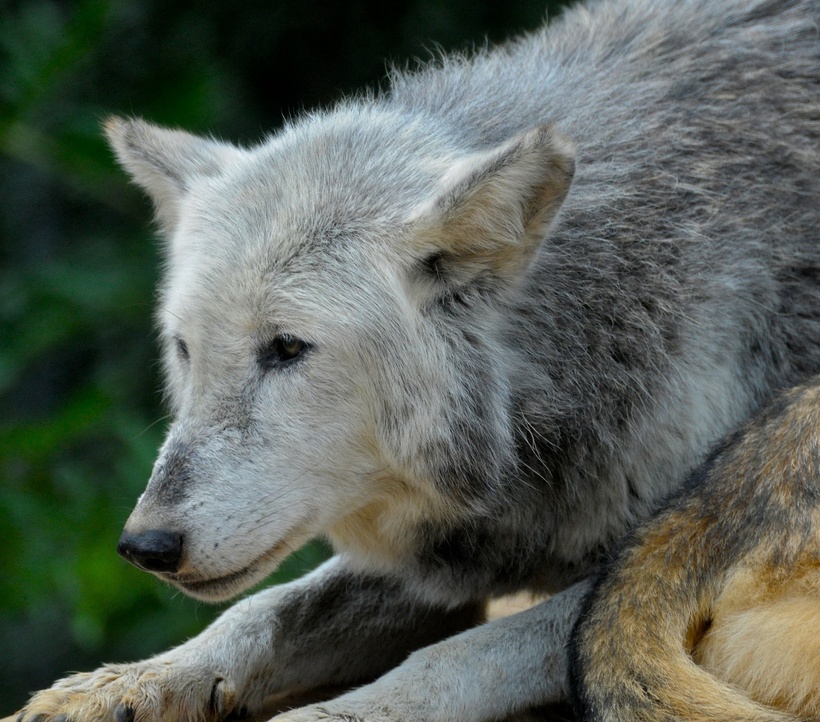}
   
    \end{subfigure}
    \begin{subfigure}[b]{0.24\linewidth}
        \centering
        \includegraphics[width=\linewidth, height=101pt]{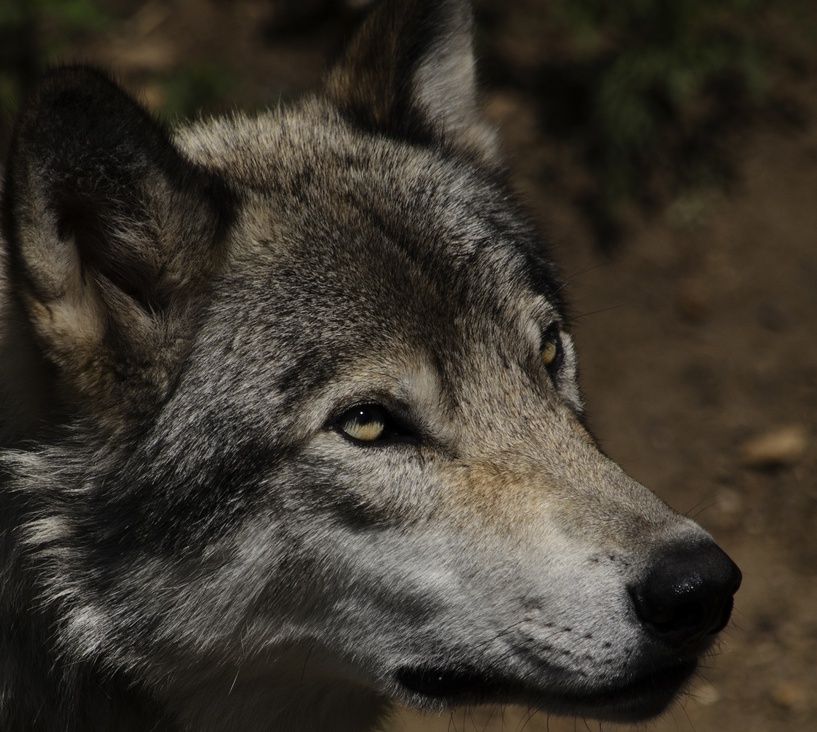}

    \end{subfigure}
    \begin{subfigure}[b]{0.24\linewidth}
        \centering
        \includegraphics[width=\linewidth, height=101pt]{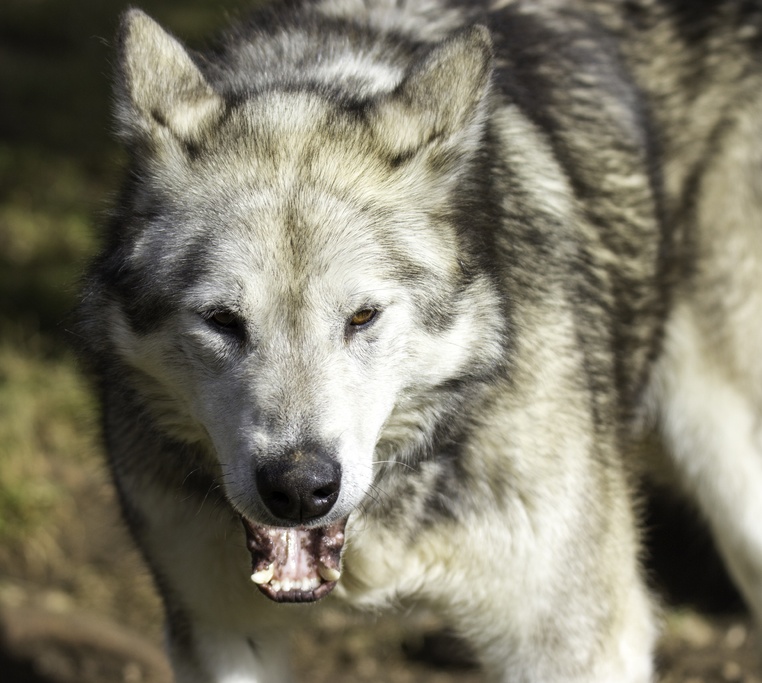}
    \end{subfigure}\\
    \begin{subfigure}[b]{0.24\linewidth}
        \centering
        \includegraphics[width=\linewidth, height=101pt]{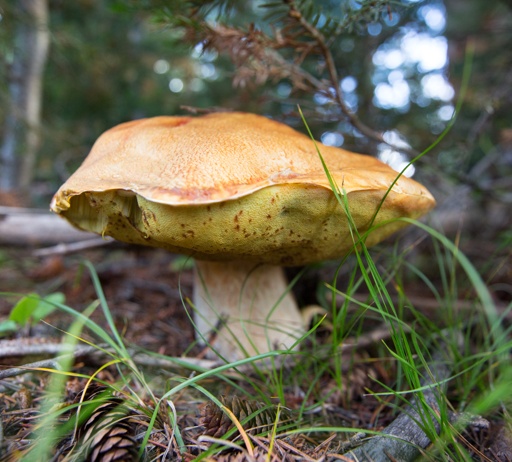}
   
    \end{subfigure}
     \begin{subfigure}[b]{0.24\linewidth}
        \centering
        \includegraphics[width=\linewidth, height=101pt]{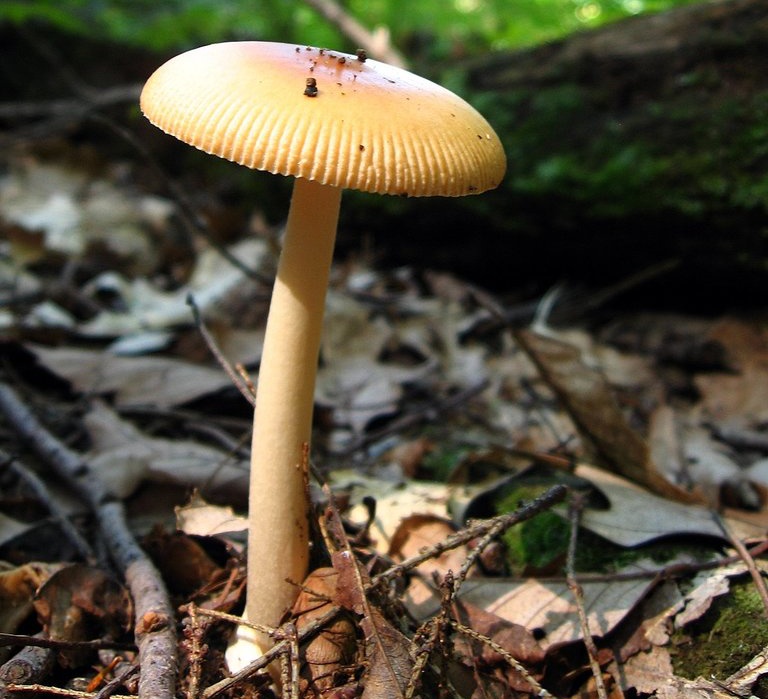}
   
    \end{subfigure}
    \begin{subfigure}[b]{0.24\linewidth}
        \centering
        \includegraphics[width=\linewidth, height=101pt]{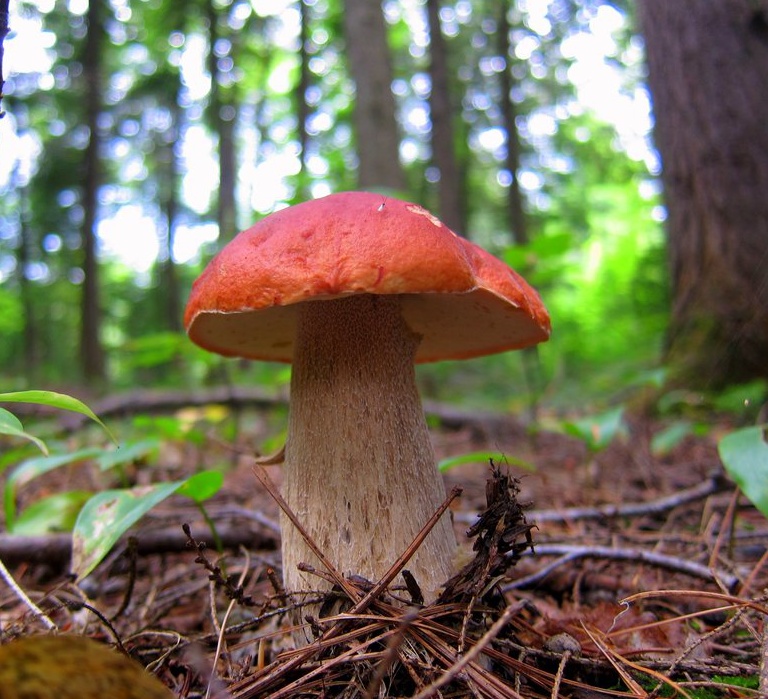}

    \end{subfigure}
    \begin{subfigure}[b]{0.24\linewidth}
        \centering
        \includegraphics[width=\linewidth, height=101pt]{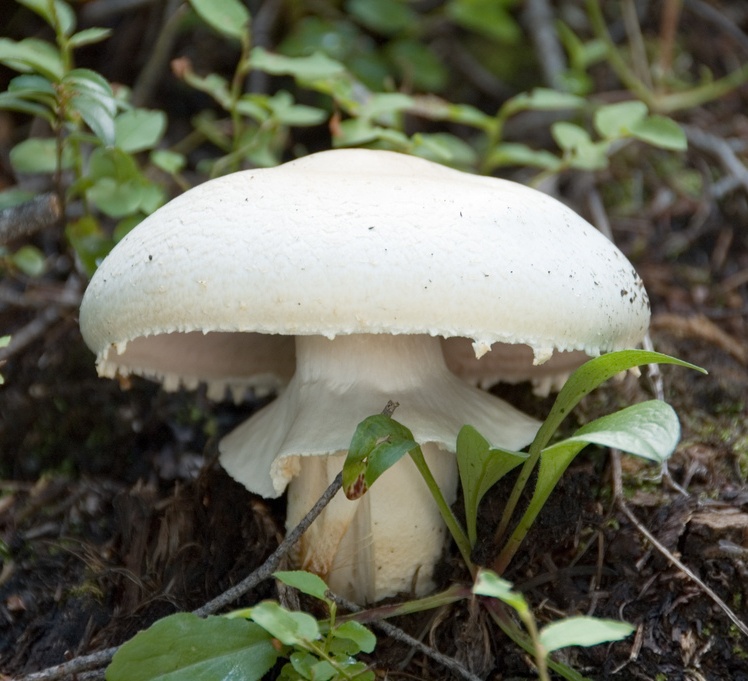}
    \end{subfigure}\\
    \begin{subfigure}[b]{0.24\linewidth}
        \centering
        \includegraphics[width=\linewidth, height=101pt]{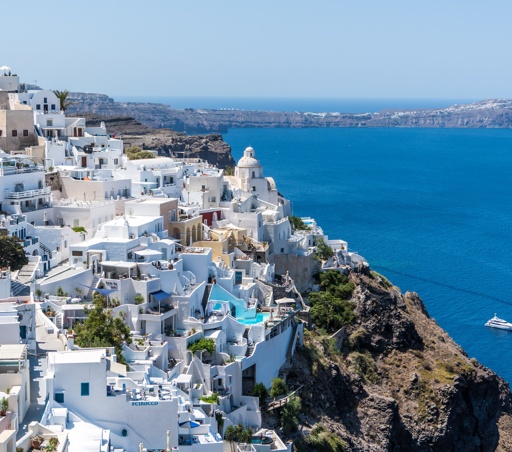}
        \subcaption{Ground Truth}
    \end{subfigure}
     \begin{subfigure}[b]{0.24\linewidth}
        \centering
        \includegraphics[width=\linewidth, height=101pt]{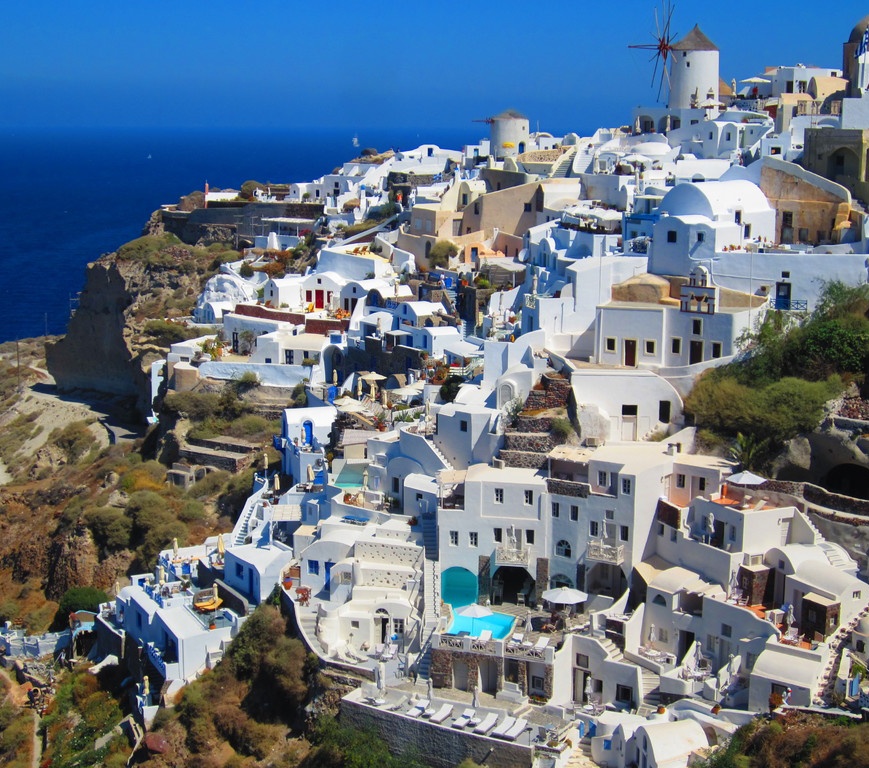}
        \subcaption{NN-1}
    \end{subfigure}
    \begin{subfigure}[b]{0.24\linewidth}
        \centering
        \includegraphics[width=\linewidth, height=101pt]{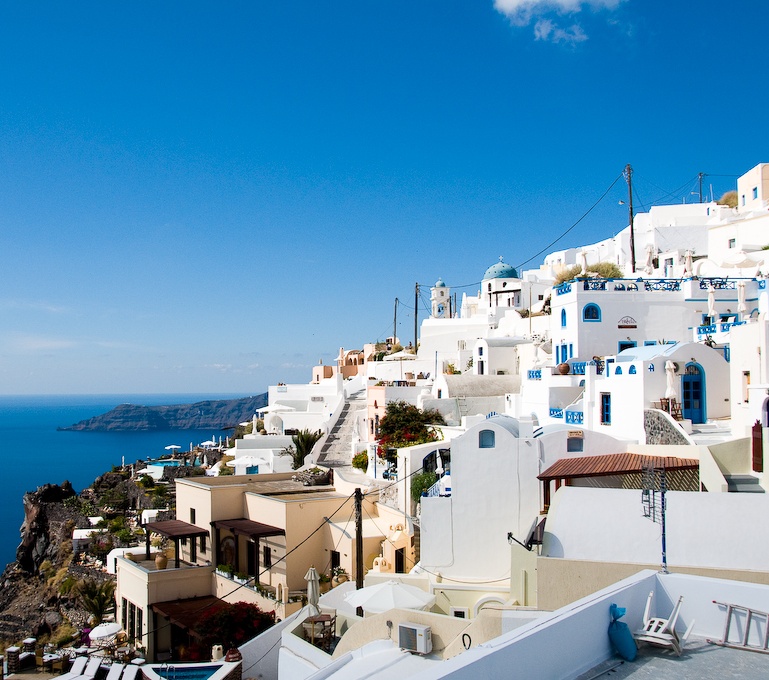}
        \subcaption{NN-2}
    \end{subfigure}
    \begin{subfigure}[b]{0.24\linewidth}
        \centering
        \includegraphics[width=\linewidth, height=101pt]{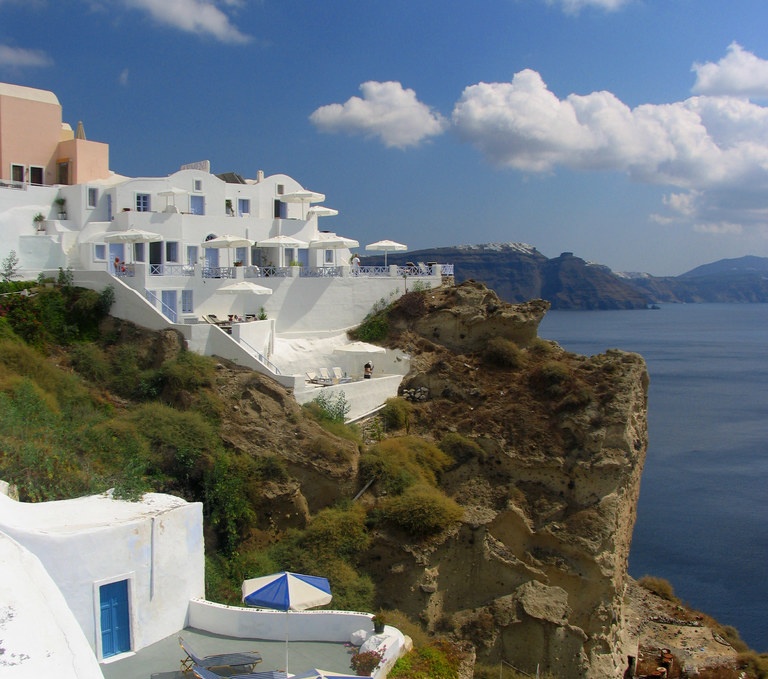}
        \subcaption{NN-3}
    \end{subfigure}
    \caption{Examples of images retrieved from Google Open Dataset \citep{OpenImages} using CLIP \citep{CLIP} for super resolution with scale factor of 8 ($64^2\rightarrow 512^2$).}
    \label{fig:supp_SRx8_NN}
\end{figure*}

\begin{figure*}
\captionsetup[subfigure]{labelformat=empty}
    \centering
    \begin{subfigure}[b]{0.16\linewidth}
        \centering
        \includegraphics[width=\linewidth]{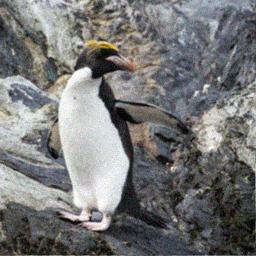}
    \end{subfigure}
     \begin{subfigure}[b]{0.16\linewidth}
        \centering
        \includegraphics[width=\linewidth]{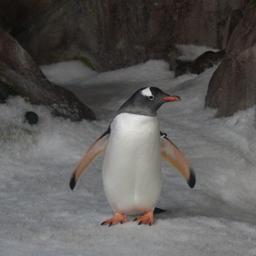}
    \end{subfigure}
    \begin{subfigure}[b]{0.16\linewidth}
        \centering
        \includegraphics[width=\linewidth]{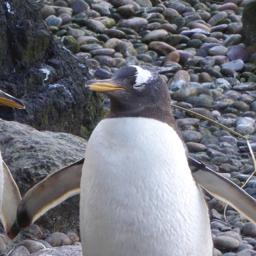}
    \end{subfigure}
    \begin{subfigure}[b]{0.16\linewidth}
        \centering
        \includegraphics[width=\linewidth]{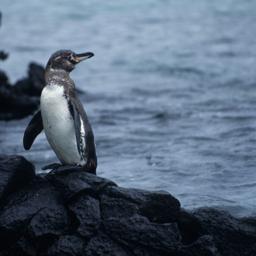}
    \end{subfigure}
    \begin{subfigure}[b]{0.16\linewidth}
        \centering
        \includegraphics[width=\linewidth]{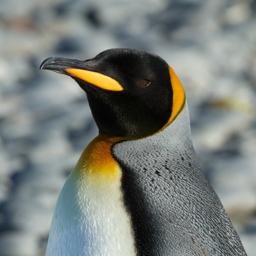}
    \end{subfigure}
    \begin{subfigure}[b]{0.16\linewidth}
        \centering
        \includegraphics[width=\linewidth]{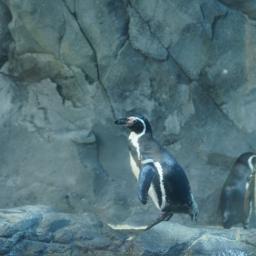}
    \end{subfigure}
    \\
    \begin{subfigure}[b]{0.16\linewidth}
        \centering
        \includegraphics[width=\linewidth]{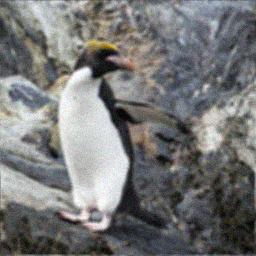}
    \end{subfigure}
     \begin{subfigure}[b]{0.16\linewidth}
        \centering
        \includegraphics[width=\linewidth]{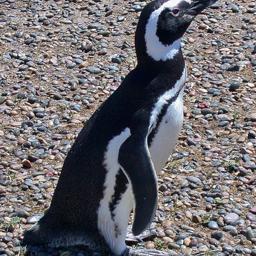}
    \end{subfigure}
    \begin{subfigure}[b]{0.16\linewidth}
        \centering
        \includegraphics[width=\linewidth]{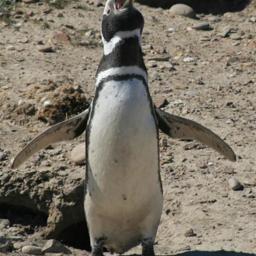}
    \end{subfigure}
    \begin{subfigure}[b]{0.16\linewidth}
        \centering
        \includegraphics[width=\linewidth]{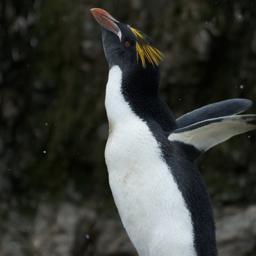}
    \end{subfigure}
    \begin{subfigure}[b]{0.16\linewidth}
        \centering
        \includegraphics[width=\linewidth]{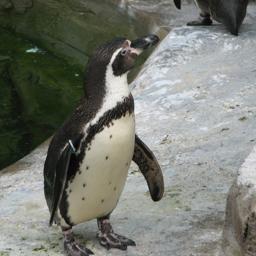}
    \end{subfigure}
    \begin{subfigure}[b]{0.16\linewidth}
        \centering
        \includegraphics[width=\linewidth]{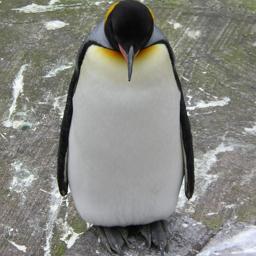}
    \end{subfigure} \\
    
    \begin{subfigure}[b]{0.16\linewidth}
        \centering
        \includegraphics[width=\linewidth]{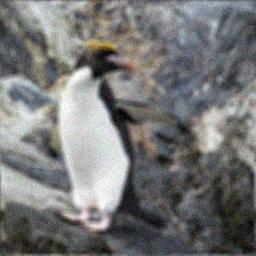}
    \end{subfigure}
     \begin{subfigure}[b]{0.16\linewidth}
        \centering
        \includegraphics[width=\linewidth]{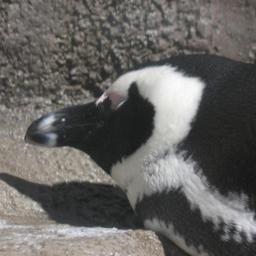}
    \end{subfigure}
    \begin{subfigure}[b]{0.16\linewidth}
        \centering
        \includegraphics[width=\linewidth]{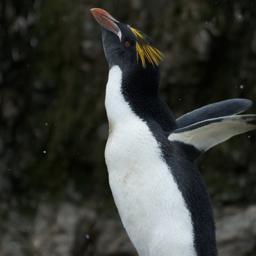}
    \end{subfigure}
    \begin{subfigure}[b]{0.16\linewidth}
        \centering
        \includegraphics[width=\linewidth]{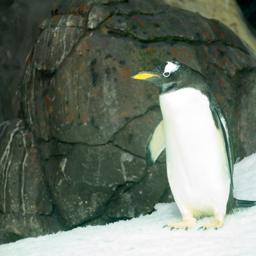}
    \end{subfigure}
    \begin{subfigure}[b]{0.16\linewidth}
        \centering
        \includegraphics[width=\linewidth]{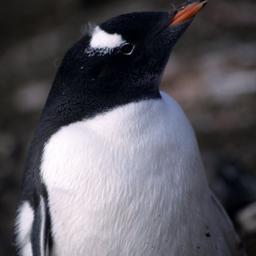}
    \end{subfigure}
    \begin{subfigure}[b]{0.16\linewidth}
        \centering
        \includegraphics[width=\linewidth]{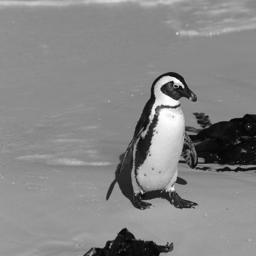}
    \end{subfigure}\\
    \begin{subfigure}[b]{0.16\linewidth}
        \centering
        \includegraphics[width=\linewidth]{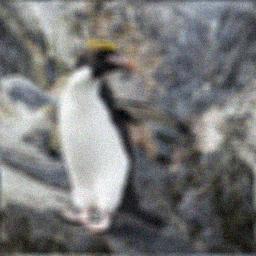}
    \end{subfigure}
     \begin{subfigure}[b]{0.16\linewidth}
        \centering
        \includegraphics[width=\linewidth]{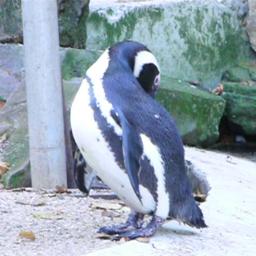}
    \end{subfigure}
    \begin{subfigure}[b]{0.16\linewidth}
        \centering
        \includegraphics[width=\linewidth]{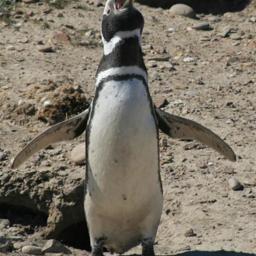}
    \end{subfigure}
    \begin{subfigure}[b]{0.16\linewidth}
        \centering
        \includegraphics[width=\linewidth]{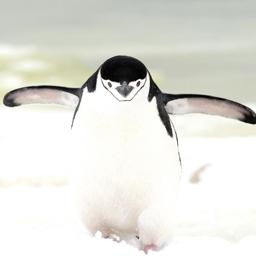}
    \end{subfigure}
    \begin{subfigure}[b]{0.16\linewidth}
        \centering
        \includegraphics[width=\linewidth]{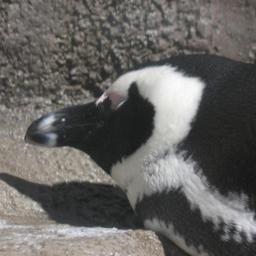}
    \end{subfigure}
    \begin{subfigure}[b]{0.16\linewidth}
        \centering
        \includegraphics[width=\linewidth]{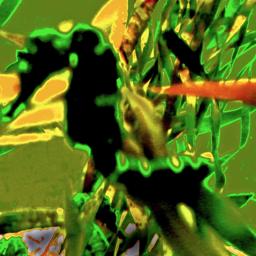}
    \end{subfigure}\\
    \begin{subfigure}[b]{0.16\linewidth}
        \centering
        \includegraphics[width=\linewidth]{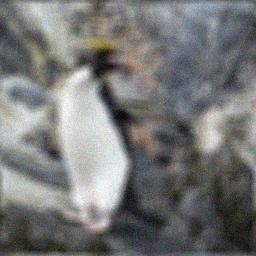}
        \subcaption{$\y$}
    \end{subfigure}
     \begin{subfigure}[b]{0.16\linewidth}
        \centering
        \includegraphics[width=\linewidth]{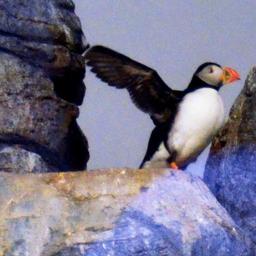}
        \subcaption{NN-16}
    \end{subfigure}
    \begin{subfigure}[b]{0.16\linewidth}
        \centering
        \includegraphics[width=\linewidth]{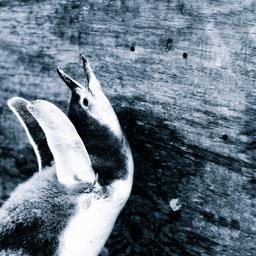}
        \subcaption{NN-17}
    \end{subfigure}
    \begin{subfigure}[b]{0.16\linewidth}
        \centering
        \includegraphics[width=\linewidth]{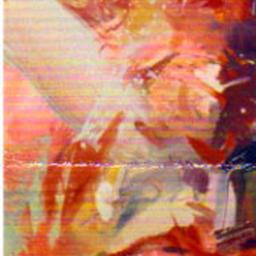}
        \subcaption{NN-18}
    \end{subfigure}
    \begin{subfigure}[b]{0.16\linewidth}
        \centering
        \includegraphics[width=\linewidth]{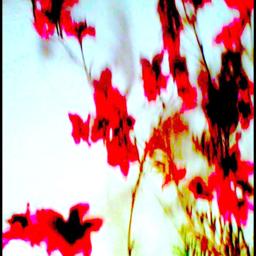}
        \subcaption{NN-19}
    \end{subfigure}
    \begin{subfigure}[b]{0.16\linewidth}
        \centering
        \includegraphics[width=\linewidth]{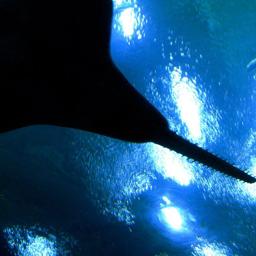}
        \subcaption{NN-20}
    \end{subfigure}
    \caption{The effect of $\A$ on the K-NN retrieval: Each row represent a different blur operator $\A$, and each column shows the 5 least similar images from the 20 retrieved nearest images from Google Open Dataset \citep{OpenImages} using CLIP \citep{CLIP}. In all cases we used a box filter with support $3\times3, 5\times5, 7\times7, 9\times9$ and $11\times11$, respective to the row number.}
    \label{fig:supp_A_effect_on_NN}
\end{figure*}

\footnotesize In the following we
\begin{itemize}
    \footnotesize
    \item Summarization of the method from section \ref{sec:method_guided_diff} in an algorithm fashion (Algorithm \ref{alg:guided_diff})
    \item Demonstrate the adaptation effectiveness on celebrity images, where the K-NN images are taken from the web (Figure \ref{fig:Faces}).
    \item Provide quantitative results for SR, deblurring, and colorization tasks (Tables \ref{table:SR}, \ref{table:GD_deblur} and \ref{table:GD_color}).
    \item Provide the hyper-parameters used for ADIR (Table \ref{table:appndx_configs}).
    \item Present an Ablation study where we compare ADIR to fine-tuning using random images and $K$-NN retrieved w.r.t. the MSE (Table \ref{table:KNN_ablation}).
    \item Examples of retrieved nearest neighbors images (Figure \ref{fig:supp_SRx8_NN}).
    \item Examine the effect of $\A$ on the $K$-NN retrieval (Figure \ref{fig:supp_A_effect_on_NN}).
    \item Show results for super resolution with scaling factor of 8.
    \item Show additional results of deblurring task.
    \item Show more results of colorization use-case.
    \item Compare our method to Stable Diffusion for editing task in multiple scenarios.
\end{itemize}

\begin{algorithm}[bh]
\footnotesize
\centering
    \caption{\footnotesize Proposed GD sampling for image reconstruction.}
    \label{alg:guided_diff}
    \begin{algorithmic}[1]
        \Require $(\bepsilon_\theta(\cdot), \bSigma_\theta(\cdot))$, $\y$, $s$
        \State $\x_{T} \leftarrow$ sample from $\mathcal{N}(\mathbf{0}, \I_n)$
        \For {$t$ from $T$ to $1$}
            \State $\hat{\bepsilon}$, $\hat{\bSigma}\leftarrow \bepsilon_\theta (\x_t,t)$, $\bSigma_\theta (\x_t,t)$
            \State $\hat{\bmu}\leftarrow \frac{1}{\sqrt{\alpha_t}} (\x_t - \frac{1-\alpha_t}{\sqrt{1 - \bar{\alpha}_t}}\hat{\bepsilon})$           
            \State $\y_t \leftarrow \sqrt{\bar{\alpha}_t}\y + \sqrt{1 - \bar{\alpha}_t}\A \hat{\epsilon}$
            \State $\g \leftarrow -2\A^T(\A\hat{\bmu} - \y_t)$            
            \State $\x_{t-1} \leftarrow$ sample from $\mathcal{N}(\hat{\bmu} + s \hat{\bSigma}\g, \hat{\bSigma})$
        \EndFor\\
        \Return $\x_0$
    \end{algorithmic}
\end{algorithm}

\begin{table}[!ht]
\centering
\resizebox{\textwidth}{!}{%
\begin{tabular}{@{}cccc@{}}
\begin{tabular}[t]{@{}lccc@{}}
\toprule
\textbf{Method} & \textbf{LPIPS$\downarrow$} & \textbf{AVA$\uparrow$} & \textbf{KonIQ$\uparrow$} \\
\midrule
IPT & 0.237 & 5.02 & 64.19 \\
USRNet & 0.249 & 4.38 & 45.76 \\
SwinIR & \textcolor{blue}{0.232} & 4.98 & 64.19 \\
SRDiff & \textbf{0.135} & 4.76 & 60.87 \\
Real-ESRGAN & 0.317 & \textcolor{blue}{5.02} & \textbf{69.42} \\
DeepRED & 0.475 & 3.20 & 22.77 \\
DDRM & 0.297 & 3.42 & 28.96 \\
GD (Baseline) & 0.325 & 4.88 & 64.63 \\
ADIR (GD) & 0.335 & \textbf{5.06} & \textcolor{blue}{66.33} \\
\bottomrule
\end{tabular}
&
\begin{tabular}[t]{@{}lccc@{}}
\toprule
\textbf{Method} & \textbf{LPIPS$\downarrow$} & \textbf{AVA$\uparrow$} & \textbf{KonIQ$\uparrow$} \\
\midrule
IPT & 0.221 & 4.90 & 65.38 \\
USRNet & 0.234 & 4.51 & 59.10 \\
SwinIR & \textcolor{blue}{0.218} & 4.88 & 65.08 \\
SRDiff & 0.237 & 4.76 & 62.64 \\
DeepRED & 0.405 & 3.25 & 25.26 \\
Real-ESRGAN & 0.305 & 4.93 & 69.11 \\
Stable Diffusion & 0.331 & \textcolor{blue}{5.07} & \textcolor{blue}{69.18} \\
ADIR (SD) & \textbf{0.213} & \textbf{5.51} & \textbf{72.56} \\
\bottomrule
\end{tabular}
&
\begin{tabular}[t]{@{}lccc@{}}
\toprule
\textbf{Method} & \textbf{LPIPS$\downarrow$} & \textbf{AVA$\uparrow$} & \textbf{KonIQ$\uparrow$} \\
\midrule
SwinIR & 0.424 & \textbf{4.54} & 48.04 \\
DeepRED & 0.591 & 2.99 & 17.43 \\
DDRM & 0.572 & 3.13 & 20.68 \\
GD (Baseline) & \textcolor{blue}{0.365} & 4.36 & \textcolor{blue}{53.99} \\
ADIR (GD) & \textbf{0.347} & \textcolor{blue}{4.41} & \textbf{55.89} \\
\bottomrule
\end{tabular}
\end{tabular}%
}
\caption{\small Super-resolution performance comparison: Guided Diffusion (SRx4 left, SRx8 middle) and Stable Diffusion (SRx4 right). Best in \textbf{bold}, second-best in \textcolor{blue}{blue}.}
\label{table:SR}
\end{table}

\begin{table}[th]
\centering
\resizebox{0.85\linewidth}{!}{%
\begin{tabular}{@{}l *{3}{ccc @{\hspace{1.5em}}}@{}}
\toprule
 & \multicolumn{3}{c}{Box ($256$)} & \multicolumn{3}{c}{Box ($512$)} & \multicolumn{3}{c}{Gauss ($256$)} \\
\textbf{Method} & LPIPS$\downarrow$ & AVA$\uparrow$ & KonIQ$\uparrow$ & LPIPS$\downarrow$ & AVA$\uparrow$ & KonIQ$\uparrow$ & LPIPS$\downarrow$ & AVA$\uparrow$ & KonIQ$\uparrow$ \\
\midrule
\textbf{M3SNet} \citep{gao2023mountain} & 0.477 & 3.13 & 26.16 & 0.468 & 2.93 & 47.42 & 0.481 & 2.75 & 31.13 \\
\textbf{DeepRED} \citep{mataev2019deepred} & 0.561 & 3.61 & 22.12 & 0.557 & 3.57 & 27.59 & 0.572 & 3.59 & 19.13 \\
\textbf{Restormer} \citep{zamir2022restormer} & \textbf{0.341} & 3.74 & 40.11 & \textcolor{blue}{0.377} & 4.67 & 55.03 & 0.518 & 3.61 & 36.70 \\
\textbf{MPRNet}  \citep{mehri2021mprnet} & 0.395 & 3.08 & 26.90 & 0.429 & 3.63 & 37.87 & 0.491 & 3.01 & 20.96 \\
\textbf{Guided Diffusion} \ref{sec:method_guided_diff} & 0.423 & \textcolor{blue}{4.20} & \textcolor{blue}{49.19} & 0.411 & \textbf{4.81} & \textcolor{blue}{58.66} & \textcolor{blue}{0.424} & \textcolor{blue}{4.01} & \textcolor{blue}{48.11} \\
\textbf{ADIR} & \textcolor{blue}{0.394} & \textbf{4.31} & \textbf{55.78} & \textbf{0.312} & \textcolor{blue}{4.77} & \textbf{60.13} & \textbf{0.415} & \textbf{4.19} & \textbf{51.80} \\
\bottomrule
\end{tabular}%
}
\caption{\small Deblurring with 10 noise levels results for the unconditional guided diffusion model \citep{guidedDiff}. Similar to SR in Table~\ref{table:SR}, the results are averaged on the first 50 images of the DIV2K validation set.}
\label{table:GD_deblur}
\end{table}

\begin{table}[t]
\centering
\begin{minipage}[t]{0.48\textwidth}
\centering
\resizebox{0.8\linewidth}{!}{%
\begin{tabular}{@{}lcc@{}}
\toprule
\textbf{Method} & \textbf{AVA$\uparrow$} & \textbf{KonIQ$\uparrow$} \\
\midrule
DDRM \citep{kawar2022denoising} & 4.012 & 53.458 \\
Guided Diffusion (GD) & \textcolor{blue}{4.195} & \textcolor{blue}{56.044} \\
ADIR (GD) & \textbf{4.214} & \textbf{58.679} \\
\bottomrule
\end{tabular}%
}
\caption{\small Image colorization results
on the first 50 images of the DIV2K validation set \citep{div2k}. }
\label{table:GD_color}
\end{minipage}
\hfill
\begin{minipage}[t]{0.48\textwidth}
\centering
\resizebox{\linewidth}{!}{%
\begin{tabular}{@{}lccccc@{}}
\toprule
\textbf{Method} & \textbf{IA Iter.} & \textbf{LR} & \textbf{NN Img.} & $\mathbf{s}$ & \textbf{Steps}\\
\midrule
ADIR-GD & 400 & $10^{-4}$ & 20 & 10 & 1000 \\
ADIR-SD & 400 & $10^{-4}$ & 50 & -- & 50 \\ 
\bottomrule
\end{tabular}%
}
\caption{\small Configurations used for ADIR. Abbreviations: Iter.=Iterations, Img.=Images, Steps=Diffusion steps.}
\label{table:appndx_configs}
\end{minipage}
\end{table}

\begin{table}[th]
\centering
\resizebox{\linewidth}{!}{
\begin{tabular}{@{}l @{\hspace{3em}} cccc @{\hspace{3em}} cccc @{}}
\toprule
 & \multicolumn{4}{c}{SRx8 ($512\times512$)} & \multicolumn{4}{c}{Deblur ($256\times256$)} \\
\textbf{Method} & LPIPS$\downarrow$ & AVA$\uparrow$ & KonIQ$\uparrow$ & Runtime [s] & LPIPS$\downarrow$ & AVA$\uparrow$ & KonIQ$\uparrow$ & Runtime [s] \\
\midrule
GD (baseline) & \textcolor{blue}{0.365} & \textcolor{blue}{4.36} & \textcolor{blue}{53.99} & \textbf{830} & \textcolor{blue}{0.423} & \textcolor{blue}{4.20} & \textcolor{blue}{49.19} & \textbf{425} \\
IA w/ random images & 0.430 & 4.14 & 52.89 & \textcolor{blue}{1300} & 0.433 & 3.68 & 40.05 & \textcolor{blue}{895} \\
IA w/ MSE based NN & 0.434 & 4.28 & 53.12 & 2700 & 0.428 & 3.62 & 39.13 & 2500 \\
\textbf{ADIR} & \textbf{0.347} & \textbf{4.41} & \textbf{55.89} & 1308 & \textbf{0.394} & \textbf{4.31} & \textbf{55.78} & 903 \\
\bottomrule
\end{tabular}%
}
\caption{Ablation study of the adaptation advantage of ADIR. We compare our method to adaptation using random images sampled from the dataset, adaptation using MSE as a retrieval distance, and the approach from Section \ref{sec:method_guided_diff}. We use the traditional LPIPS \citep{zhang2018unreasonable} as well as the state-of-the-art no reference perceptual losses AVA-MUSIQ and KonIQ-MUSIQ \citep{ke2021musiq} for evaluation. We also benchmark the runtime of each method and report the inference time per image on a single NVIDIA RTX A6000 GPU. The best results are in bold black, and the second best is highlighted in blue. In ADIR we achieve much smaller runtimes compared to the MSE-based retrieval because we use an efficient K-D tree data structure on the embedding space that is much smaller than the naive pixel domain.}
\label{table:KNN_ablation}
\end{table}

\newpage

\begin{figure*}[!hb]
\captionsetup[subfigure]{labelformat=empty}
    \centering
    \begin{subfigure}[b]{0.2\linewidth}
        \centering
        \begin{overpic}[width=\linewidth]{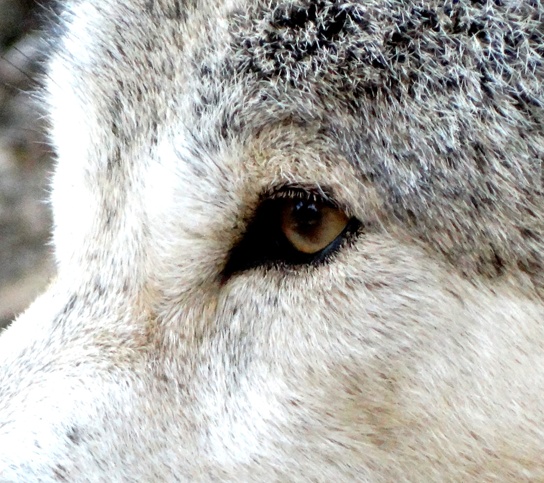}
            \put(0.5\linewidth,0){\frame{\includegraphics[width=0.5\linewidth, trim={230 250 160 100}, clip]{LDM_SR/DIV2K_004_GT.jpg}} }
        \end{overpic}
   
    \end{subfigure}
     \begin{subfigure}[b]{0.2\linewidth}
        \centering
        \begin{overpic}[width=\linewidth]{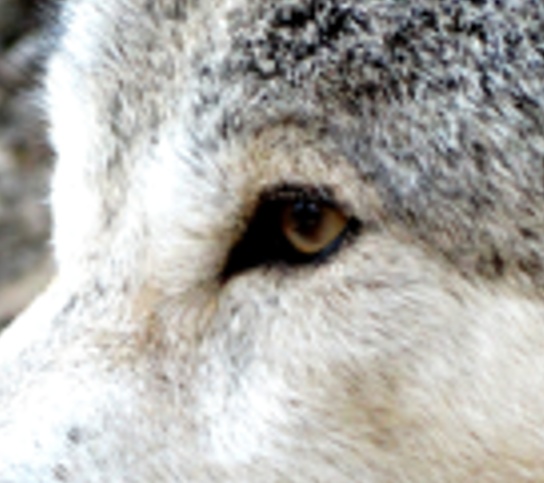}
            \put(0.5\linewidth,0){\frame{\includegraphics[width=0.5\linewidth, trim={230 250 160 100}, clip]{LDM_SR/DIV2K_004_Bicubic.jpg}} }
        \end{overpic}
   
    \end{subfigure}
    \begin{subfigure}[b]{0.2\linewidth}
        \centering
        \begin{overpic}[width=\linewidth]{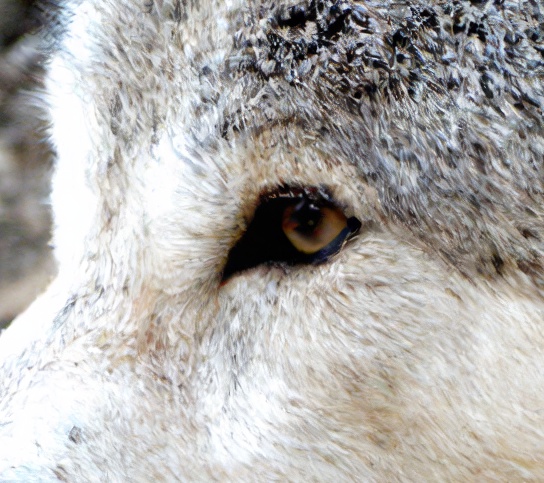}
            \put(0.5\linewidth,0){\frame{\includegraphics[width=0.5\linewidth, trim={230 250 160 100}, clip]{LDM_SR/DIV2K_004_LDM.jpg}} }
        \end{overpic}

    \end{subfigure}
    \begin{subfigure}[b]{0.2\linewidth}
        \centering
        \begin{overpic}[width=\linewidth]{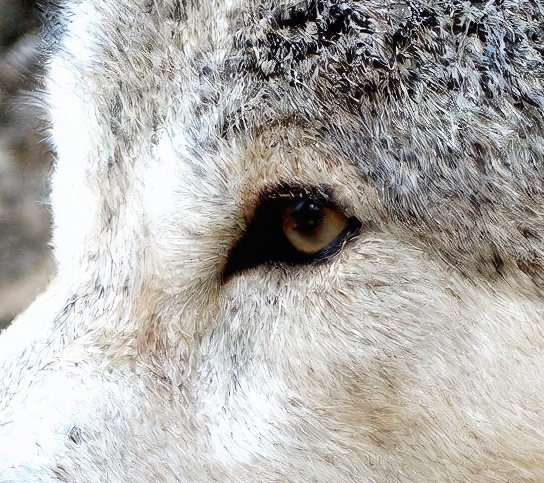}
            \put(0.5\linewidth,0){\frame{\includegraphics[width=0.5\linewidth, trim={230 250 160 100}, clip]{LDM_SR/DIV2K_004_ADIR.jpg}} }
        \end{overpic}
        
    \end{subfigure}\\
    \begin{subfigure}[b]{0.2\linewidth}
        \centering
        \begin{overpic}[width=\linewidth]{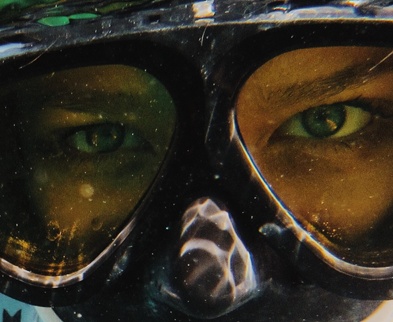}
            \put(0.4\linewidth,0.1\linewidth){\frame{\includegraphics[width=0.6\linewidth, trim={260 170 0 90}, clip]{LDM_SR/DIV2K_016_GT.jpg}} }
        \end{overpic}
   
    \end{subfigure}
     \begin{subfigure}[b]{0.2\linewidth}
        \centering
        \begin{overpic}[width=\linewidth]{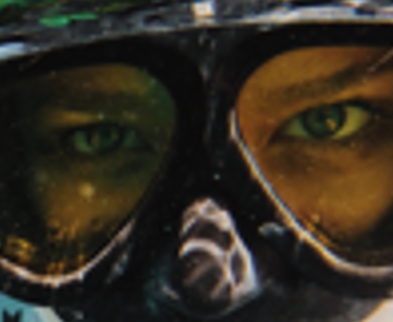}
            \put(0.4\linewidth,0.1\linewidth){\frame{\includegraphics[width=0.6\linewidth, trim={260 170 0 90}, clip]{LDM_SR/DIV2K_016_Bicubic.jpg}} }
        \end{overpic}
   
    \end{subfigure}
    \begin{subfigure}[b]{0.2\linewidth}
        \centering
        \begin{overpic}[width=\linewidth]{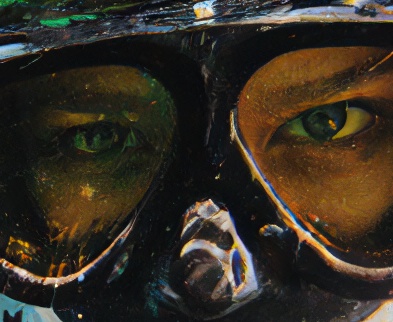}
            \put(0.4\linewidth,0.1\linewidth){\frame{\includegraphics[width=0.6\linewidth, trim={260 170 0 90}, clip]{LDM_SR/DIV2K_016_LDM.jpg}} }
        \end{overpic}

    \end{subfigure}
    \begin{subfigure}[b]{0.2\linewidth}
        \centering
        \begin{overpic}[width=\linewidth]{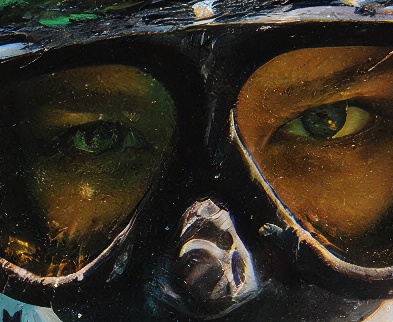}
            \put(0.4\linewidth,0.1\linewidth){\frame{\includegraphics[width=0.6\linewidth, trim={260 170 0 90}, clip]{LDM_SR/DIV2K_016_ADIR.jpg}} }
        \end{overpic}
        
    \end{subfigure}\\
    \begin{subfigure}[b]{0.2\linewidth}
        \centering
        \begin{overpic}[width=\linewidth]{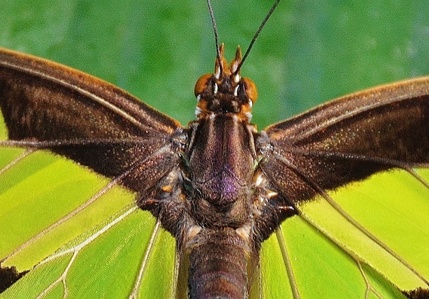}
            \put(0.4\linewidth,0.0\linewidth){\frame{\includegraphics[width=0.6\linewidth, trim={170 130 90 60}, clip]{LDM_SR/DIV2K_028_GT.jpg}} }
        \end{overpic}
        \subcaption{GT}
   
    \end{subfigure}
     \begin{subfigure}[b]{0.2\linewidth}
        \centering
        \begin{overpic}[width=\linewidth]{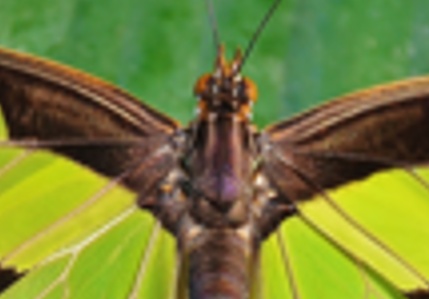}
            \put(0.4\linewidth,0.0\linewidth){\frame{\includegraphics[width=0.6\linewidth, trim={170 130 90 60}, clip]{LDM_SR/DIV2K_028_Bicubic.jpg}} }
        \end{overpic}
        
        \subcaption{Bicubic}
    \end{subfigure}
    \begin{subfigure}[b]{0.2\linewidth}
        \centering
        \begin{overpic}[width=\linewidth]{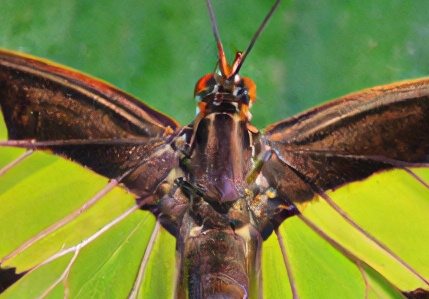}
            \put(0.4\linewidth,0.0\linewidth){\frame{\includegraphics[width=0.6\linewidth, trim={170 130 90 60}, clip]{LDM_SR/DIV2K_028_LDM.jpg}} }
        \end{overpic}
        \subcaption{Stable Diffusion}
    \end{subfigure}
    \begin{subfigure}[b]{0.2\linewidth}
        \centering
        \begin{overpic}[width=\linewidth]{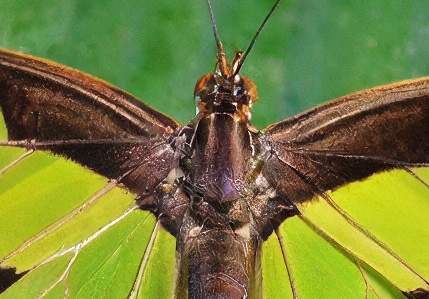}
            \put(0.4\linewidth,0.0\linewidth){\frame{\includegraphics[width=0.6\linewidth, trim={170 130 90 60}, clip]{LDM_SR/DIV2K_028_ADIR.jpg}} }
        \end{overpic}
        \subcaption{ADIR}
    \end{subfigure}

    \caption{Comparison of super resolution ($256^2 \rightarrow 1024^2$) results of Stable Diffusion model\citep{rombach2022high} and our method (ADIR). As can be seen from the images, our method outperforms Stable Diffusion in both sharpness and reconstructing details.}

    \label{fig:LDM_SRx4}
\end{figure*}

\begin{figure*}
\centering
    \includegraphics[width=0.19\linewidth]{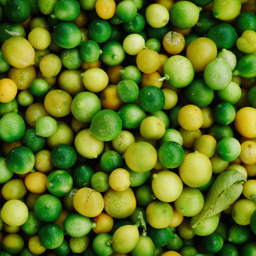}
    \includegraphics[width=0.19\linewidth]{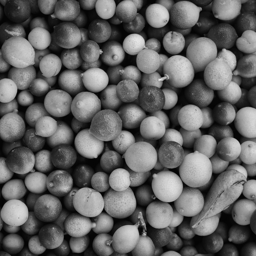}
    \includegraphics[width=0.19\linewidth]{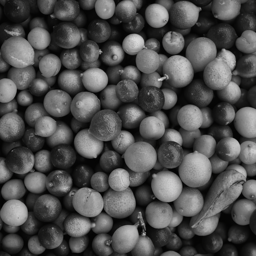}
    \includegraphics[width=0.19\linewidth]{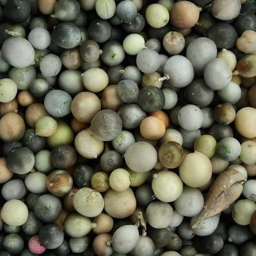}
    \includegraphics[width=0.19\linewidth]{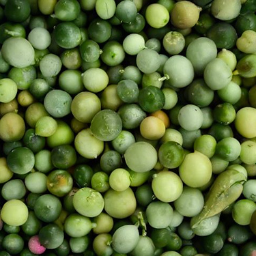}
    \includegraphics[width=0.19\linewidth]{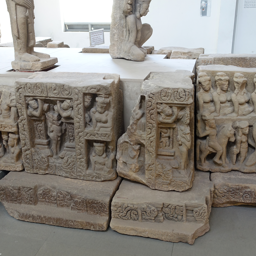}
    \includegraphics[width=0.19\linewidth]{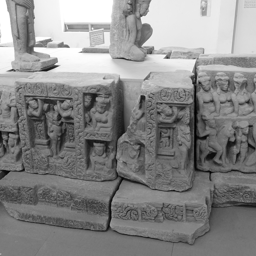}
    \includegraphics[width=0.19\linewidth]{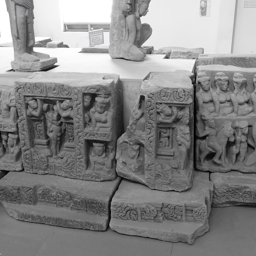}
    \includegraphics[width=0.19\linewidth]{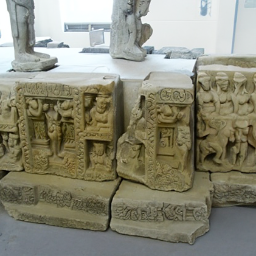}
    \includegraphics[width=0.19\linewidth]{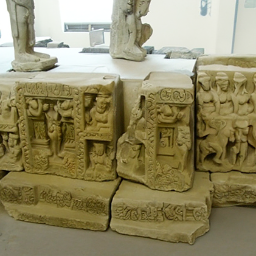}
    \includegraphics[width=0.19\linewidth]{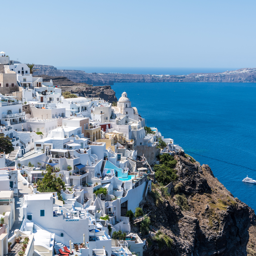}
    \includegraphics[width=0.19\linewidth]{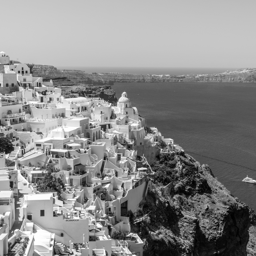}
    \includegraphics[width=0.19\linewidth]{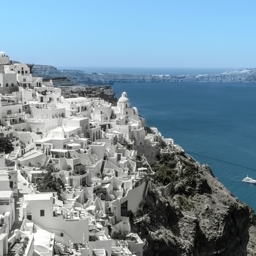}
    \includegraphics[width=0.19\linewidth]{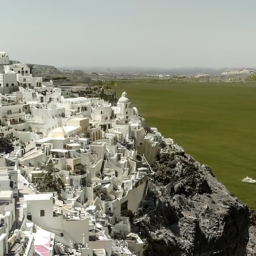}
    \includegraphics[width=0.19\linewidth]{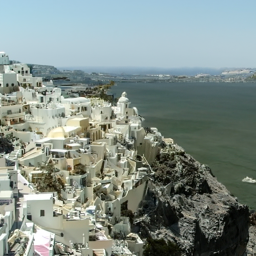}
    \subcaptionbox*{Original Image}{\includegraphics[width=0.19\linewidth]{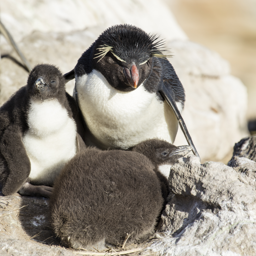}}
    \subcaptionbox*{Grayscale}{\includegraphics[width=0.19\linewidth]{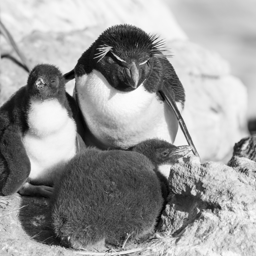}}
    \subcaptionbox*{DDRM}{\includegraphics[width=0.19\linewidth]{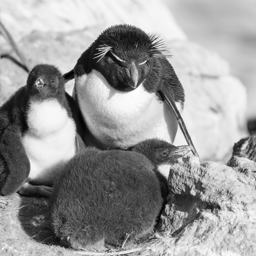}}
    \subcaptionbox*{Guided Diffusion}{\includegraphics[width=0.19\linewidth]{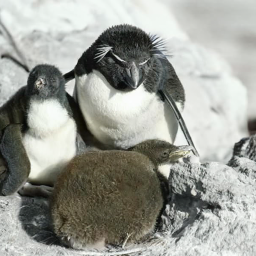}}
    \subcaptionbox*{ADIR}{\includegraphics[width=0.19\linewidth]{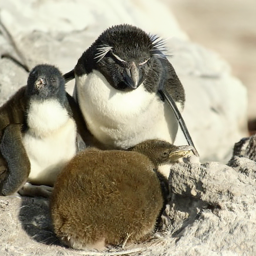}}
  \caption{Image colorization results comparison between DDRM \citep{kawar2022denoising}, Guided diffusion proposed in section \ref{sec:method_guided_diff}, and our adaptive approach ADIR. As can be seen, adapting the denoiser network to the given image can improve the results significantly.}
  \label{fig:GD_color}
\end{figure*}

\begin{figure}[!h]
\captionsetup[subfigure]{labelformat=empty}
    \centering
    \begin{subfigure}[b]{0.24\linewidth}
        \centering
        \includegraphics[width=\linewidth]{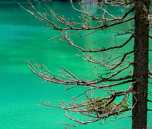}
   
    \end{subfigure}
     \begin{subfigure}[b]{0.24\linewidth}
        \centering
        \includegraphics[width=\linewidth]{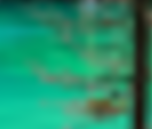}
   
    \end{subfigure}
    \begin{subfigure}[b]{0.24\linewidth}
        \centering
        \includegraphics[width=\linewidth]{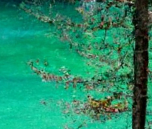}

    \end{subfigure}
    \begin{subfigure}[b]{0.24\linewidth}
        \centering
        \includegraphics[width=\linewidth]{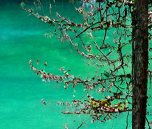}
    \end{subfigure}\\
    \begin{subfigure}[b]{0.24\linewidth}
        \centering
        \includegraphics[width=\linewidth]{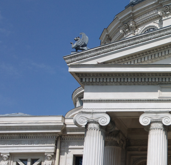}
   
    \end{subfigure}
     \begin{subfigure}[b]{0.24\linewidth}
        \centering
        \includegraphics[width=\linewidth]{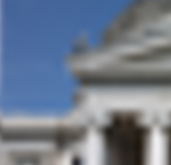}
   
    \end{subfigure}
    \begin{subfigure}[b]{0.24\linewidth}
        \centering
        \includegraphics[width=\linewidth]{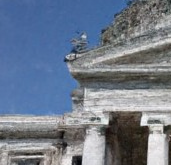}

    \end{subfigure}
    \begin{subfigure}[b]{0.24\linewidth}
        \centering
        \includegraphics[width=\linewidth]{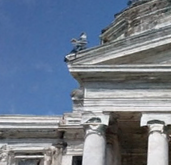}
    \end{subfigure}\\
    \begin{subfigure}[b]{0.24\linewidth}
        \centering
        \includegraphics[width=\linewidth]{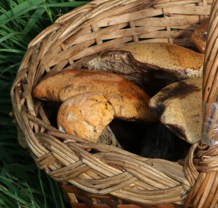}
   
    \end{subfigure}
     \begin{subfigure}[b]{0.24\linewidth}
        \centering
        \includegraphics[width=\linewidth]{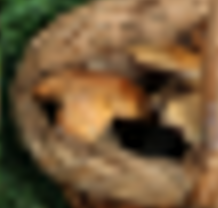}
   
    \end{subfigure}
    \begin{subfigure}[b]{0.24\linewidth}
        \centering
        \includegraphics[width=\linewidth]{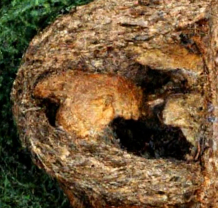}

    \end{subfigure}
    \begin{subfigure}[b]{0.24\linewidth}
        \centering
        \includegraphics[width=\linewidth]{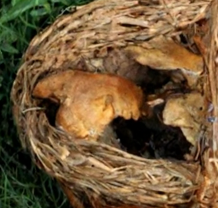}
    \end{subfigure}\\
    \begin{subfigure}[b]{0.24\linewidth}
        \centering
        \includegraphics[width=\linewidth, trim={0, 10pt, 0, 0}, clip]{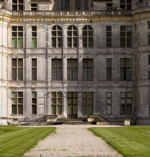}
   
    \end{subfigure}
     \begin{subfigure}[b]{0.24\linewidth}
        \centering
        \includegraphics[width=\linewidth, trim={0, 10pt, 0, 0}, clip]{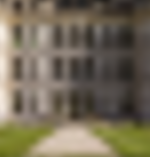}
   
    \end{subfigure}
    \begin{subfigure}[b]{0.24\linewidth}
        \centering
        \includegraphics[width=\linewidth, trim={0, 10pt, 0, 0}, clip]{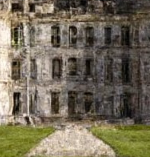}

    \end{subfigure}
    \begin{subfigure}[b]{0.24\linewidth}
        \centering
        \includegraphics[width=\linewidth, trim={0, 10pt, 0, 0}, clip]{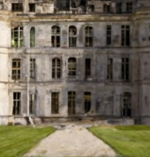}
    \end{subfigure}\\
    \begin{subfigure}[b]{0.24\linewidth}
        \centering
        \includegraphics[width=\linewidth]{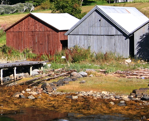}
        \subcaption{GT}
   
    \end{subfigure}
     \begin{subfigure}[b]{0.24\linewidth}
        \centering
        \includegraphics[width=\linewidth]{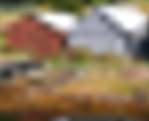}
        \subcaption{Bicubic}
    \end{subfigure}
    \begin{subfigure}[b]{0.24\linewidth}
        \centering
        \includegraphics[width=\linewidth]{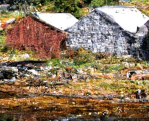}
        \subcaption{Guided Diffusion}
    \end{subfigure}
    \begin{subfigure}[b]{0.24\linewidth}
        \centering
        \includegraphics[width=\linewidth]{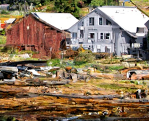}
        \subcaption{ADIR}
    \end{subfigure}
    \caption{Comparison of super resolution ($64^2 \rightarrow 512^2$) results of Guided Diffusion from section 3.2 and our method (ADIR), using the unconditional model from \citep{rombach2022high}. As can be seen from the images, our method outperforms guided diffusion in both sharpness and reconstruction details.}
\end{figure}

\begin{figure}
\captionsetup[subfigure]{labelformat=empty}
    \centering
    \begin{subfigure}[b]{0.45\linewidth}
        \centering
        \begin{overpic}[width=\linewidth]{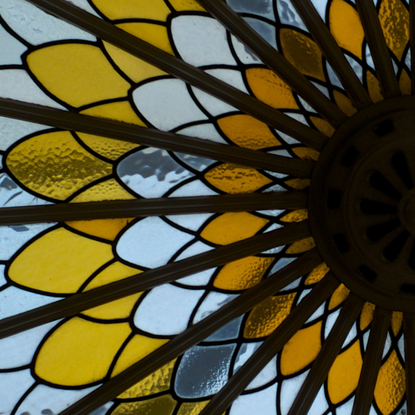}
            \put(0.6\linewidth,0){\frame{\includegraphics[width=0.3999\linewidth, trim={130 0 170 300}, clip]{supp_figs/LDM_SR/020_GT.png}} }
        \end{overpic}
        \subcaption{GT}
   
    \end{subfigure}
     \begin{subfigure}[b]{0.45\linewidth}
        \centering
        \begin{overpic}[width=\linewidth]{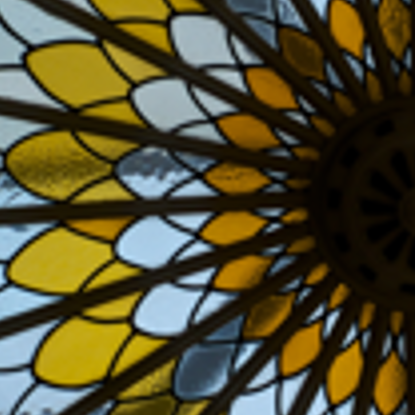}
            \put(0.6\linewidth,0){\frame{\includegraphics[width=0.3999\linewidth, trim={130 0 170 300}, clip]{supp_figs/LDM_SR/020_SRx4_Sty.png}} }
        \end{overpic}
        \subcaption{Bicubic}
    \end{subfigure}\\
    \begin{subfigure}[b]{0.45\linewidth}
        \centering
        
        \begin{overpic}[width=\linewidth]{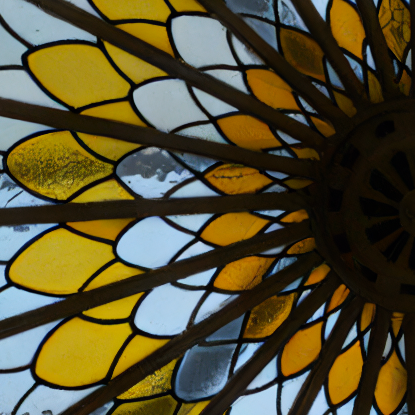}
            \put(0.6\linewidth,0){\frame{\includegraphics[width=0.3999\linewidth, trim={130 0 170 300}, clip]{supp_figs/LDM_SR/020_LDM_SRx4_config0.png}} }
        \end{overpic}
        \subcaption{Stable Diffusion}
    \end{subfigure}
    \begin{subfigure}[b]{0.45\linewidth}
        \centering
        \begin{overpic}[width=\linewidth]{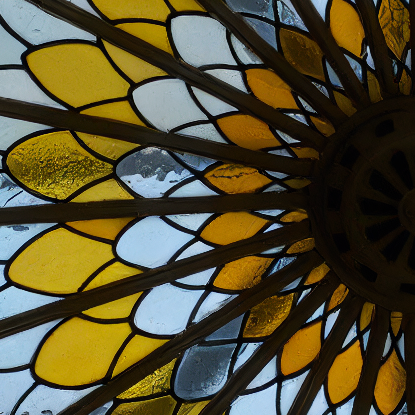}
            \put(0.6\linewidth,0){\frame{\includegraphics[width=0.3999\linewidth, trim={130 0 170 300}, clip]{supp_figs/LDM_SR/020_LDM_SRx4__IA400_50NN_config34_google.png}} }
        \end{overpic}
        \subcaption{ADIR}
    \end{subfigure}
    \caption{Comparison of super resolution ($256^2 \rightarrow 1024^2$) results of Stable Diffusion \citep{rombach2022high} and our method (ADIR), using the unconditional model from \citep{rombach2022high}. As can be seen from the images, our method outperforms guided diffusion in both sharpness and reconstruction details.}
\end{figure}

\begin{figure}
\captionsetup[subfigure]{labelformat=empty}
    \centering
    \begin{subfigure}[b]{0.45\linewidth}
        \centering
        
        \begin{overpic}[width=\linewidth]{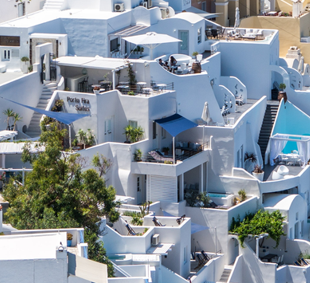}
            \put(0.6\linewidth,0){\frame{\includegraphics[width=0.3999\linewidth, trim={160 65 50 160}, clip]{supp_figs/LDM_SR/022_GT.png}} }
        \end{overpic}
        \subcaption{GT}
   
    \end{subfigure}
     \begin{subfigure}[b]{0.45\linewidth}
        \centering
        
        \begin{overpic}[width=\linewidth]{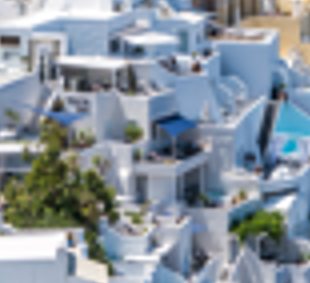}
            \put(0.6\linewidth,0){\frame{\includegraphics[width=0.3999\linewidth, trim={160 65 50 160}, clip]{supp_figs/LDM_SR/022_SRx4_Sty.png}} }
        \end{overpic}
        \subcaption{Bicubic}
    \end{subfigure}\\
    \begin{subfigure}[b]{0.45\linewidth}
        \centering
        
        \begin{overpic}[width=\linewidth]{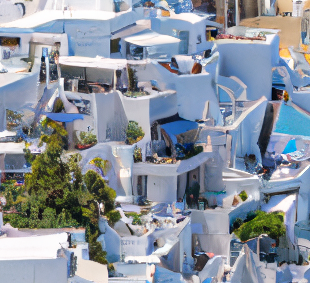}
            \put(0.6\linewidth,0){\frame{\includegraphics[width=0.3999\linewidth, trim={160 65 50 160}, clip]{supp_figs/LDM_SR/022_LDM_SRx4_config0.png}} }
        \end{overpic}
        \subcaption{Stable Diffusion}
    \end{subfigure}
    \begin{subfigure}[b]{0.45\linewidth}
        \centering
        \begin{overpic}[width=\linewidth]{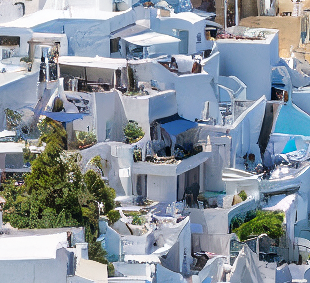}
            \put(0.6\linewidth,0){\frame{\includegraphics[width=0.3999\linewidth, trim={160 65 50 160}, clip]{supp_figs/LDM_SR/022_LDM_SRx4__IA400_50NN_config34_google.png}} }
        \end{overpic}
        \subcaption{ADIR}
    \end{subfigure}
    \caption{Comparison of super resolution ($256^2 \rightarrow 1024^2$) results of Stable Diffusion \citep{rombach2022high} and our method (ADIR), using the unconditional model from \citep{rombach2022high}. As can be seen from the images, our method outperforms guided diffusion in both sharpness and reconstruction details.}
\end{figure}

\begin{figure}
\captionsetup[subfigure]{labelformat=empty}
    \centering
    \begin{subfigure}[b]{0.45\linewidth}
        \centering
        
        \begin{overpic}[width=\linewidth]{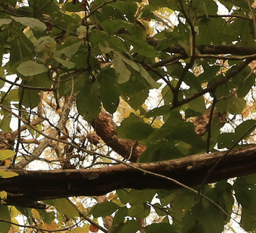}
            \put(0.6\linewidth,0){\frame{\includegraphics[width=0.3999\linewidth, trim={180 165 0 0}, clip]{supp_figs/LDM_SR/048_GT.png}} }
        \end{overpic}
        \subcaption{GT}
   
    \end{subfigure}
     \begin{subfigure}[b]{0.45\linewidth}
        \centering
        
        \begin{overpic}[width=\linewidth]{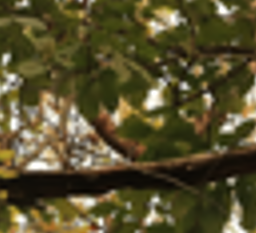}
            \put(0.6\linewidth,0){\frame{\includegraphics[width=0.3999\linewidth, trim={180 165 0 0}, clip]{supp_figs/LDM_SR/048_SRx4_Sty.png}} }
        \end{overpic}
        \subcaption{Bicubic}
    \end{subfigure}\\
    \begin{subfigure}[b]{0.45\linewidth}
        \centering
        
        \begin{overpic}[width=\linewidth]{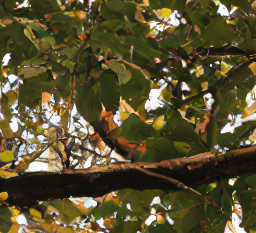}
            \put(0.6\linewidth,0){\frame{\includegraphics[width=0.3999\linewidth, trim={180 165 0 0}, clip]{supp_figs/LDM_SR/048_LDM_SRx4_config0.png}} }
        \end{overpic}
        \subcaption{Stable Diffusion}
    \end{subfigure}
    \begin{subfigure}[b]{0.45\linewidth}
        \centering
        \begin{overpic}[width=\linewidth]{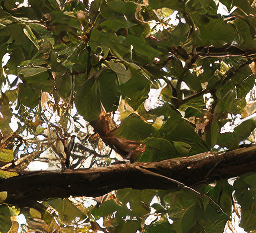}
            \put(0.6\linewidth,0){\frame{\includegraphics[width=0.3999\linewidth, trim={180 165 0 0}, clip]{supp_figs/LDM_SR/048_LDM_SRx4__IA400_50NN_config34_google.png}} }
        \end{overpic}
        \subcaption{ADIR}
    \end{subfigure}
    \caption{Comparison of super resolution ($256^2 \rightarrow 1024^2$) results of Stable Diffusion \citep{rombach2022high} and our method (ADIR), using the unconditional model from \citep{rombach2022high}. As can be seen from the images, our method outperforms guided diffusion in both sharpness and reconstruction details.}
\end{figure}

\begin{figure*}[t]
\captionsetup[subfigure]{labelformat=empty}
    \centering
    \begin{subfigure}[b]{0.24\linewidth}
        \centering
        \begin{overpic}[width=\linewidth]{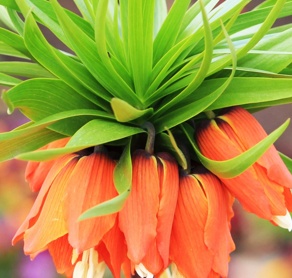}
            \put(0.6\linewidth,0){\frame{\includegraphics[width=0.3999\linewidth, trim={130 85 70 85}, clip]{GD_deblur/00003_GT.jpg}} }
        \end{overpic}
   
    \end{subfigure}
     \begin{subfigure}[b]{0.24\linewidth}
        \centering
        \begin{overpic}[width=\linewidth]{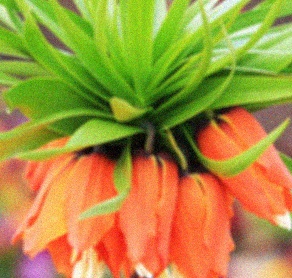}
            \put(0.6\linewidth,0){\frame{\includegraphics[width=0.3999\linewidth, trim={130 85 70 85}, clip]{GD_deblur/00003_deblur_y.jpg}} }
        \end{overpic}
   
    \end{subfigure}
    \begin{subfigure}[b]{0.24\linewidth}
        \centering
        \begin{overpic}[width=\linewidth]{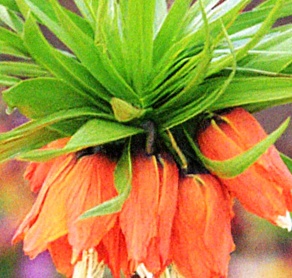}
            \put(0.6\linewidth,0){\frame{\includegraphics[width=0.3999\linewidth, trim={130 85 70 85}, clip]{GD_deblur/00003_deblur_s10_config0_ADIR.jpg}} }
        \end{overpic}

    \end{subfigure}
    \begin{subfigure}[b]{0.24\linewidth}
        \centering
        \begin{overpic}[width=\linewidth]{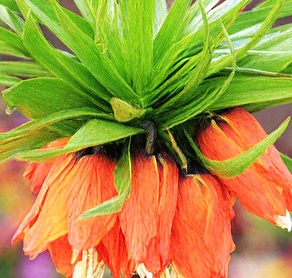}
            \put(0.6\linewidth,0){\frame{\includegraphics[width=0.3999\linewidth, trim={130 85 70 85}, clip]{GD_deblur/00003_deblur_s10_LS_IA400_20NN_config84.jpg}} }
        \end{overpic}
    \end{subfigure}\\
    \begin{subfigure}[b]{0.24\linewidth}
        \centering
        \begin{overpic}[width=\linewidth]{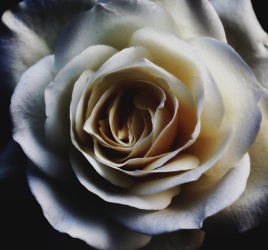}
            \put(0.6\linewidth,0){\frame{\includegraphics[width=0.3999\linewidth, trim={130 85 70 85}, clip]{GD_deblur/00043_GT.jpg}} }
        \end{overpic}        
        \subcaption{GT}
   
    \end{subfigure}
     \begin{subfigure}[b]{0.24\linewidth}
        \centering
        \begin{overpic}[width=\linewidth]{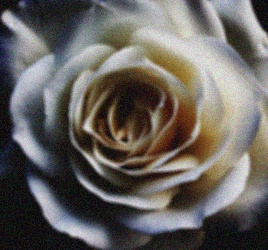}
            \put(0.6\linewidth,0){\frame{\includegraphics[width=0.3999\linewidth, trim={130 85 70 85}, clip]{GD_deblur/00043_deblur_y.jpg}} }
        \end{overpic}
        \subcaption{Blurry}
    \end{subfigure}
    \begin{subfigure}[b]{0.24\linewidth}
        \centering
        \begin{overpic}[width=\linewidth]{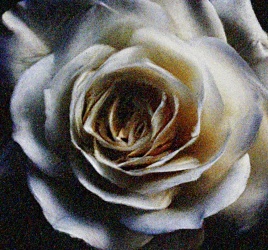}
            \put(0.6\linewidth,0){\frame{\includegraphics[width=0.3999\linewidth, trim={130 85 70 85}, clip]{GD_deblur/00043_deblur_s10_config0_ADIR.jpg}} }
        \end{overpic}
        \subcaption{Guided Diffusion}
    \end{subfigure}
    \begin{subfigure}[b]{0.24\linewidth}
        \centering
        \begin{overpic}[width=\linewidth]{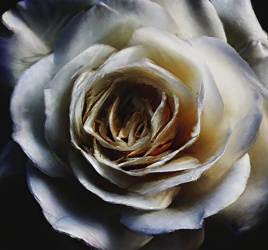}
            \put(0.6\linewidth,0){\frame{\includegraphics[width=0.3999\linewidth, trim={130 85 70 85}, clip]{GD_deblur/00043_deblur_s10_LS_IA400_20NN_config84.jpg}} }
        \end{overpic}
        \subcaption{ADIR}
    \end{subfigure}
    \vspace{-0.15in}
    \caption{Image deblurring using Guided Diffusion approach from section \ref{sec:method_guided_diff} and ADIR, using the unconditional model from \citep{guidedDiff}. The degradation is performed using $5\times 5$ uniform blur filter with 10 levels of additive Gaussian noise. Note the better quality of our method.}
    \label{fig:GD_Deblur}
\end{figure*}

\begin{figure}
\captionsetup[subfigure]{labelformat=empty}
    \centering
    \begin{subfigure}[b]{0.24\linewidth}
        \centering
        \includegraphics[width=\linewidth]{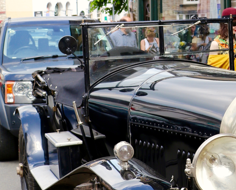}
   
    \end{subfigure}
     \begin{subfigure}[b]{0.24\linewidth}
        \centering
        \includegraphics[width=\linewidth]{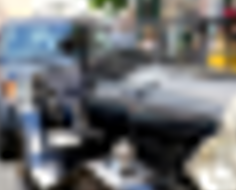}
   
    \end{subfigure}
    \begin{subfigure}[b]{0.24\linewidth}
        \centering
        \includegraphics[width=\linewidth]{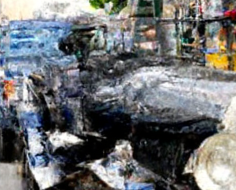}

    \end{subfigure}
    \begin{subfigure}[b]{0.24\linewidth}
        \centering
        \includegraphics[width=\linewidth]{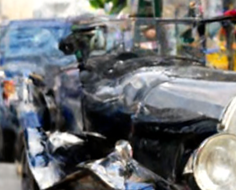}
    \end{subfigure}\\
    \begin{subfigure}[b]{0.24\linewidth}
        \centering
        \includegraphics[width=\linewidth]{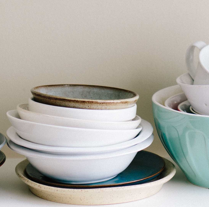}
   
    \end{subfigure}
     \begin{subfigure}[b]{0.24\linewidth}
        \centering
        \includegraphics[width=\linewidth]{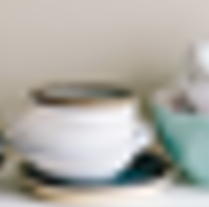}
   
    \end{subfigure}
    \begin{subfigure}[b]{0.24\linewidth}
        \centering
        \includegraphics[width=\linewidth]{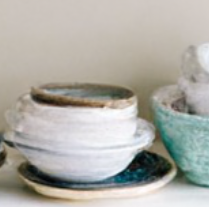}

    \end{subfigure}
    \begin{subfigure}[b]{0.24\linewidth}
        \centering
        \includegraphics[width=\linewidth]{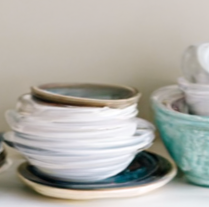}
    \end{subfigure}\\
    \begin{subfigure}[b]{0.24\linewidth}
        \centering
        \includegraphics[width=\linewidth]{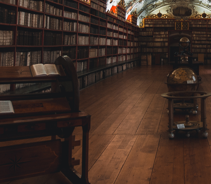}
   
    \end{subfigure}
     \begin{subfigure}[b]{0.24\linewidth}
        \centering
        \includegraphics[width=\linewidth]{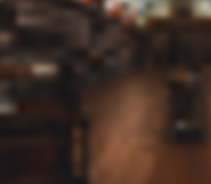}
   
    \end{subfigure}
    \begin{subfigure}[b]{0.24\linewidth}
        \centering
        \includegraphics[width=\linewidth]{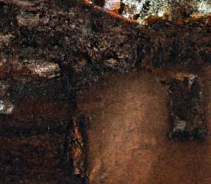}

    \end{subfigure}
    \begin{subfigure}[b]{0.24\linewidth}
        \centering
        \includegraphics[width=\linewidth]{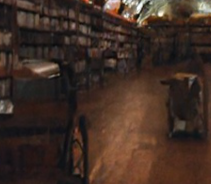}
    \end{subfigure}\\
    \begin{subfigure}[b]{0.24\linewidth}
        \centering
        \includegraphics[width=\linewidth]{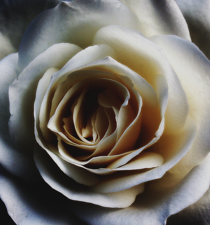}
   
    \end{subfigure}
     \begin{subfigure}[b]{0.24\linewidth}
        \centering
        \includegraphics[width=\linewidth]{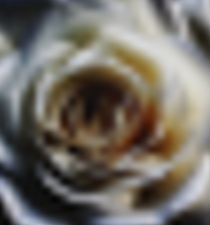}
   
    \end{subfigure}
    \begin{subfigure}[b]{0.24\linewidth}
        \centering
        \includegraphics[width=\linewidth]{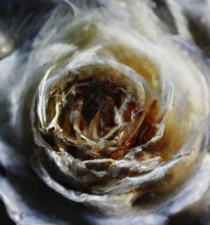}

    \end{subfigure}
    \begin{subfigure}[b]{0.24\linewidth}
        \centering
        \includegraphics[width=\linewidth]{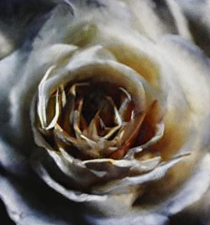}
    \end{subfigure}\\
    \begin{subfigure}[b]{0.24\linewidth}
        \centering
        \includegraphics[width=\linewidth]{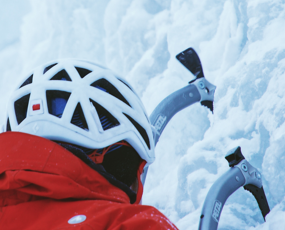}
        \subcaption{GT}
   
    \end{subfigure}
     \begin{subfigure}[b]{0.24\linewidth}
        \centering
        \includegraphics[width=\linewidth]{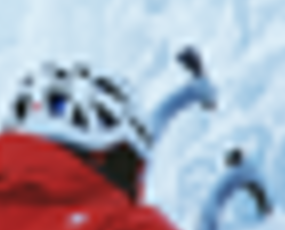}
        \subcaption{Blurry}
    \end{subfigure}
    \begin{subfigure}[b]{0.24\linewidth}
        \centering
        \includegraphics[width=\linewidth]{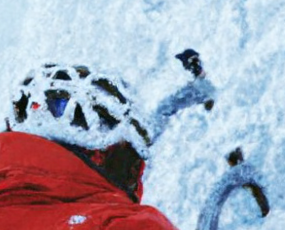}
        \subcaption{Guided Diffusion}
    \end{subfigure}
    \begin{subfigure}[b]{0.24\linewidth}
        \centering
        \includegraphics[width=\linewidth]{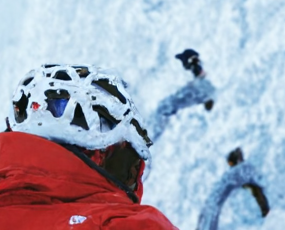}
        \subcaption{ADIR}
    \end{subfigure}
    \caption{Comparison of super resolution ($64^2 \rightarrow 512^2$) results of Guided Diffusion from section 3.2 and our method (ADIR), using the unconditional model from \citep{rombach2022high}. As can be seen from the images, our method outperforms guided diffusion in both sharpness and reconstruction details.}
\end{figure}

\begin{figure}
\captionsetup[subfigure]{labelformat=empty}
    \centering
    \begin{subfigure}[b]{0.24\linewidth}
        \centering
        \includegraphics[width=\linewidth]{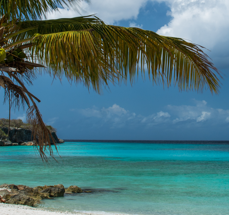}
   
    \end{subfigure}
     \begin{subfigure}[b]{0.24\linewidth}
        \centering
        \includegraphics[width=\linewidth]{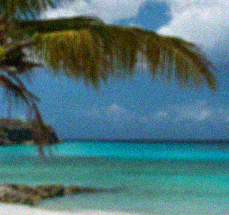}
   
    \end{subfigure}
    \begin{subfigure}[b]{0.24\linewidth}
        \centering
        \includegraphics[width=\linewidth]{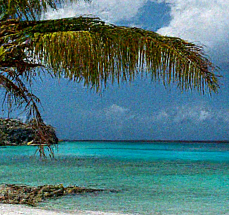}

    \end{subfigure}
    \begin{subfigure}[b]{0.24\linewidth}
        \centering
        \includegraphics[width=\linewidth]{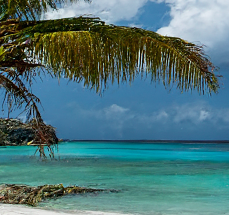}
    \end{subfigure}\\
    \begin{subfigure}[b]{0.24\linewidth}
        \centering
        \includegraphics[width=\linewidth]{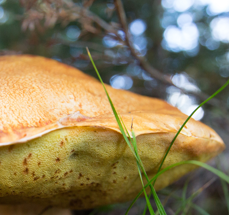}
   
    \end{subfigure}
     \begin{subfigure}[b]{0.24\linewidth}
        \centering
        \includegraphics[width=\linewidth]{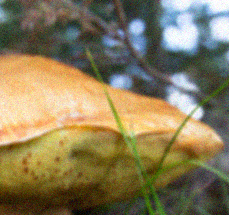}
   
    \end{subfigure}
    \begin{subfigure}[b]{0.24\linewidth}
        \centering
        \includegraphics[width=\linewidth]{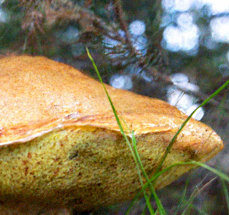}

    \end{subfigure}
    \begin{subfigure}[b]{0.24\linewidth}
        \centering
        \includegraphics[width=\linewidth]{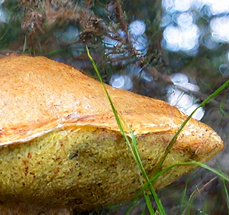}
    \end{subfigure}\\
    \begin{subfigure}[b]{0.24\linewidth}
        \centering
        \includegraphics[width=\linewidth]{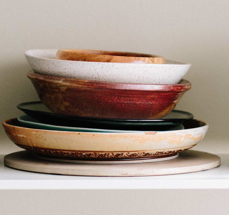}
   
    \end{subfigure}
     \begin{subfigure}[b]{0.24\linewidth}
        \centering
        \includegraphics[width=\linewidth]{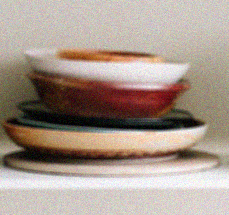}
   
    \end{subfigure}
    \begin{subfigure}[b]{0.24\linewidth}
        \centering
        \includegraphics[width=\linewidth]{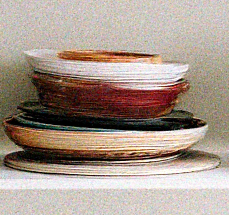}

    \end{subfigure}
    \begin{subfigure}[b]{0.24\linewidth}
        \centering
        \includegraphics[width=\linewidth]{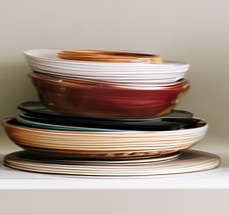}
    \end{subfigure}\\
    \begin{subfigure}[b]{0.24\linewidth}
        \centering
        \includegraphics[width=\linewidth]{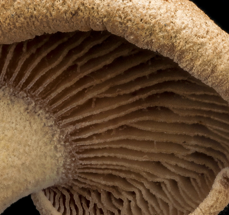}
   
    \end{subfigure}
     \begin{subfigure}[b]{0.24\linewidth}
        \centering
        \includegraphics[width=\linewidth]{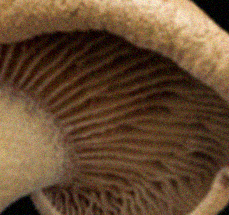}
   
    \end{subfigure}
    \begin{subfigure}[b]{0.24\linewidth}
        \centering
        \includegraphics[width=\linewidth]{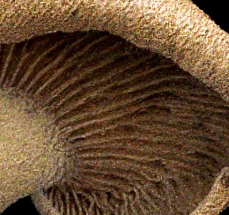}

    \end{subfigure}
    \begin{subfigure}[b]{0.24\linewidth}
        \centering
        \includegraphics[width=\linewidth]{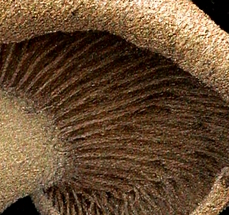}
    \end{subfigure}\\
    \begin{subfigure}[b]{0.24\linewidth}
        \centering
        \includegraphics[width=\linewidth]{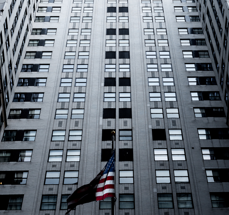}
        \subcaption{GT}
   
    \end{subfigure}
     \begin{subfigure}[b]{0.24\linewidth}
        \centering
        \includegraphics[width=\linewidth]{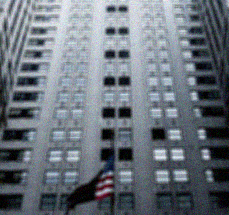}
        \subcaption{Blurry}
    \end{subfigure}
    \begin{subfigure}[b]{0.24\linewidth}
        \centering
        \includegraphics[width=\linewidth]{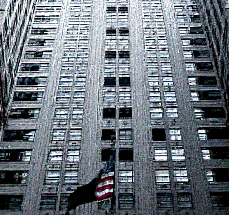}
        \subcaption{Guided Diffusion}
    \end{subfigure}
    \begin{subfigure}[b]{0.24\linewidth}
        \centering
        \includegraphics[width=\linewidth]{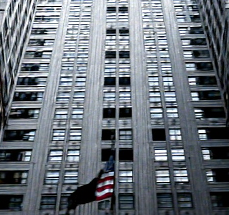}
        \subcaption{ADIR}
    \end{subfigure}
    \caption{Deblurring ($5\times5$ box filter, $\sigma = 10$) results of Guided Diffusion from section 3.2 and our method (ADIR), using the unconditional model from \citep{rombach2022high}. As can be seen from the images, our method outperforms guided diffusion in both sharpness and reconstruction details.}
\end{figure}

\begin{figure}
\captionsetup[subfigure]{labelformat=empty}
    \centering
    \begin{subfigure}[b]{0.24\linewidth}
        \centering
        \includegraphics[width=\linewidth]{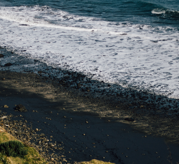}
   
    \end{subfigure}
     \begin{subfigure}[b]{0.24\linewidth}
        \centering
        \includegraphics[width=\linewidth]{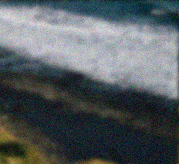}
   
    \end{subfigure}
    \begin{subfigure}[b]{0.24\linewidth}
        \centering
        \includegraphics[width=\linewidth]{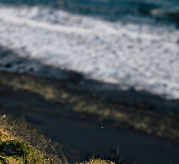}

    \end{subfigure}
    \begin{subfigure}[b]{0.24\linewidth}
        \centering
        \includegraphics[width=\linewidth]{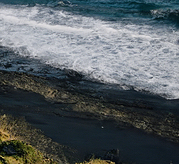}
    \end{subfigure}\\
    \begin{subfigure}[b]{0.24\linewidth}
        \centering
        \includegraphics[width=\linewidth]{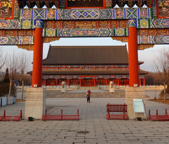}
   
    \end{subfigure}
     \begin{subfigure}[b]{0.24\linewidth}
        \centering
        \includegraphics[width=\linewidth]{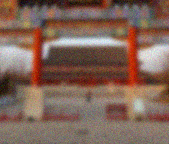}
   
    \end{subfigure}
    \begin{subfigure}[b]{0.24\linewidth}
        \centering
        \includegraphics[width=\linewidth]{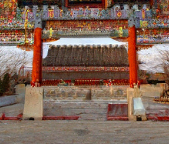}

    \end{subfigure}
    \begin{subfigure}[b]{0.24\linewidth}
        \centering
        \includegraphics[width=\linewidth]{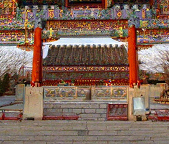}
    \end{subfigure}\\
    \begin{subfigure}[b]{0.24\linewidth}
        \centering
        \includegraphics[width=\linewidth]{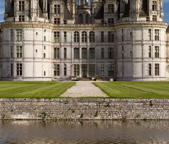}
   
    \end{subfigure}
     \begin{subfigure}[b]{0.24\linewidth}
        \centering
        \includegraphics[width=\linewidth]{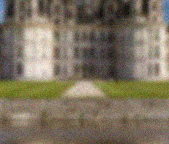}
   
    \end{subfigure}
    \begin{subfigure}[b]{0.24\linewidth}
        \centering
        \includegraphics[width=\linewidth]{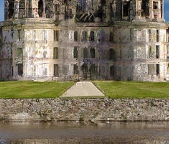}

    \end{subfigure}
    \begin{subfigure}[b]{0.24\linewidth}
        \centering
        \includegraphics[width=\linewidth]{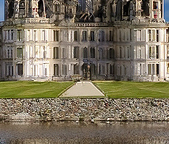}
    \end{subfigure}\\
    \begin{subfigure}[b]{0.24\linewidth}
        \centering
        \includegraphics[width=\linewidth]{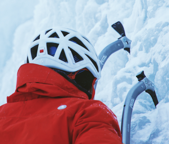}
   
    \end{subfigure}
     \begin{subfigure}[b]{0.24\linewidth}
        \centering
        \includegraphics[width=\linewidth]{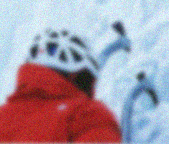}
   
    \end{subfigure}
    \begin{subfigure}[b]{0.24\linewidth}
        \centering
        \includegraphics[width=\linewidth]{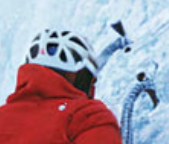}

    \end{subfigure}
    \begin{subfigure}[b]{0.24\linewidth}
        \centering
        \includegraphics[width=\linewidth]{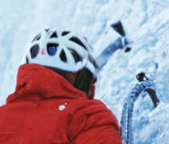}
    \end{subfigure}\\
    \begin{subfigure}[b]{0.24\linewidth}
        \centering
        \includegraphics[width=\linewidth]{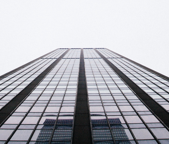}
        \subcaption{GT}
   
    \end{subfigure}
     \begin{subfigure}[b]{0.24\linewidth}
        \centering
        \includegraphics[width=\linewidth]{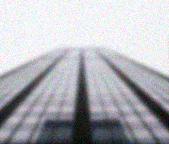}
        \subcaption{Bicubic}
    \end{subfigure}
    \begin{subfigure}[b]{0.24\linewidth}
        \centering
        \includegraphics[width=\linewidth]{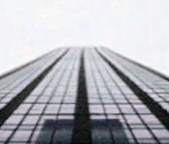}
        \subcaption{Guided Diffusion}
    \end{subfigure}
    \begin{subfigure}[b]{0.24\linewidth}
        \centering
        \includegraphics[width=\linewidth]{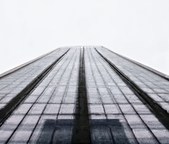}
        \subcaption{ADIR}
    \end{subfigure}
    \caption{Gaussian deblurring ($\sigma_\text{blur}=2$ and $\sigma_\text{noise}=10$) results of Guided Diffusion from section 3.2 and our method (ADIR), using the unconditional model from \citep{rombach2022high}. As can be seen from the images, our method outperforms guided diffusion in both sharpness and reconstruction details.}
\end{figure}

\begin{figure}
\captionsetup[subfigure]{labelformat=empty}
    \centering
    \begin{subfigure}[b]{0.2\linewidth}
        \centering
        \includegraphics[width=\linewidth]{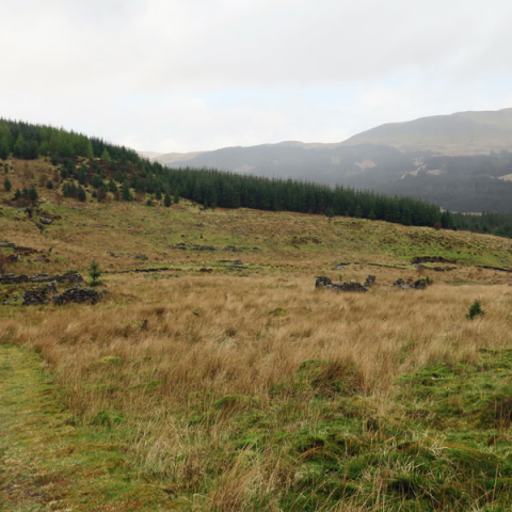}
        \subcaption{Original Image}
    \end{subfigure}
    \begin{subfigure}[b]{0.2\linewidth}
        \centering
        \includegraphics[width=\linewidth]{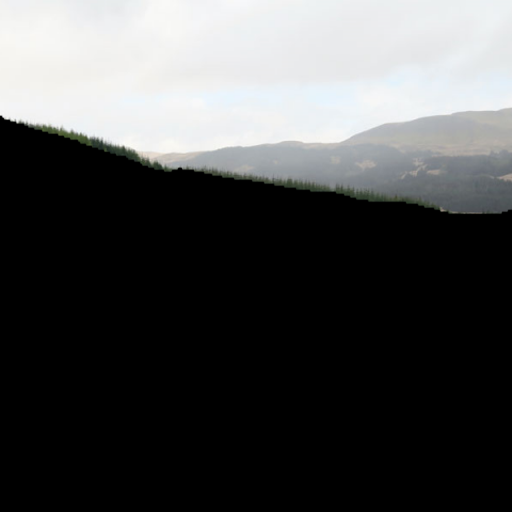}
        \subcaption{Masked}
    \end{subfigure}\\
    \begin{subfigure}[b]{0.45\linewidth}
        \centering
        \includegraphics[width=\linewidth]{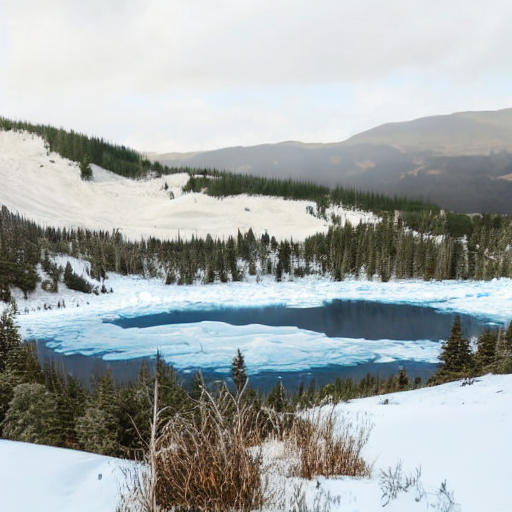}
    \end{subfigure}
     \begin{subfigure}[b]{0.45\linewidth}
        \centering
        \includegraphics[width=\linewidth]{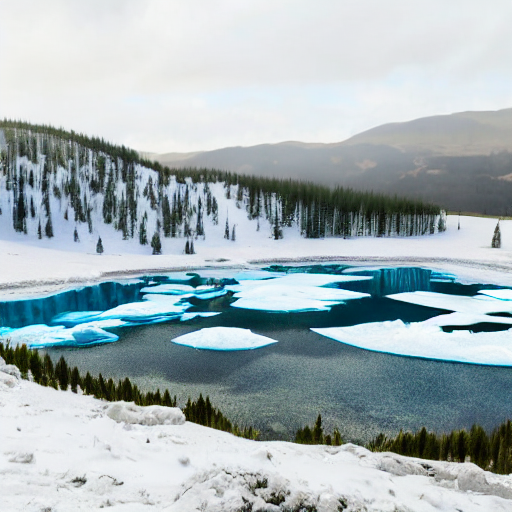}
    \end{subfigure}
    \\Stable Diffusion \\ 
    
    \begin{subfigure}[b]{0.45\linewidth}
        \centering
        \includegraphics[width=\linewidth]{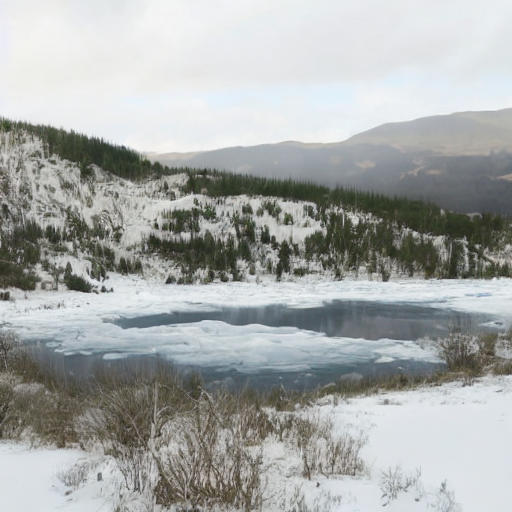}
   
    \end{subfigure}
     \begin{subfigure}[b]{0.45\linewidth}
        \centering
        \includegraphics[width=\linewidth]{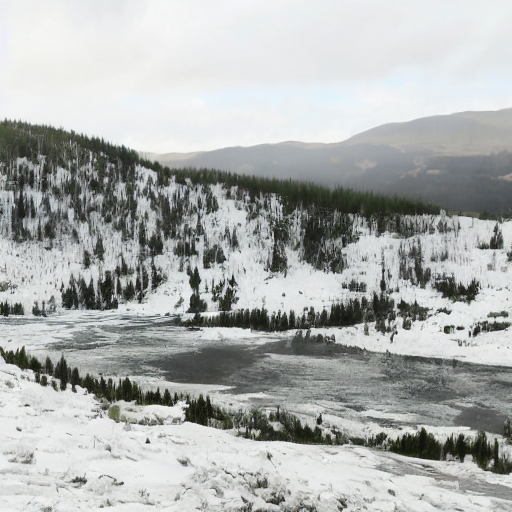}
    \end{subfigure}
   \\ ADIR \\
    \caption{Text-based image editing comparison between Stable Diffusion \citep{rombach2022high} and ADIR, using the prompt ``A beautiful frozen lake between mountains in the snow'' for two different seeds.}
    
\end{figure}

\begin{figure*}[th]
\captionsetup[subfigure]{labelformat=empty}
    \centering
    \begin{subfigure}[b]{0.2\linewidth}
        \centering
        \includegraphics[width=\linewidth]{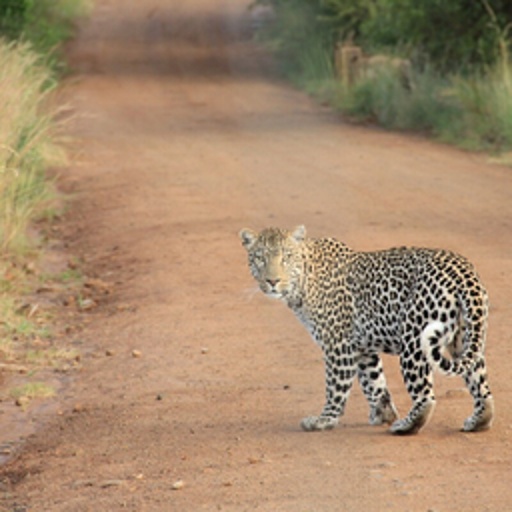}
        \subcaption{Original Image}
    \end{subfigure}
    \begin{subfigure}[b]{0.2\linewidth}
        \centering
        \includegraphics[width=\linewidth]{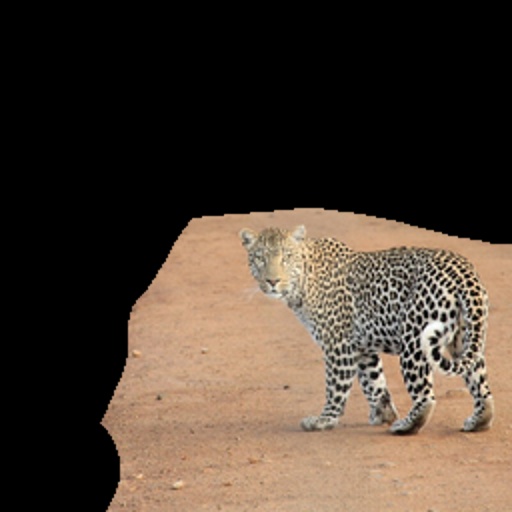}
        \subcaption{Masked}
    \end{subfigure}\\
    \begin{subfigure}[b]{0.45\linewidth}
        \centering
        \includegraphics[width=\linewidth]{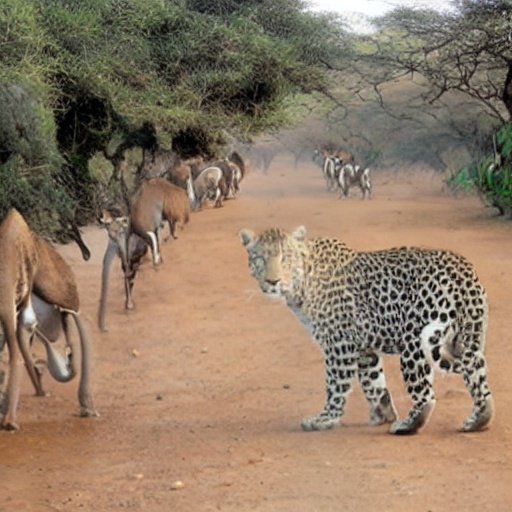}
    \end{subfigure}
    \begin{subfigure}[b]{0.45\linewidth}
        \centering
        \includegraphics[width=\linewidth]{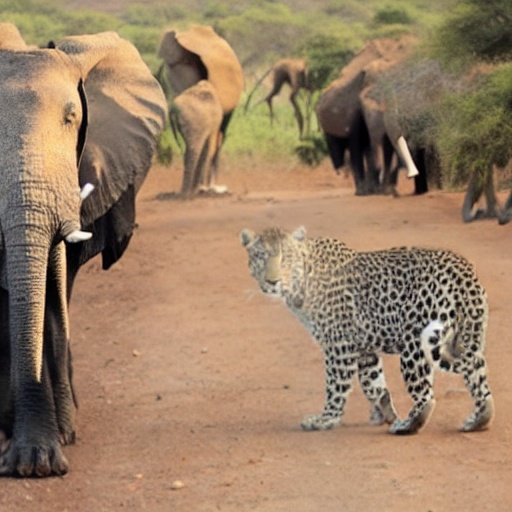}
    \end{subfigure}
    \\Stable Diffusion\\
    \begin{subfigure}[b]{0.45\linewidth}
        \centering
        \includegraphics[width=\linewidth]{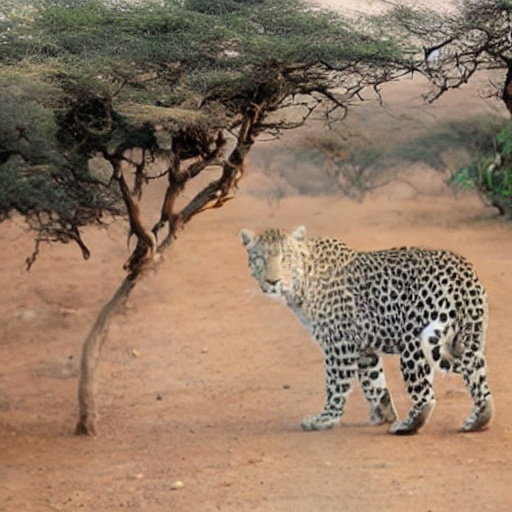}
    \end{subfigure}
    \begin{subfigure}[b]{0.45\linewidth}
        \centering
        \includegraphics[width=\linewidth]{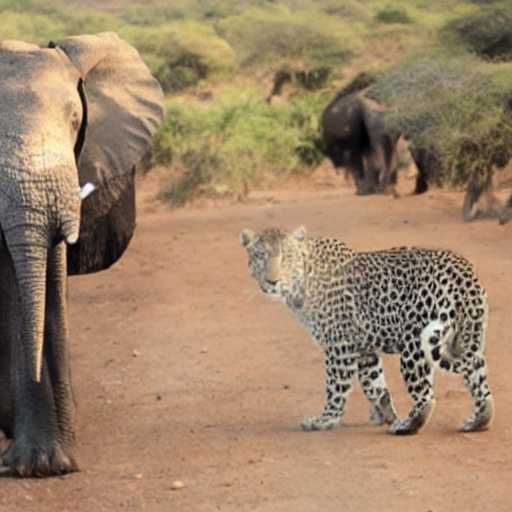}
    \end{subfigure}
    \\ADIR\\
    \caption{Text-based editing comparison between Stable Diffusion and ADIR, using the prompt ``Africa'' for two different seeds. Note that Stable diffusion adds partial animals while ADIR completes the scene more naturally.}
\end{figure*}

\begin{figure}
\captionsetup[subfigure]{labelformat=empty}
    \centering
    \begin{subfigure}[b]{0.2\linewidth}
        \centering
        \includegraphics[width=\linewidth]{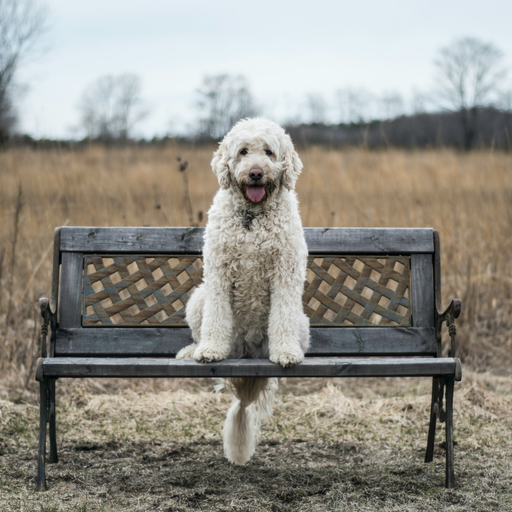}
        \subcaption{Original Image}
    \end{subfigure}
    \begin{subfigure}[b]{0.2\linewidth}
        \centering
        \includegraphics[width=\linewidth]{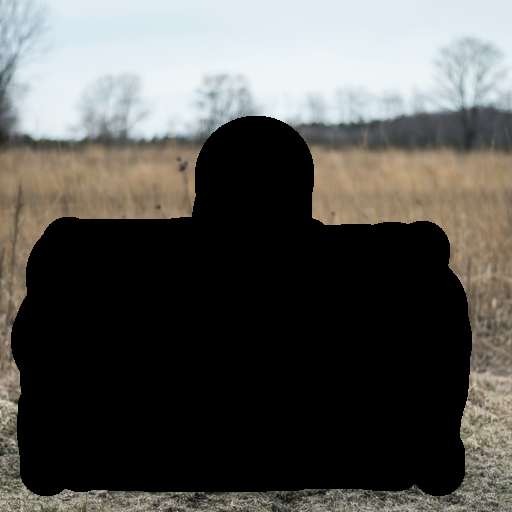}
        \subcaption{Masked}
    \end{subfigure}\\
    \begin{subfigure}[b]{0.45\linewidth}
        \centering
        \includegraphics[width=\linewidth]{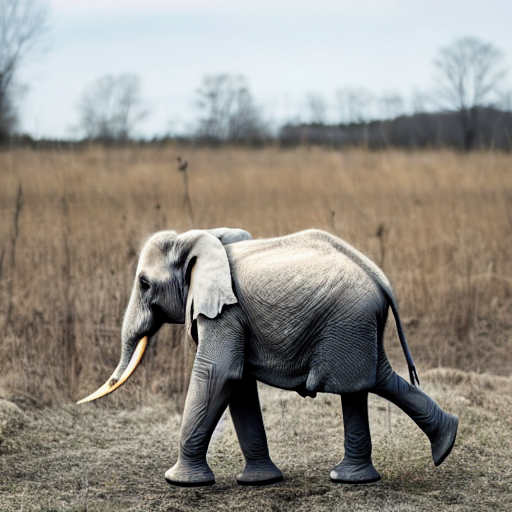}
   
    \end{subfigure}
    \begin{subfigure}[b]{0.45\linewidth}
        \centering
        \includegraphics[width=\linewidth]{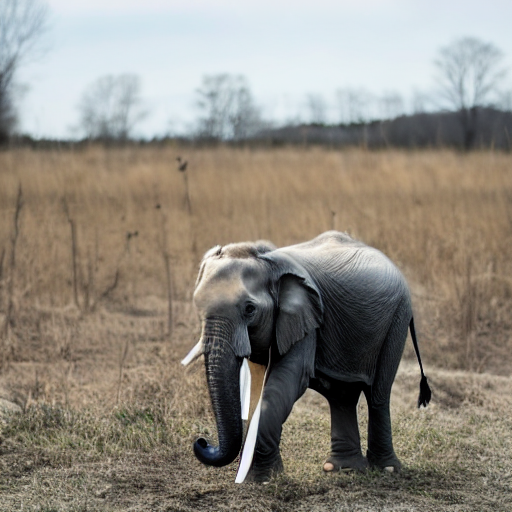}
    \end{subfigure}
    \\ Stable Diffusion \\
    \begin{subfigure}[b]{0.45\linewidth}
        \centering
        \includegraphics[width=\linewidth]{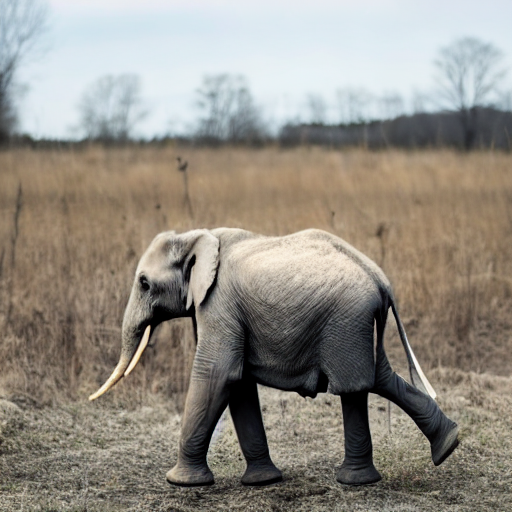}
   
    \end{subfigure}
    \begin{subfigure}[b]{0.45\linewidth}
        \centering
        \includegraphics[width=\linewidth]{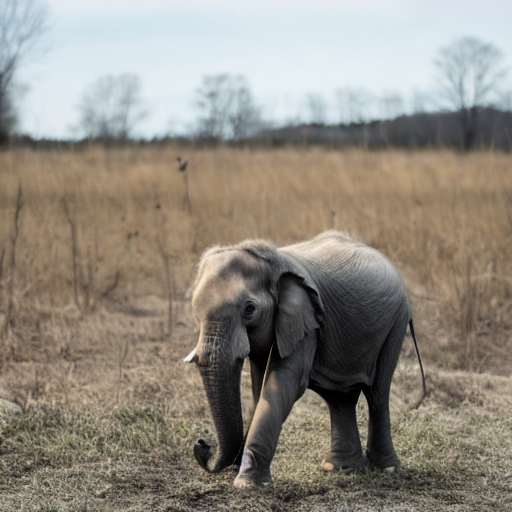}
    \end{subfigure}
    \\ADIR \\
    \caption{Text-based image editing comparison between Stable Diffusion \citep{rombach2022high} and ADIR, using the prompt ``An elephant walking'' for two different seeds.}
    
\end{figure}

\begin{figure}
\captionsetup[subfigure]{labelformat=empty}
    \centering
    \begin{subfigure}[b]{0.2\linewidth}
        \centering
        \includegraphics[width=\linewidth]{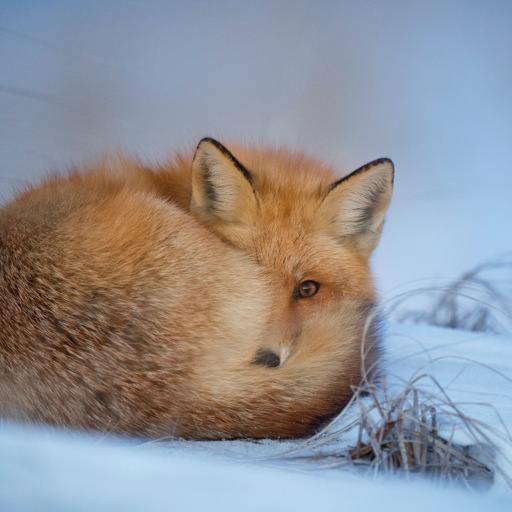}
        \subcaption{Original Image}
    \end{subfigure}
    \begin{subfigure}[b]{0.2\linewidth}
        \centering
        \includegraphics[width=\linewidth]{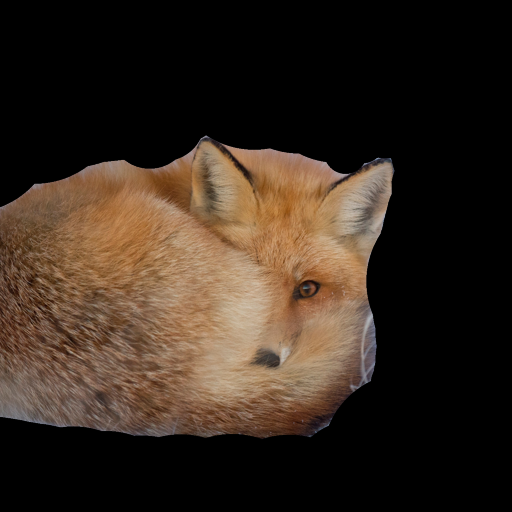}
        \subcaption{Masked}
    \end{subfigure}\\
    \begin{subfigure}[b]{0.45\linewidth}
        \centering
        \includegraphics[width=\linewidth]{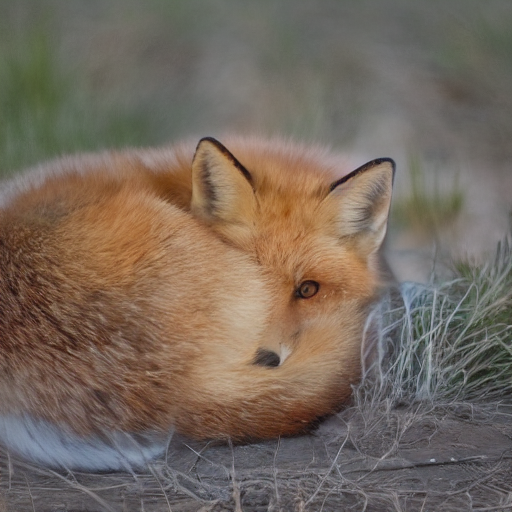}
   
    \end{subfigure}
    \begin{subfigure}[b]{0.45\linewidth}
        \centering
        \includegraphics[width=\linewidth]{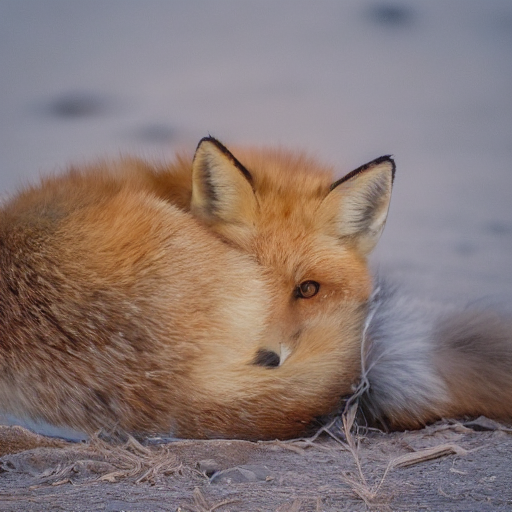}
    \end{subfigure}
    \\ Stable Diffusion \\
    \begin{subfigure}[b]{0.45\linewidth}
        \centering
        \includegraphics[width=\linewidth]{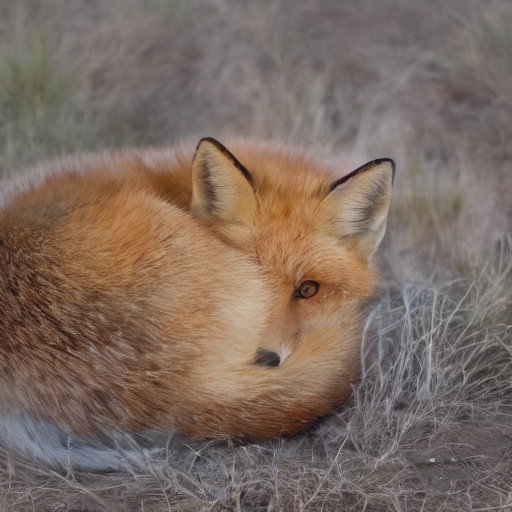}
   
    \end{subfigure}
    \begin{subfigure}[b]{0.45\linewidth}
        \centering
        \includegraphics[width=\linewidth]{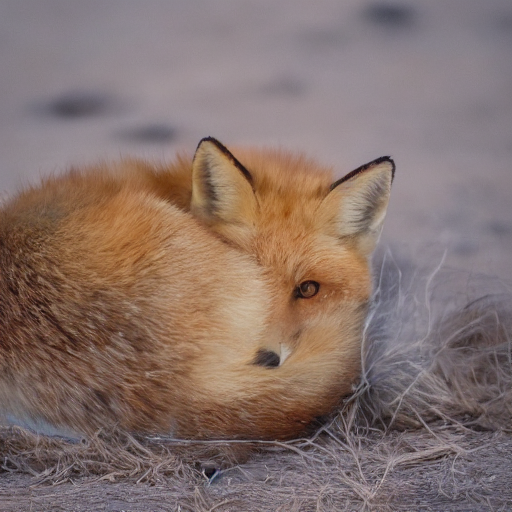}
    \end{subfigure}
    \\ADIR \\
    \caption{Text-based image editing comparison between Stable Diffusion \citep{rombach2022high} and ADIR applied to the Stable Diffusion model, for the prompt ``A fox sitting in the middle of the desert''}
    
\end{figure}

\begin{figure}
\captionsetup[subfigure]{labelformat=empty}
    \centering
    \begin{subfigure}[b]{0.2\linewidth}
        \centering
        \includegraphics[width=\linewidth]{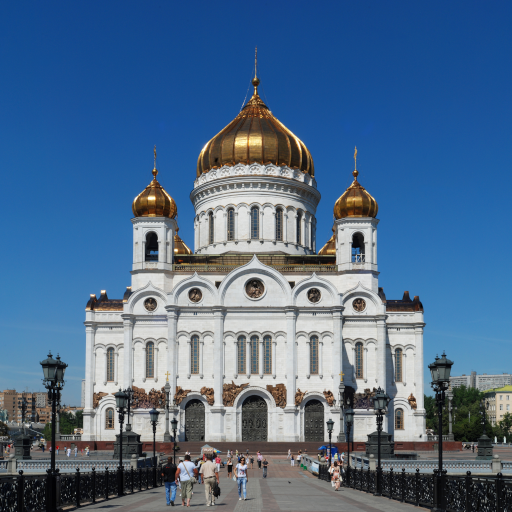}
        \subcaption{Original Image}
    \end{subfigure}
    \begin{subfigure}[b]{0.2\linewidth}
        \centering
        \includegraphics[width=\linewidth]{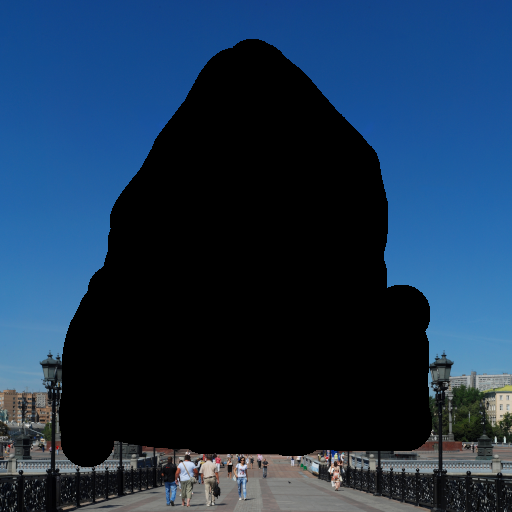}
        \subcaption{Masked}
    \end{subfigure}\\
    \begin{subfigure}[b]{0.45\linewidth}
        \centering
        \includegraphics[width=\linewidth]{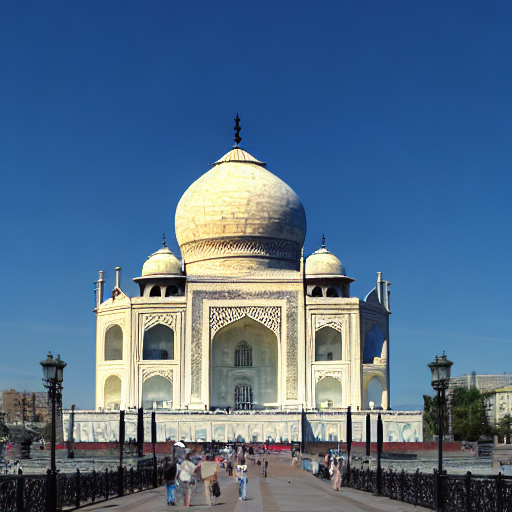}
    \end{subfigure}
     \begin{subfigure}[b]{0.45\linewidth}
        \centering
        \includegraphics[width=\linewidth]{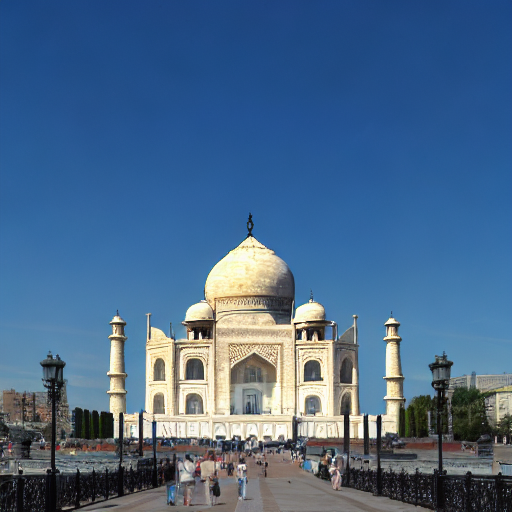}
    \end{subfigure}
    \\ Stable Diffusion \\
    \begin{subfigure}[b]{0.45\linewidth}
        \centering
        \includegraphics[width=\linewidth]{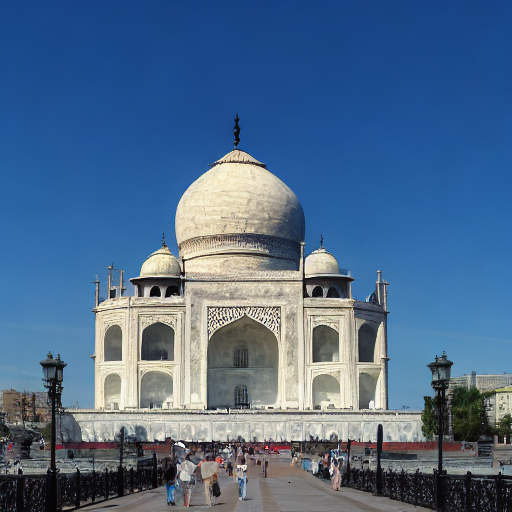}
    \end{subfigure}
     \begin{subfigure}[b]{0.45\linewidth}
        \centering
        \includegraphics[width=\linewidth]{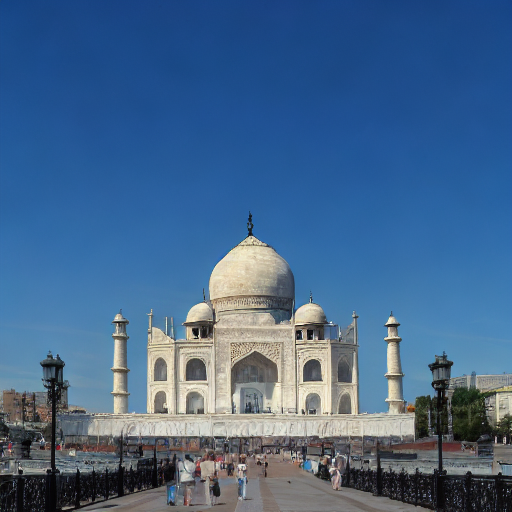}
    \end{subfigure}
    \\ ADIR \\
    \caption{Text-based image editing comparison between Stable Diffusion \citep{rombach2022high} and ADIR applied to the Stable Diffusion model, for the prompt ``Taj Mahal''}
    
\end{figure}

\begin{figure}[t]
\captionsetup[subfigure]{labelformat=empty}
    \centering
    ``A vase of flowers on the table of a living room'' \\
    \begin{subfigure}[b]{0.24\linewidth}
        \centering
        \includegraphics[width=\linewidth]{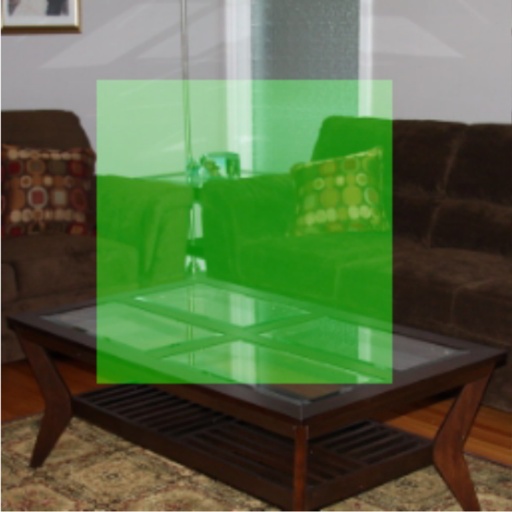}
   
    \end{subfigure}
    \begin{subfigure}[b]{0.24\linewidth}
        \centering
        \includegraphics[width=\linewidth]{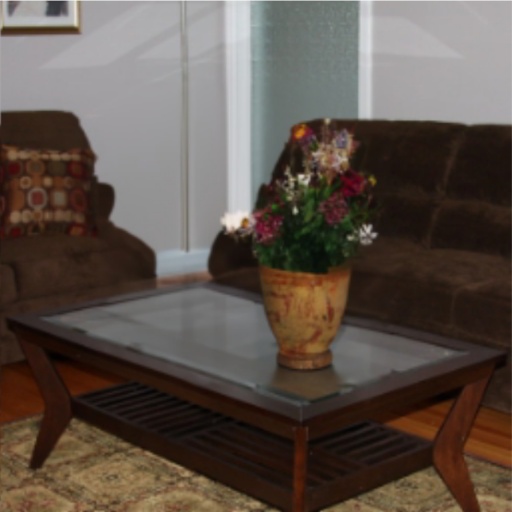}
    \end{subfigure}
     \begin{subfigure}[b]{0.24\linewidth}
        \centering
        \includegraphics[width=\linewidth]{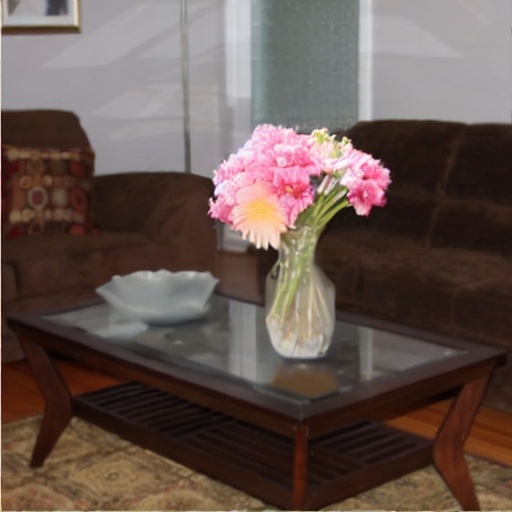}
   
    \end{subfigure}
    \begin{subfigure}[b]{0.24\linewidth}
        \centering
        \includegraphics[width=\linewidth]{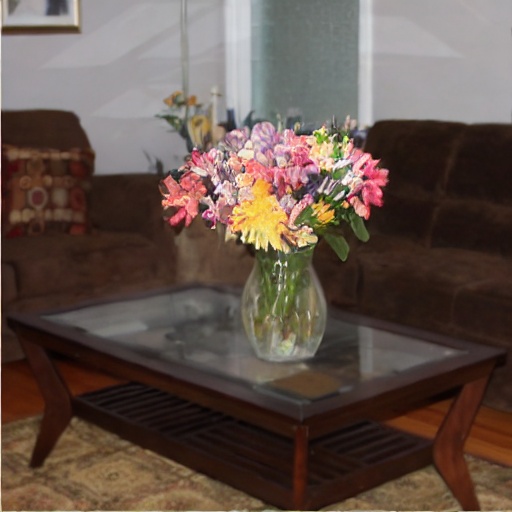}

    \end{subfigure}\\
    ``A man with red hair'' \\
    \begin{subfigure}[b]{0.24\linewidth}
        \centering
        \includegraphics[width=\linewidth]{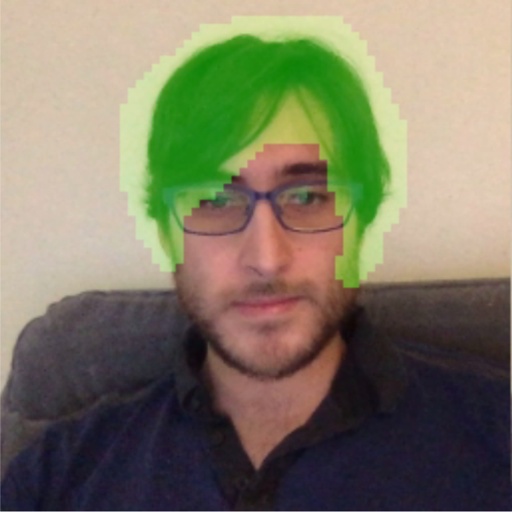}
   
    \end{subfigure}
    \begin{subfigure}[b]{0.24\linewidth}
        \centering
        \includegraphics[width=\linewidth]{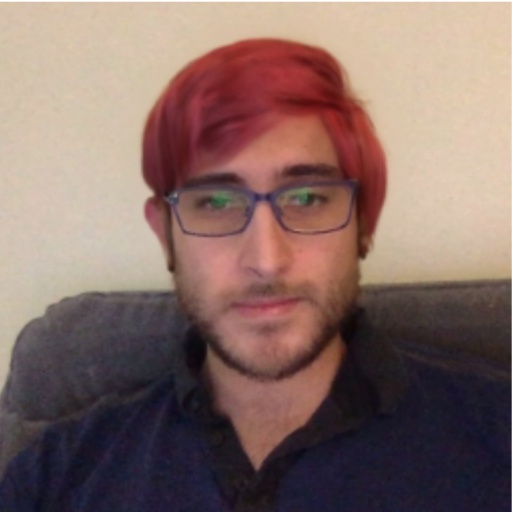}
    \end{subfigure}
     \begin{subfigure}[b]{0.24\linewidth}
        \centering
        \includegraphics[width=\linewidth]{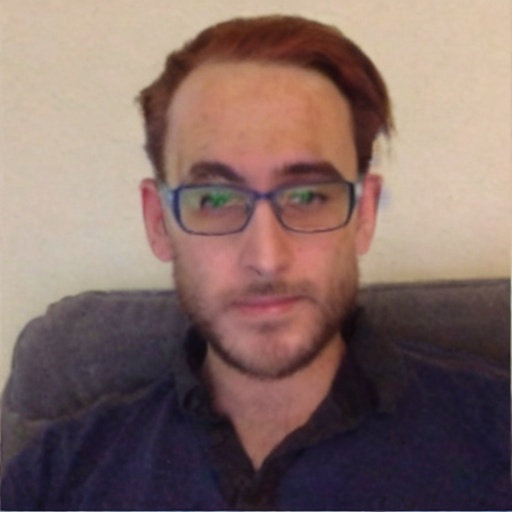}
   
    \end{subfigure}
    \begin{subfigure}[b]{0.24\linewidth}
        \centering
        \includegraphics[width=\linewidth]{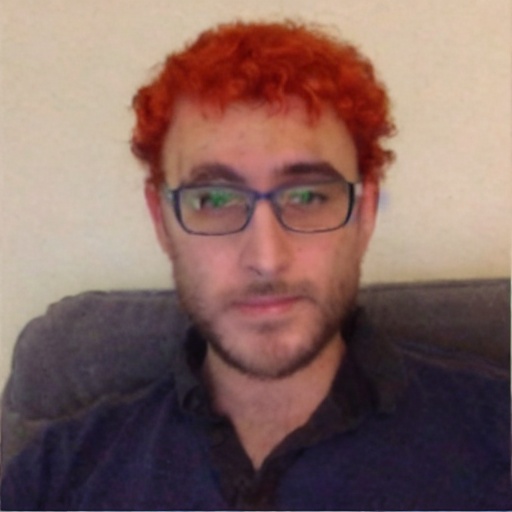}

    \end{subfigure}
    ``An old car in a snowy forest'' \\
    \begin{subfigure}[b]{0.24\linewidth}
        \centering
        \includegraphics[width=\linewidth]{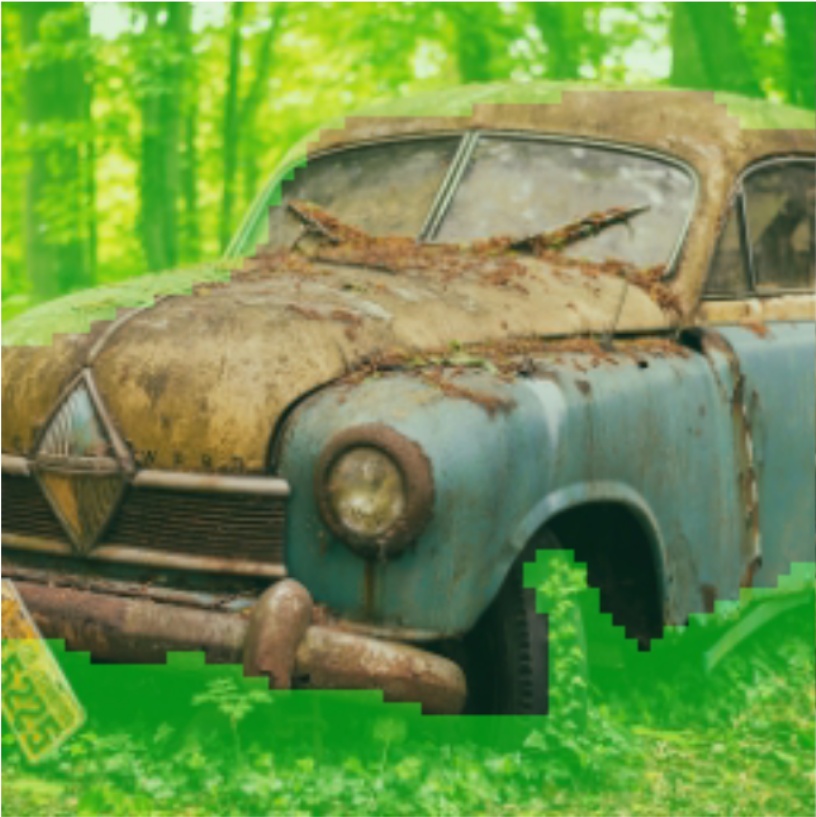}   
    \end{subfigure}
     \begin{subfigure}[b]{0.24\linewidth}
        \centering
        \includegraphics[width=\linewidth]{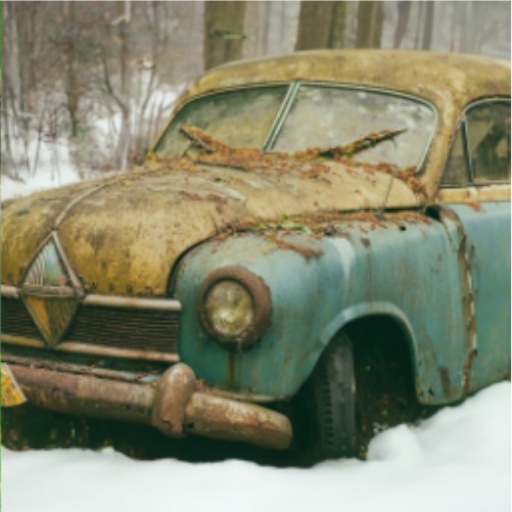}
    \end{subfigure}
    \begin{subfigure}[b]{0.24\linewidth}
        \centering
        \includegraphics[width=\linewidth]{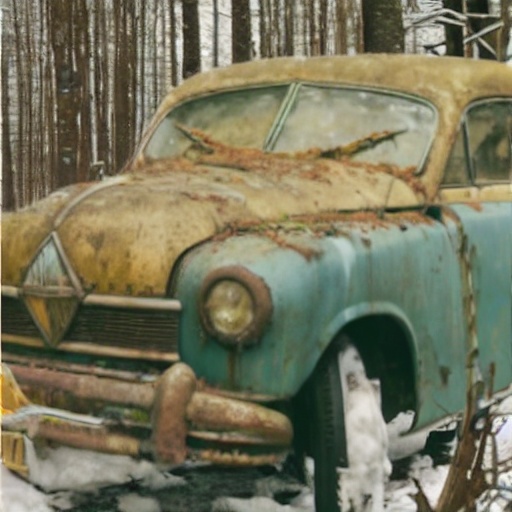}
    \end{subfigure}
    \begin{subfigure}[b]{0.24\linewidth}
        \centering
\includegraphics[width=\linewidth]{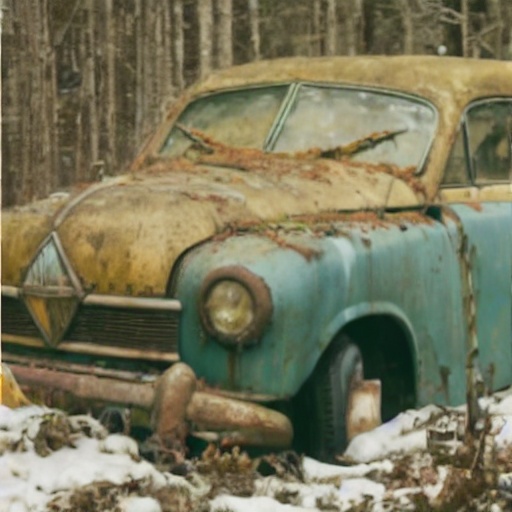}
\end{subfigure}
    \caption{Text-based image editing comparison between GLIDE (full) \citep{nichol2021glide}, Stable Diffusion \citep{rombach2022high} and ADIR applied to the Stable Diffusion model. The images are taken from \citep{nichol2021glide}, since their official high-res model was not publicly released. As can be seen, our method produces more realistic images in cases where Stable Diffusion either was not accurate (brown hair instead of red) or in terms of artifacts.}
    \label{fig:Editing_GLIDE}
\end{figure}

\begin{figure*}
    \includegraphics[width=0.19\linewidth]{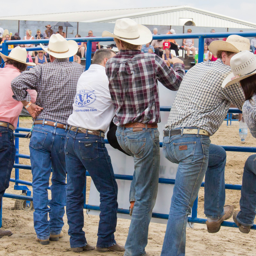}
    \includegraphics[width=0.19\linewidth]{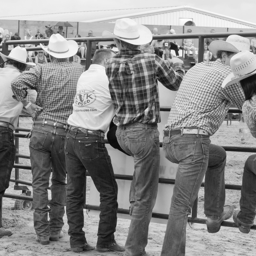}
    \includegraphics[width=0.19\linewidth]{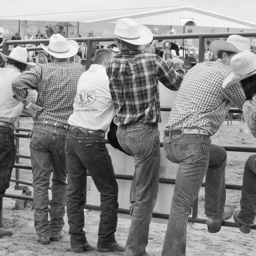}
    \includegraphics[width=0.19\linewidth]{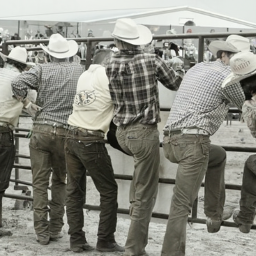}
    \includegraphics[width=0.19\linewidth]{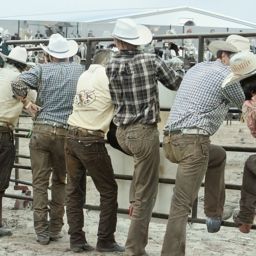}\\
    \includegraphics[width=0.19\linewidth]{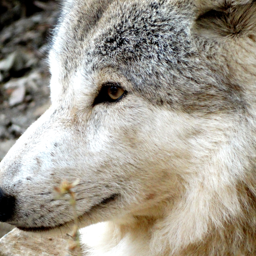}
    \includegraphics[width=0.19\linewidth]{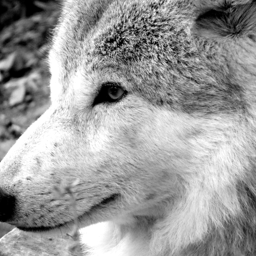}
    \includegraphics[width=0.19\linewidth]{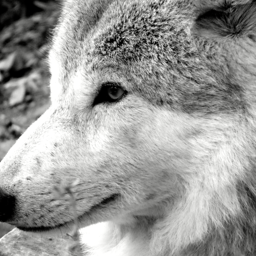}
    \includegraphics[width=0.19\linewidth]{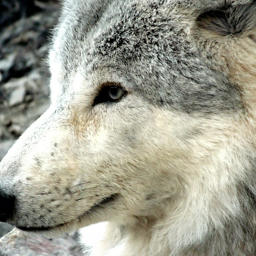}
    \includegraphics[width=0.19\linewidth]{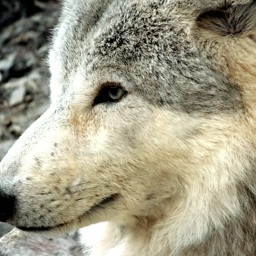}\\
    \includegraphics[width=0.19\linewidth]{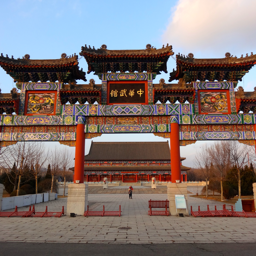}
    \includegraphics[width=0.19\linewidth]{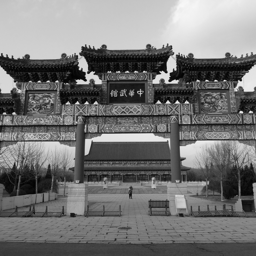}
    \includegraphics[width=0.19\linewidth]{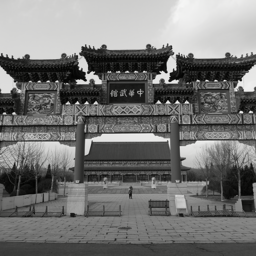}
    \includegraphics[width=0.19\linewidth]{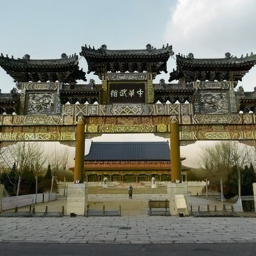}
    \includegraphics[width=0.19\linewidth]{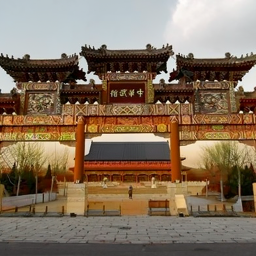}\\
    \includegraphics[width=0.19\linewidth]{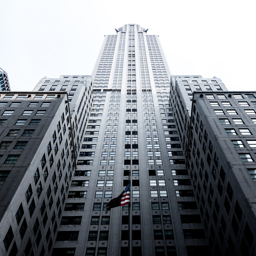}
    \includegraphics[width=0.19\linewidth]{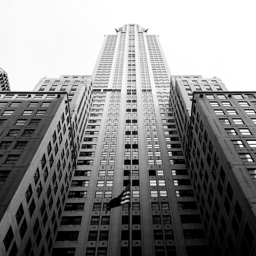}
    \includegraphics[width=0.19\linewidth]{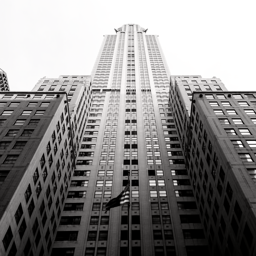}
    \includegraphics[width=0.19\linewidth]{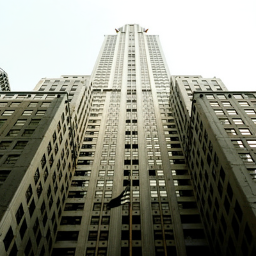}
    \includegraphics[width=0.19\linewidth]{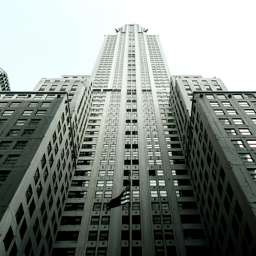}\\
    \subcaptionbox*{Original Image}{\includegraphics[width=0.19\linewidth]{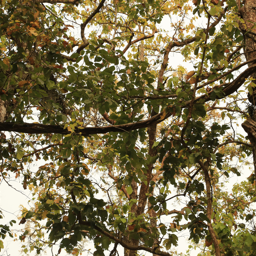}}
    \subcaptionbox*{Grayscale}{\includegraphics[width=0.19\linewidth]{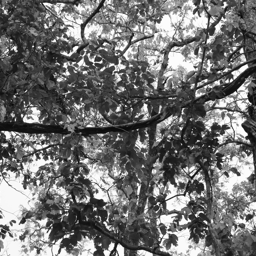}}
    \subcaptionbox*{DDRM}{\includegraphics[width=0.19\linewidth]{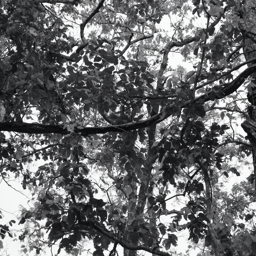}}
    \subcaptionbox*{Guided Diffusion}{\includegraphics[width=0.19\linewidth]{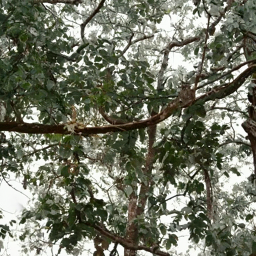}}
    \subcaptionbox*{ADIR}{\includegraphics[width=0.19\linewidth]{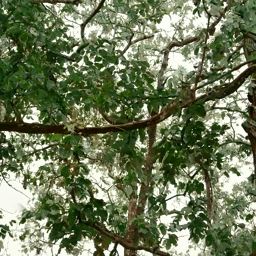}}\\
  \caption{Image colorization results comparison between DDRM \citep{kawar2022denoising}, Guided diffusion proposed in section 3.2, and our adaptive approach ADIR. As can be seen, adapting the denoiser network to the given image can improve the results significantly.}
  \label{fig:GD_color2}
\end{figure*}

\end{document}